\documentclass{article}

\usepackage[preprint]{neurips_2023}

\usepackage[utf8]{inputenc} % allow utf-8 input
\usepackage[T1]{fontenc}    % use 8-bit T1 fonts
\usepackage{hyperref}       % hyperlinks
\usepackage{url}            % simple URL typesetting
\usepackage{booktabs}       % professional-quality tables
\usepackage{amsfonts}       % blackboard math symbols
\usepackage{nicefrac}       % compact symbols for 1/2, etc.
\usepackage{microtype}      % microtypography
\usepackage{xcolor}         % colors
\usepackage{graphicx}
\usepackage{comment}
\usepackage{amsmath}
\usepackage{amssymb}
\usepackage{mathtools}
\usepackage{amsthm}
\usepackage{bbm}
\usepackage{graphbox}
\usepackage{subfigure}
\usepackage{booktabs} % for professional tables
\usepackage{soul} %remove later, added for strikethrough
\usepackage{multirow}
\usepackage{float}
\usepackage{todonotes}
\definecolor{applegreen}{rgb}{0.01, 0.75, 0.24}
\definecolor{brightmaroon}{rgb}{0.66, 0.13, 0.24}
\definecolor{warningyellow}{rgb}{0.86, 0.58, 0.17}

\title{Unveiling the Hessian's Connection to~the~Decision~Boundary}

\author{%
  Mahalakshmi Sabanayagam\thanks{Equal contribution.}
  \\
  School of Computation, Information and Technology,
  Technical University of Munich, Germany \\
  \texttt{sabanaya@cit.tum.de} \\
  \And
  Freya Behrens$^*$ \\
  Statistical Physics of Computation Lab, %\\
  École Polytechnique Fédérale de Lausanne, Switzerland \\
  \texttt{freya.behrens@epfl.ch} \\
  \And
  Urte Adomaityte \\
  Department of Mathematics, King’s College London, United Kingdom \\
  \texttt{urte.adomaityte@kcl.ac.uk} \\
  \And
  Anna Dawid\thanks{Corresponding author.} \\
  Center for Computational Quantum Physics, Flatiron Institute, USA\\
  \texttt{adawid@flatironinstitute.org}
}

\begin{document}

\maketitle

\begin{abstract}
Understanding the properties of well-generalizing minima is at the heart of deep learning research.
On the one hand, the generalization of neural networks has been connected to the decision boundary complexity, which is hard to study in the high-dimensional input space. Conversely, the flatness of a minimum has become a controversial proxy for generalization.
In this work, we provide the missing link between the two approaches and show that the Hessian top eigenvectors characterize the decision boundary learned by the neural network. Notably, the number of outliers in the Hessian spectrum is proportional to the complexity of the decision boundary. Based on this finding, we 
provide a new and straightforward approach to studying the complexity of a high-dimensional decision boundary;
show that this connection naturally inspires a new generalization measure;
and finally, we develop a novel margin estimation technique which, in combination with the generalization measure, precisely identifies minima with simple wide-margin boundaries. 
Overall, this analysis establishes the connection between the Hessian and the decision boundary and provides a new method to identify minima with simple wide-margin decision boundaries.
\end{abstract}

\section{Introduction}

The loss landscape of a deep neural network is a high-dimensional non-convex object exhibiting multiple equivalent local minima and saddle points \citep{auer1995exponentially, Dauphin2014, Choromanska2015, Sagun2016, Alain2019}.
The complex geometry of the loss landscape makes it notoriously difficult to analyze.
In the context of gradient descent-based optimization, it is widely observed that the network converges to a local minimum that generalizes reasonably well \citep{Keskar2017}. Still, the properties of minima exhibiting good generalization are highly debated.

%%%% DECISION BOUNDARY WORK
To understand those properties, some works study the decision boundary corresponding to a given minimum.
Researchers often follow Occam's razor by assuming that among minima with similarly high training accuracy, the ones with simpler decision boundaries will have a higher test accuracy \citep{guan2020analysis}.
Then, they attempt to define the complexity of the decision boundary using various approaches, e.g., using topological measures \citep{ramamurthy2018topological}, or curvature of the loss around the boundary in the input space \citep{Fawzi2018topology}, or via generation of adversarial examples \citep{guan2020analysis, Karimi2022DeepDIG}. Another proxy for the decision boundary complexity is the number of its linear segments \citep{Kienitz2023comparingcomplexities} but it is limited to the low-dimensional input space.

%%%% HESSIAN WORKS
Other works study minima by analyzing their curvature in the model parameter space and developing heuristics indicating their generalization abilities.
A few notable results suggest that flat minima generalize better than sharp minima \citep{Keskar2017, Wu2017, Izmailov2018, He2019}.
One approach to analyzing the curvature of the minimum is through the Hessian of the training loss. Specifically, the intuition that a flat minimum has a smaller sum of Hessian eigenvalues (trace) than a sharp minimum is used as a straightforward metric for generalization ~\citep{hochreiter1997flat, Keskar2017}. 
However, these results are extensively contested and discussed \citep{Zhang2021rethinking}, since flatness is not a well-defined concept in non-convex landscapes of deep models~\citep{Dinh2017}.
On the one hand, works such as \citet{Sagun2018, jastrzebski2019relation, petzka2021relative, andriushchenko2023modern} illustrate the superfluousness of Hessian-based generalization measures and observe that flatness is not well correlated with generalization.
On the other hand, \citet{kwon2021asam} and \citet{petzka2021relative} suggest improvements to the direct Hessian-based metric and compute the adaptive flatness and relative flatness of the minimum, respectively.

%%%%% OUR PART: We are the missing link
Despite the intuitive understanding that the simple decision boundary and a properly defined flatness of minima together promote good generalization of neural networks, 
to the best of our knowledge, no explicit connection between the two has been established so far.
Advancing an understanding of this connection is precisely the goal of this work.
%%%%% What info we leverage
To do so, we take a closer look at properties of the Hessian that are observed to be universal across different deep learning setups. 
Firstly, the spectrum of the Hessian at a minimum separates into the bulk centered around zero and a few outliers, whose number is roughly equal to the number of classes in the data \citep{Sagun2016, Sagun2018, Ghorbani2019, Papyan2019structure, Papyan2020structure}. We ask \emph{what is the significance of those outliers, and why is their number approximately equal to the number of classes?}
Secondly, \emph{why does the gradient information reside in a small subspace spanned by the Hessian top eigenvectors} as noted by \cite{GurAri2019}?

%%% TODO PAPAYAN
In our work, we give an understanding of these properties by revealing their connection to the decision boundary.
In particular, we compare the gradient directions of the loss for individual training data with the Hessian eigenvectors and see that they align when the samples are at the decision boundary. As a consequence, we propose a new generalization measure and a margin estimation technique that show promising empirical success in capturing the generalization of neural networks.

\textbf{Contributions.} We perform a rigorous numerical analysis of the deep neural network loss landscape for classification tasks through the Hessian of the training loss, and we observe the following: \\
%%%
    \textbf{(1)} The top eigenvectors of the Hessian of the training loss encode the decision boundary learned by the neural network. In particular, there is a clear information separation across the eigenvectors, which encode separate sections of the decision boundary. \\
    % \item 
    \textbf{(2)} The number of encoding eigenvectors is usually equal to the number of spectrum outliers which is directly proportional to the complexity of the decision boundary. To elaborate, more eigenvectors are needed to encode a complex, highly non-linear decision boundary than a simpler counterpart. \\
    % \item 
    \textbf{(3)} We propose a new, improved generalization measure that considers the simplicity of the decision boundary via the Hessian eigenvectors. In addition, we develop a technique to estimate the narrowest margin of the decision boundary in the input space.
%%%

\begin{figure*}[ht]
    \centering
    \includegraphics[trim=0 1cm 0 0, clip, width=0.98\textwidth]{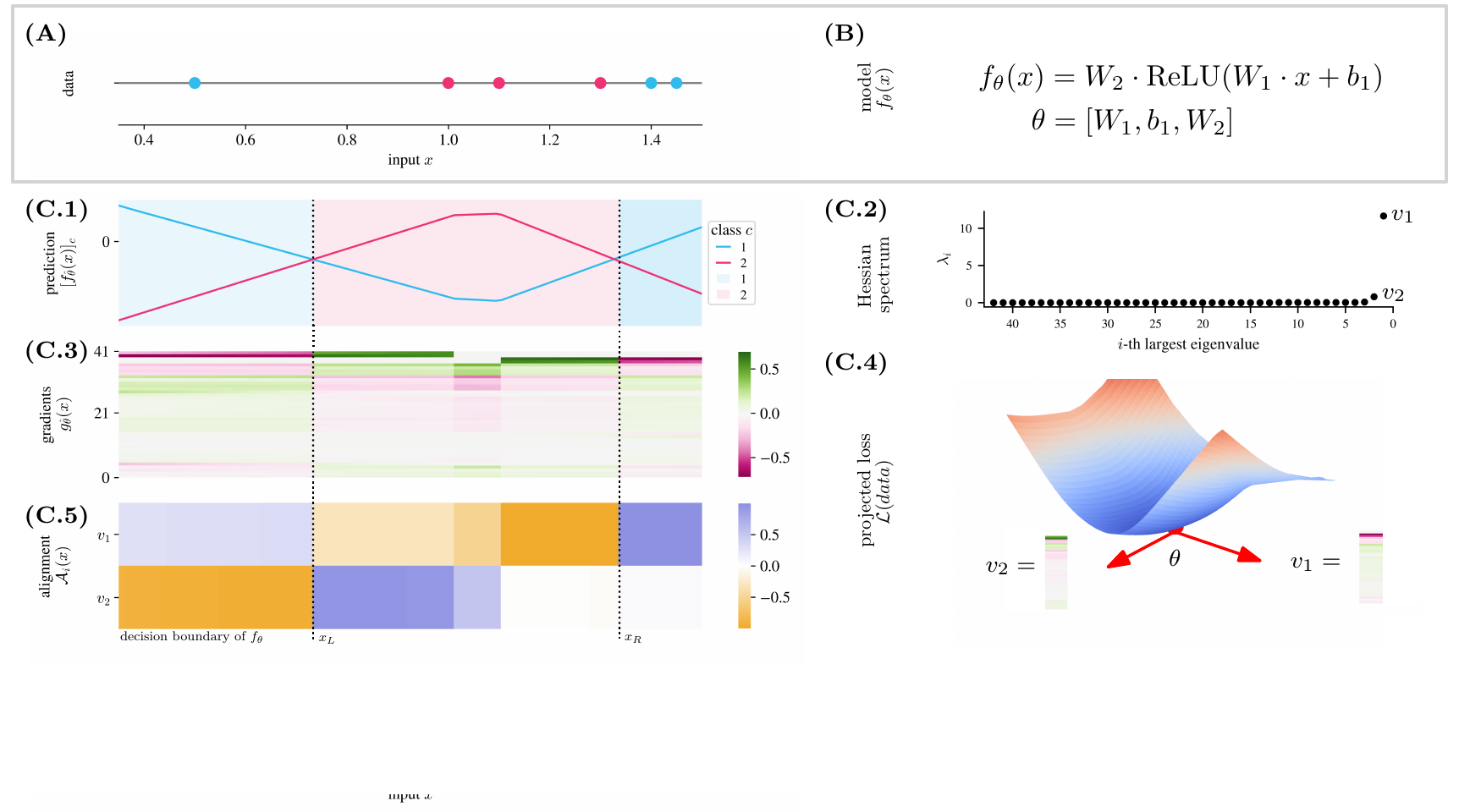}\vspace{-2em}
\caption{\textbf{Analysis of the Hessian.} 
\textbf{(A)} 1D toy dataset with 5 input points and 2 classes $\{$pink, cyan$\}$. 
\textbf{(B)} A model $f_{\theta}$ parameterized by $\theta$ that takes an input $x$ and returns logit probabilities for each class. % returns a value for each class for an input $x$. 
\textbf{(C.1)} Predictions of $f_{\hat{\theta}}$ across the input space with $\hat{\theta}$ being a specific set of parameters that correctly classify the training data. 
\textbf{(C.2)} There are two outliers in the Hessian eigenspectrum of the training loss calculated at the minimum $\hat{\theta}$. They correspond to the eigenvectors $v_1$ and $v_2$ that are directions in the parameter space shown in \textbf{(C.4)}.
When measuring their cosine similarity with gradients of the loss of individual points from the input space \textbf{(C.3)}, we obtain the alignment in \textbf{(C.5)}.
Each outlier encodes one section of the decision boundary of $f_{\hat{\theta}}$ respect to the data that induced the loss landscape.}
\label{fig:method_illustration}
\vspace{-1em}
\end{figure*}

\textbf{Our approach.}
To detail our approach, consider a simple two-layer fully-connected ReLU network $f_\theta$ trained to classify one-dimensional (1D) training data into two classes as presented in Figure~\ref{fig:method_illustration} (A)-(B). In this 1D input space, the network learns a decision boundary located at two points, $x_L$ and $x_R$ (Figure~\ref{fig:method_illustration} (C.1)). Consider gradients of the loss function of individual data points in the input space as in Figure~\ref{fig:method_illustration} (C.3). Those gradients align with the directions in the parameter space corresponding to the largest increase of error on the data. Overall, the largest error is made on the data that is on the boundary by shifting it across the boundary. 
Moreover, gradients on the opposite sides of the boundary point in opposite directions as moving the boundary benefits samples from one class but hurts samples from another. Indeed, we see that gradients on either side of $x_L$ and $x_R$ point in opposite directions. 
These directions align with the top two Hessian eigenvectors $v_1$ and $v_2$ corresponding to the two outliers in its eigenspectrum (Figure~\ref{fig:method_illustration} (C.2,C.4)). We see that gradients around $x_R$ align perfectly with the top eigenvector $v_1$: The cosine similarity flips from $-1$ to $1$ as the decision boundary is crossed (Figure~\ref{fig:method_illustration} (C.5)). The gradients around $x_L$ align with the second top eigenvector $v_2$. We conclude that the top Hessian eigenvectors encode separate pieces of the decision boundary learned by the network.

In the remainder of this work, we give a formal definition of our framework in Section~\ref{sec:hessian_analysis} along with the mathematical intuition and main observations; propose the generalization measure and margin estimation technique in Section~\ref{sec:gen_and_margin}; validate our results on real data in Section~\ref{sec:real_data}; discuss the implications of our findings in Section~\ref{sec:discussion} and conclude in Section~\ref{s:conclusions}.

\section{Hessian-gradient analysis and the complexity of
 the decision boundary}\label{sec:hessian_analysis}

We consider a $C$ class classification problem with training data $\mathcal{D} = \{x_i,y_i\}_{i=1}^{n}$ where $x_i \in \mathbb{R}^d$ and $y_i \in \{1,\dots, C\}$ is the class label. 
Let the neural network be $f_{\theta}: \mathbb{R}^d \to \mathbb{R}^\mathcal{C}$ parameterized by $\theta \in \mathbb{R}^p$ where we focus on the over-parameterized setting, that is, $p \gg nd$. 
We obtain the class prediction as $\hat{y}_i = \arg \max f_\theta(x_i)$. 
The training of the network $f_\theta$ is done using stochastic gradient descent (SGD) and cross-entropy loss 
$\mathcal{L}(\theta; \mathcal{D}) = \sum_{i=1}^n \sum_{c=1}^{\mathcal{C}} \mathbbm{1}[y_i = c] \log [f_\theta(x_i)]_c
$ where $\mathbbm{1}[ \cdot ]$ is the indicator function.
The pairwise decision boundary of the classifier (between two classes $c$ and $c'$) is defined as a set of points in the input space $\mathcal{B} = \{ z : [f_{\theta}(z)]_c = [f_{\theta}(z)]_{c'} = \mathrm{max}f_{\theta}(z) \}$ which $f$ classifies as being equally likely to belong to class $c$ and $c'$. Within this work, we focus on datasets where we can visualize the decision boundary. Therefore, we follow \cite{Kienitz2023comparingcomplexities} and define the geometric complexity of the decision boundary as the number of its linear segments.

In a similar spirit to \citet{fort2019emerglocalgeom}, we define the \emph{reinforcing gradient} $g_{\theta}: \mathbb{R}^d \to \mathbb{R}^p$ of a given input  $x$ to be the gradient direction in parameter space that strengthens the dominating class in the distribution of the current prediction from $f_\theta$ at an input $x$:
\begin{align}\label{eq:reinf_grads}
   g_\theta(x) = \frac{\partial}{\partial \theta} \mathcal{L}\left(\theta;\{x,\hat{y}\}\right)\,.
\end{align}
Let the Hessian matrix of the training loss on data $\mathcal{D}$ be the square matrix $H \in \mathbb{R}^{p \times p}$ such that 
$H_{i,j} = \frac{\partial^2}{\partial \theta_i \theta_j} \mathcal{L}(\mathcal{D})$.
Each eigenvalue and its corresponding normalized eigenvector of $H$ is denoted by $\lambda_i$ and $v_i \in \mathbb{R}^p, \forall i \in [p]$ respectively. We assume that the ordering is descending in value of the eigenvalues $\lambda_i$.
Therefore, the top $k$ Hessian eigenvectors correspond to the first $k$ largest Hessian eigenvalues.
Then, we define the \emph{alignment} between the reinforcing gradient $g_\theta(x)$ of an input $x$ with eigenvector $v_i$ in terms of the cosine similarity as
\begin{align}\label{eq:alignment}
     \mathcal{A}_i(x) = \frac{\left \langle g_\theta(x) , v_i \right\rangle}{\|g_\theta(x)\| \|v_i\|},
\end{align}
where $\langle\cdot,\cdot\rangle$ is the scalar product, and $\|\cdot\|$ is the Euclidean norm of the vector. Note that this alignment crucially depends on $\theta$ in parameter space from which the Hessian $H$ is derived.
The cosine similarity is more informative than the common scalar product, as discussed in Appendix~\ref{app:dotvscosine}.
\begin{figure*}[t]
    \centering
\includegraphics[width=\textwidth]{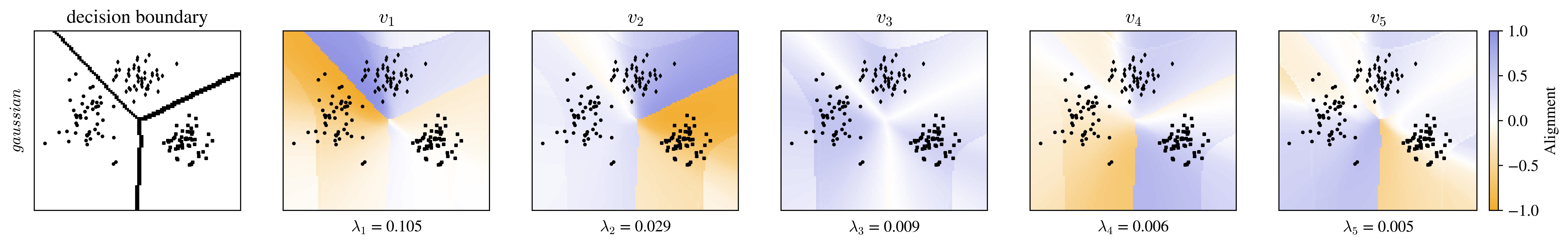}\vspace{-1em}
\caption{\textbf{Experimental results on \emph{gaussian} dataset.} \emph{(First column)} The decision boundary in the data space obtained by training a two-layer fully connected network. \emph{(Other columns)} The alignment of reinforcing gradients with the top five eigenvectors illustrates that the top eigenvectors encode the decision boundary.}\label{fig:gauss_normal_training}\vspace{-1em}
\end{figure*}
%\section{Results}\label{sec:results}
To showcase the connection between the Hessian top eigenvectors and decision boundary as well as the effectiveness of the generalization measure, we conduct a series of experiments on small datasets.

\paragraph{Datasets and architectures.}
We consider five two-dimensional (2D) simulated datasets: \emph{gaussian} with three classes sampled from Gaussian mixtures, concentric \emph{circle} and \emph{half-moon} datasets with two classes each, \emph{hierarchical gaussian} with four classes, and \emph{checkerboard} dataset with two classes. We also validate our findings on real datasets such as \emph{Iris} and \emph{MNIST}. In the main body of the manuscript, we focus on results for \emph{gaussian}, \emph{checkerboard}, and \emph{MNIST-017}. The other results are presented in Appendix~\ref{app:sim_datasets}.
We study both two-layered fully-connected neural networks and convolutional neural networks with a number of model parameters around $10^5$. Detailed descriptions of the datasets and the models are provided in the code.\footnote{The code is available at: \url{https://github.com/Shmoo137/Hessian-and-Decision-Boundary}.}
Since the models and datasets are of tractable sizes, we compute the Hessian exactly using the \texttt{torch.autograd} module from PyTorch \citep{PyTorch}.

We measure the alignment of the training data
%either all training data or an ample number of input points 
with respect to each eigenvector of the Hessian for a converged network $f_{\hat{\theta}}$ where the training loss is converged but not necessarily equal to $0$, and conduct extensive empirical analysis leading to the following results.

\subsection{Top Hessian eigenvectors encode the decision boundary}\label{ss:results-top-heigenvectors}
We plot the alignment of reinforcing gradients and each of the top $k=5$ eigenvectors for the 2D \emph{gaussian} dataset in Figure~\ref{fig:gauss_normal_training}.
For the topmost eigenvectors, \emph{we observe a close-to-one absolute alignment with gradients of loss of the points on the decision boundary learned by the network.} Moreover, for these points, we see a transition from maximal positive to negative cosine similarity values, i.e., a switch of the alignment sign. Those results hold for all other considered 2D datasets (\emph{circle}, \emph{half-moon}, and \emph{hierarchical gaussian}) as presented in Appendix~\ref{app:sim_datasets}.

We interpret this alignment of the topmost eigenvectors with gradients of samples on the boundary in two ways. 
When we shift the model parameters $\theta$ along the direction of the top eigenvector, the points in input space with high alignment to this vector would either reinforce their class by increasing the output corresponding to their class prediction or weaken their class prediction by decreasing the corresponding output. 
Alternatively, we can think of a gradient aligning with the direction in the parameter space corresponding to the direction of the largest error increase on the respective data. On the boundary, the largest error occurs by shifting the boundary; therefore, gradients there align with parameters whose change would shift the decision boundary. In every case, we see that the reinforcing gradients on the boundary align with the top Hessian eigenvectors indicating that they encode the same information as the direction in the parameter space that would shift the boundary.

We can strengthen this observation mathematically by expanding the loss around a minimum $\theta^*$ using second-order Taylor's approximation at $\theta^* + \Delta \theta$ and considering $\Delta \theta = \frac{g_\theta(x)}{|| g_\theta (x)||}$, resulting in
\begin{align}
    \mathcal{L}\left( \mathcal{Y} + \nabla_\theta f \left( \theta^*, \mathcal{X} \right)^T \frac{g_\theta(x)}{|| g_\theta(x) ||} , \mathcal{Y} \right) &= \dfrac{1}{2} \sum_{i=1}^p \lambda_i \mathcal{A}_{i}(x)^2. \label{eq:taylor_l}
\end{align}
From \eqref{eq:taylor_l}, for the loss to have a maximal change, the reinforcing gradient of $x$ should be aligned with the direction of the steepest ascent of $f(\theta^*, \mathcal{X})$. 
This implies that moving data $x$ in the direction of the gradient of $f(\theta^*, \mathcal{X})$ potentially changes the predicted class for $x$, thus increasing the loss.
In other words, the alignment of the reinforcing gradient of $x$ with the function's gradient is high for $x$ near the decision boundary.
From this understanding and the right-hand side of ~\eqref{eq:taylor_l}, we infer that the alignment of $x$ with the top Hessian eigenvectors is larger for $x$ near the boundary than data points farther away, explaining our numerical observation. A detailed analysis is provided in Appendix~\ref{sec:theory}. 

Additionally, we see that \emph{each top eigenvector may capture only a section of the complete boundary.} This information on the sections of decision boundary can be well separated across eigenvectors as in the case of \emph{gaussian} in Figure~\ref{fig:gauss_normal_training}, where the top eigenvector encodes a section of the boundary between one pair of classes, and the second top eigenvector between another pair of classes. 
Furthermore, the alignment does not necessarily switch between extreme values $+1$ and $-1$ across the decision boundary. \emph{The exact extreme values do not seem informative in contrast to the sign switch itself.} However, for the topmost eigenvectors and standard training, the absolute alignment value is usually close to $1$. Finally, the largest alignment across the input space is always for points on the boundary.

In Appendix~\ref{app:other_directions}, we showcase that \emph{the top few eigenvectors are sufficient to encode the entire decision boundary}, and the other directions in parameter space do not exhibit the same property. Interestingly, when we analyze the Hessian with respect to the loss that only considers a specific class $c$, we observe that the top eigenvectors are now restricted to the boundaries that are relevant to deciding the ``all-against-one'' for the selected class $c$ (Appendix~\ref{app:per-class-boundary}).
%the top few eigenvectors are not only sufficient to encode the whole decision boundary but also that other directions in parameter space do not exhibit the same property.
Moreover, in Appendix~\ref{app:ablation}, we show that the connection between the topmost eigenvectors and the decision boundary is invariant to the architecture and loss function. The dependence on the optimizer is more subtle, and we discuss it in more detail in the same appendix.
In Appendix~\ref{app:grad_hessian}, we show that the top eigenvectors of the covariance matrix of training samples' gradients actually encode the same information as the top Hessian eigenvectors at the minimum as observed by \citet{Ghorbani2019} and \citet{fort2019emerglocalgeom}.
Finally, we see that the same connection between the top Hessian eigenvectors and decision boundary is not restricted to a minimum and persists throughout the training (Appendix~\ref{app:dynamics}).

\subsection{A complex boundary is characterized by many eigenvectors}\label{ss:results-complex-many-heigenvectors}
\vspace{-0.25em}
Past works 
%\citep{Sagun2016, Sagun2018, Ghorbani2019, Papyan2019structure, Papyan2020structure} 
indicate that the number of outliers in the Hessian spectrum is roughly equal to the number of classes in the dataset. However, we hypothesize that \emph{the number of outliers depends on the simplicity of the learned decision boundary.} An increased number of eigenvectors, corresponding to an increased number of outliers, is needed to characterize a more complex decision boundary.

\begin{figure*}[t]
    \centering
    \vspace{-1em}
    \includegraphics[width=0.82\textwidth]{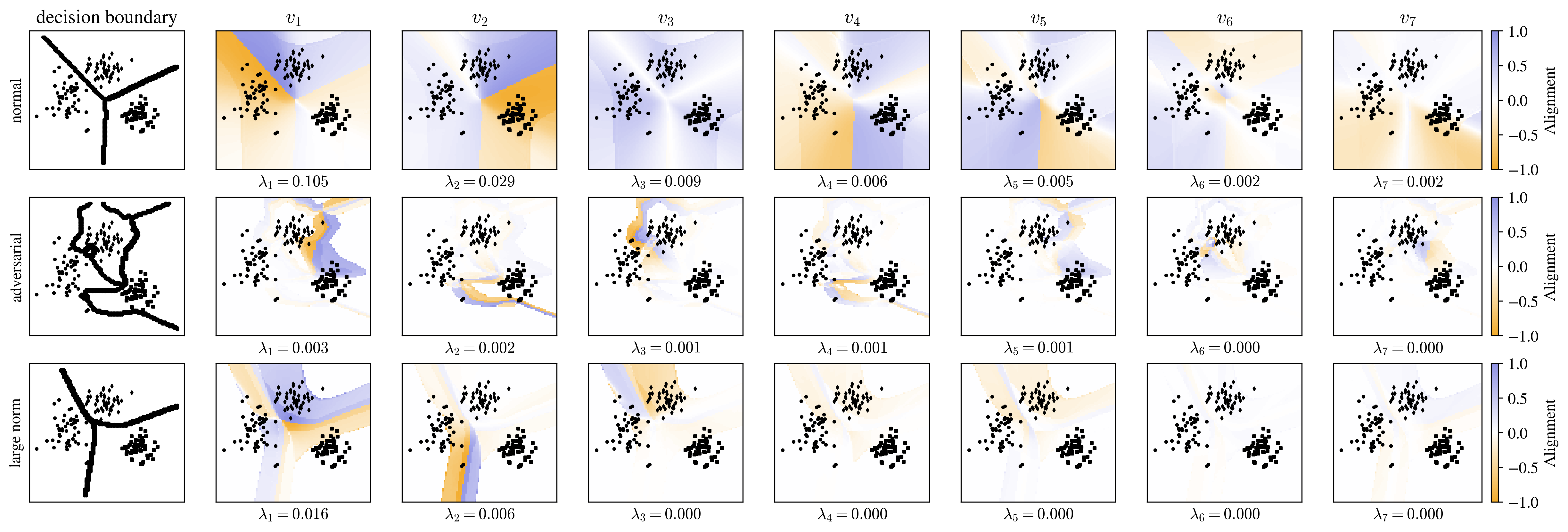}
    \includegraphics[width=0.139\textwidth]
    {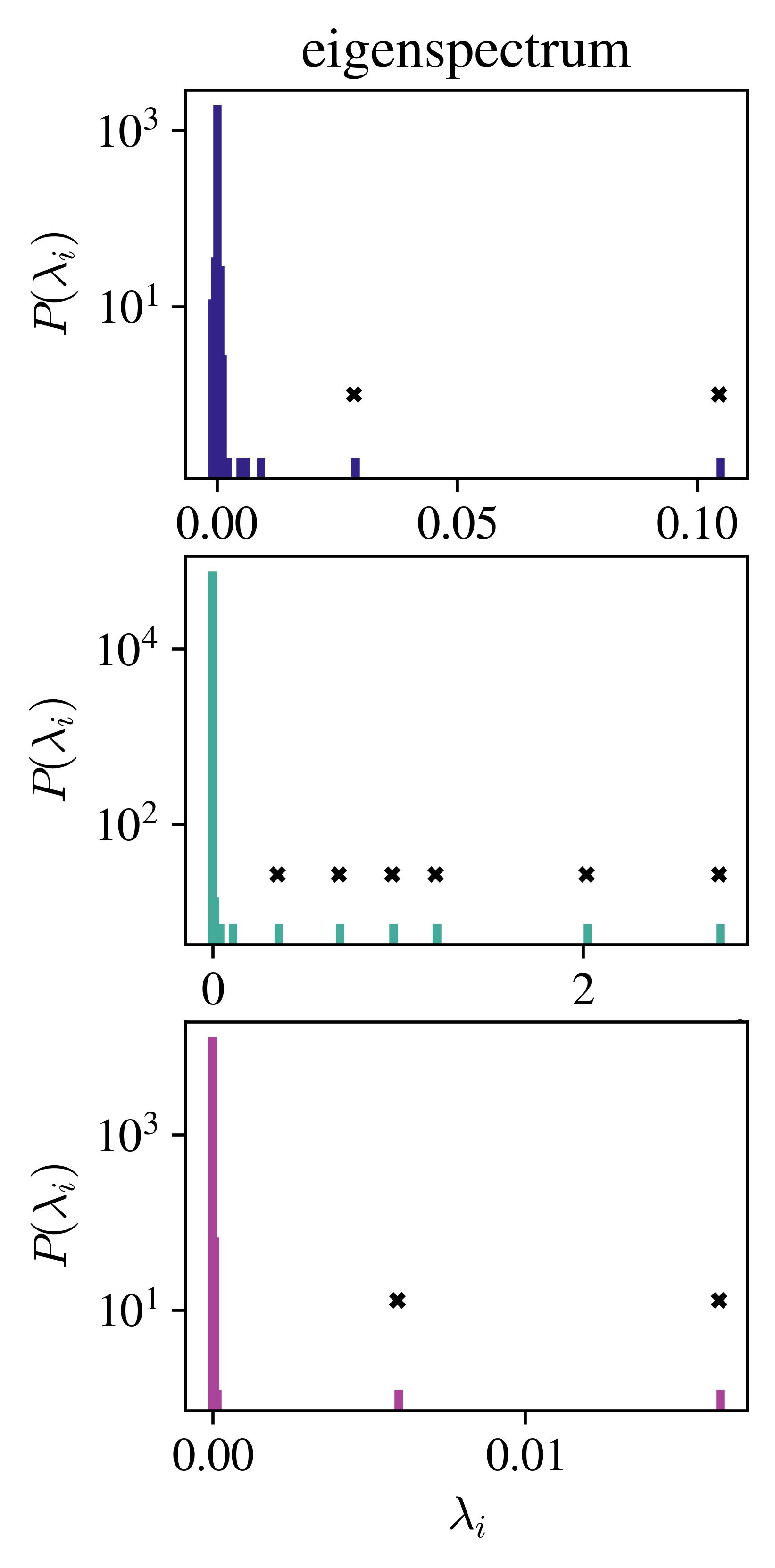}
    %\vspace{-0.5em}
    \caption{\textbf{Decision boundaries of different complexities for \emph{gaussian}.} Alignment plots and histograms of the Hessian spectra for models obtained from normal training, an adversarial initialization \citep{Liu2020badminima}, and a large norm initialization.} %\emph{(Right)} Comparison of the spectrum and the proposed generalization measure for all models.
    %\vspace{-1em}
\label{fig:bad_minima_gaussian}
\end{figure*}

To verify our hypothesis, we follow different training procedures to reach poorly generalizing minima which, by Occam's razor, usually imply complex decision boundaries.
We use two such methods. One is an adversarial initialization as introduced by \citet{Liu2020badminima}. Briefly, the procedure consists in initializing the network with parameters $\theta$ that fit the data with random labels. We notice that such an initialization always exhibits a large $L_2$ norm. Therefore, another method we use consists in simply large norm initialization of the model. Usually, the adversarial and large norm initializations lead to much more complex and slightly more complex decision boundary than the regular initialization, respectively, as presented in the first column of Figure~\ref{fig:bad_minima_gaussian}.
%We hypothesize that tuning a model pretrained on a complex problem for a simple task is analogous to large norm initialization as illustrated in Figure 1 of \citet{Srinivas2022flatten}.
We compute the alignment of reinforcing gradients with the top Hessian eigenvectors corresponding to the outliers for all the initialization methods on \emph{gaussian} as shown in Figure~\ref{fig:bad_minima_gaussian}, which demonstrates \emph{more eigenvectors are needed to describe the learned decision boundary from both adversarial and large norm initializations.}

To complete the picture, 
%Figure~\ref{fig:hist_plot_gaussian} 
the last column of Figure~\ref{fig:bad_minima_gaussian} shows the histogram of Hessian eigenvalues for \emph{gaussian} dataset illustrating that \emph{different training procedures lead to a different number of outliers}.
In particular, normal training leads to $2$ outliers following the conjectures from the past works, whereas the adversarial initialization shows more outliers. It is important to note that the number of the top Hessian eigenvectors that encode sections of the decision boundary does not correspond one-to-one to the number of outliers in the spectra.

\begin{figure}[t]
    \centering
    \includegraphics[width=0.9\textwidth]{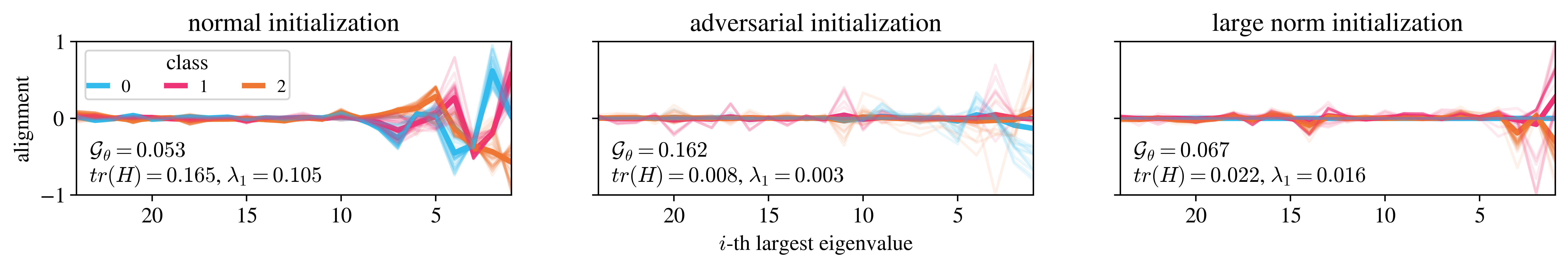}
    %\vspace{-0.75em}
    \caption{\textbf{Alignment of all training data with the top $25$ Hessian eigenvectors for \emph{gaussian} with classes $\{0,1,2\}$ and different initializations.} The dark lines show the mean of each class alignment.\vspace{-0.75em}}
\label{fig:bad_minima_gaussian_only_train}
\end{figure}

So far, we have studied the alignment of reinforcing gradients of samples across the whole input space, which is feasible only for low input space dimensions. In realistic setups, we only have access to training gradients. Therefore, we plot the alignment of gradients of loss of training samples with the top Hessian eigenvectors in Figure~\ref{fig:bad_minima_gaussian_only_train} and confirm that our main observation holds also for this subset of the input space: indeed, a more complex decision boundary leads to a larger number of Hessian eigenvectors with non-zero alignment with training reinforcing gradients. Moreover, for simpler decision boundary (normal initialization) the gradients are much more localized in the Hessian space, that is, the alignment is significantly greater than $0$ only for the top eigenvectors, than in the other initializations. Interestingly, we also see that gradients at the better generalizing minimum are more aligned with one another according to their classes.

%\vspace{-0.25em}
\section{Generalization measure and margin estimation technique}\label{sec:gen_and_margin}
%\vspace{-0.5em}
Going with the conventional wisdom that a simple decision boundary generalizes better than a complex one and the results from Section~\ref{sec:hessian_analysis}, we naturally define a generalization measure $\mathcal{G}_\theta$ that counts the number of eigenvectors needed to describe the decision boundary. 
Mathematically, we define $\mathcal{G}_\theta$ to be the ratio of Hessian eigenvectors with non-zero absolute mean alignment $\mathcal{A}$ with the training samples to the total number of eigenvectors, computed at the minimum ${\theta}$:
\begin{align}
    m_i = \dfrac{1}{n} \sum_{s=1}^{n} |\mathcal{A}_i(x_s)| \quad ; \quad
    \mathcal{G_\theta} = \dfrac{1}{p} \sum_{i=1}^{p} \mathbbm{1}[ m_i > \epsilon ], \label{eq:gen_measure}
\end{align}
where $\epsilon$ is close to zero,\footnote{To be precise, $\epsilon$ is set to a small number being the average maximum alignment with several random directions in a parameter space (see Appendix~\ref{app:other_directions}). Usually, it is around $10^{-2}$.} $|\cdot|$ is the absolute value, and $m_i$ denotes the mean of absolute alignment of the training samples with respect to eigenvector $v_i$. 
A better generalizing minimum has a smaller number of eigenvectors with a non-zero alignment of individual training data gradients, $\mathcal{G_\theta}$, signifying a simpler (therefore, better generalizing) decision boundary compared to other minima of the same network on the same data. In other words, there are fewer directions in the parameter space whose shift corresponds to large errors in the training data. Finally, as $\mathcal{G}_\theta$ depends crucially on training samples and the number of Hessian eigenvectors, the comparison of $\mathcal{G}_\theta$ between minima is meaningful only when they are reached with models with the same architecture and trained on the same data. Note that its value changes between training procedures with fixed hyperparameters due to randomness in the initialization. %(in our experiments, it was never greater than 0.05).

\subsection{Our generalization measure captures the complexity of the decision boundary}

We compute our generalization measure $\mathcal{G}_\theta$ as in \eqref{eq:gen_measure} for all the datasets trained from normal, adversarial, and large norm initializations, leading to decision boundaries of varied complexity and observe that \emph{$\mathcal{G}_\theta$ captures the correct generalization order of the three minima} in all cases as seen in Figure~\ref{fig:bad_minima_gaussian} and Table~\ref{tab:gen_meas_comp_main}.
We also compare the $\mathcal{G}_\theta$ with the standard flatness measures like the Hessian trace and its spectral norm.
We confirm their superfluousness in Table~\ref{tab:gen_meas_comp_main}, where the Hessian trace and spectral norm of the normally initialized network with the simplest decision boundary are larger than for networks with adversarial and large norm initializations. Interestingly, the $L_2$ norm of the parameters also fails as a generalization measure despite the observation that the well-generalizing solutions tend to have a minimum norm \citep{wilson2017marginal}. Those observations hold across the simulated and real datasets as presented in Appendices~\ref{app:generalization_measure_sim} and \ref{app:gen_real}, respectively. 

\begin{table}[h]
\centering
\caption{\textbf{Generalization measure comparison for \emph{gaussian} and \emph{MNIST-017}.} Averaged over $5$ runs.}
\small
\begin{tabular}{@{}llllll@{}}
\toprule
\multicolumn{1}{c}{\textbf{Dataset}} & \multicolumn{1}{c}{\textbf{Training}} & \multicolumn{1}{c}{\textbf{$\mathcal{G}_\theta \downarrow$}} & \multicolumn{1}{c}{\textbf{$\mathrm{trace}(H) \downarrow$}} & \multicolumn{1}{c}{\textbf{$\lambda_{\max}(H) \downarrow$}} & \multicolumn{1}{c}{\textbf{$|| \theta^* ||_2 \downarrow$}} \\ \midrule  
\multirow{3}{*}{\emph{gaussian}}   
& normal      & $ \color{applegreen}{\textbf{0.055 }} \pm 0.004$& $0.176 \pm 0.010$ & $0.114 \pm 0.010$ & $\color{applegreen}{\textbf{19.60}} \pm 0.15$ \\
 & adversarial &$0.156 \pm 0.035$&$\color{brightmaroon}{\textbf{0.003}} \pm 0.001 $&$ \color{brightmaroon}{\textbf{0.002}}\pm 0.001$&$ 105.00 \pm 0.005$             \\
& large norm &$ 0.114 \pm 0.040$&$0.021\pm 0.018$&$ 0.017\pm 0.012$&$ 98.169 \pm 0.413$\\ \midrule 
\multirow{2}{*}{\emph{MNIST-017}}  & normal      & $\color{applegreen}{\textbf{0.037}} \pm 0.028$                & $6.288 \pm 4.697$   & $3.758 \pm 3.215$     & $2237.62 \pm 3526.8$ \\
           & adversarial & $0.109 \pm 0.002$                & $\color{brightmaroon}{\textbf{0.945}} \pm 0.117$   & $\color{brightmaroon}{\textbf{0.479}} \pm 0.097$     & $\color{brightmaroon}{\textbf{938.378}} \pm 0.032 $    \\\bottomrule
\end{tabular}
\label{tab:gen_meas_comp_main}
%\vspace{-1em}
\end{table}

The proposed generalization measure also overcomes another weakness of the standard Hessian measures and is \emph{invariant to the reparametrization} as presented in Appendix \ref{app:reparameterization}, as it relies on the decision boundary complexity which is also invariant to the reparametrization. The $\mathcal{G}_\theta$ has also limitations. As it arises from the alignment of the gradients of loss of training samples with the Hessian eigenvectors encoding various sections of the boundary, it may fail to signal an increased complexity of the decision boundary far from any training sample (see Appendix~\ref{app:generalization_measure_sim} for more details). The second limitation is related to the simplicity bias of neural networks.

\subsection{Shortcomings of our method due to simplicity bias}
\label{ss:results-simplicity-bias}
Simplicity bias is the tendency of neural networks to learn ``simple'' models and has been hypothesized to explain generalization properties of neural networks \citep{memorisation_DN_2017Arpit, incr_complexity_learning2019}. \citet{Shah2022simplbias} claims that simplicity bias may instead hurt generalization (in SGD and its variants), when networks prefer simpler features over complex ones that are more informative for prediction.
%Since our generalization measure $\mathcal{G}_\theta$ relies on the prudence that simple decision boundaries are better than complex ones, it is natural to ask how it is influenced by simplicity bias.
In the context of the decision boundary, the simplicity bias is related to a bias towards a more linear boundary. \emph{As our generalization metric measures the simplicity of the decision boundary, it may fail to signal when the generalization capability of a model is diminished by simplicity bias.}

To validate it, we use the synthetic \emph{checkerboard} dataset of Gaussian clusters, similar to the one analyzed by \citet{Shah2022simplbias}; it has two classes and two features -- one feature is simple (single linear boundary sufficient for $100\%$ accuracy), and the other is complex ($100\%$ prediction needs at least  $n-1$ linear pieces for $n$ clusters). 
%The choice of this dataset is motivated by the ease of visualization and simple interpretation of simplicity bias in terms of a linear boundary, which results in a narrow margin at the expense of a wider one. 
We study the minima reached by models trained to classify this dataset
%compute the Hessian spectrum, alignment, and generalization measure for this dataset trained 
in two settings. The first setting is normal initialization, affected by simplicity bias, resulting in a more linear boundary and a narrow margin. In the second setting, which we call \emph{wide-margin}, we encourage a boundary with a wider margin by pretraining on another dataset as explained in Appendix~\ref{app:sim_datasets}, and then training on the same \textit{checkerboard} dataset as in the first setting.

In Figure~\ref{fig:simplbias_margin}, we show the decision boundary and the alignment of reinforcing gradients in input space with the top three eigenvectors $v_i$. The Hessian eigenspectrum, presented in the fifth column, exhibits two outliers for both initializations.
%(refer $x$-axis labels of Figure~\ref{fig:simplbias_margin}), resulting in training data aligning with $v_1$ and $v_2$, encoding two sections of the boundary. 
While the second setting has a wider margin and thus exhibits better generalization, the difference between $\mathcal{G}_\theta$ for normal and wide-margin initializations is small. To properly distinguish those solutions, we can instead estimate the margin width of the decision boundary as described in the next section.

\subsection{The order of eigenvectors is related to the margin width of the decision boundary}\label{ss:results-order-of-eigenvectors}
We observe in every setup (Figure~\ref{fig:gauss_normal_training} and Appendix~\ref{app:sim_datasets}) that \emph{the order of the top eigenvectors follows the increasing margin of the encoded sections of the boundary}. The topmost eigenvector captures the boundary section that separates the closest training data from two classes in the input space.

\begin{figure}[t]
    \centering
    \vspace{-0.5em}
   \includegraphics[width=0.675\textwidth]{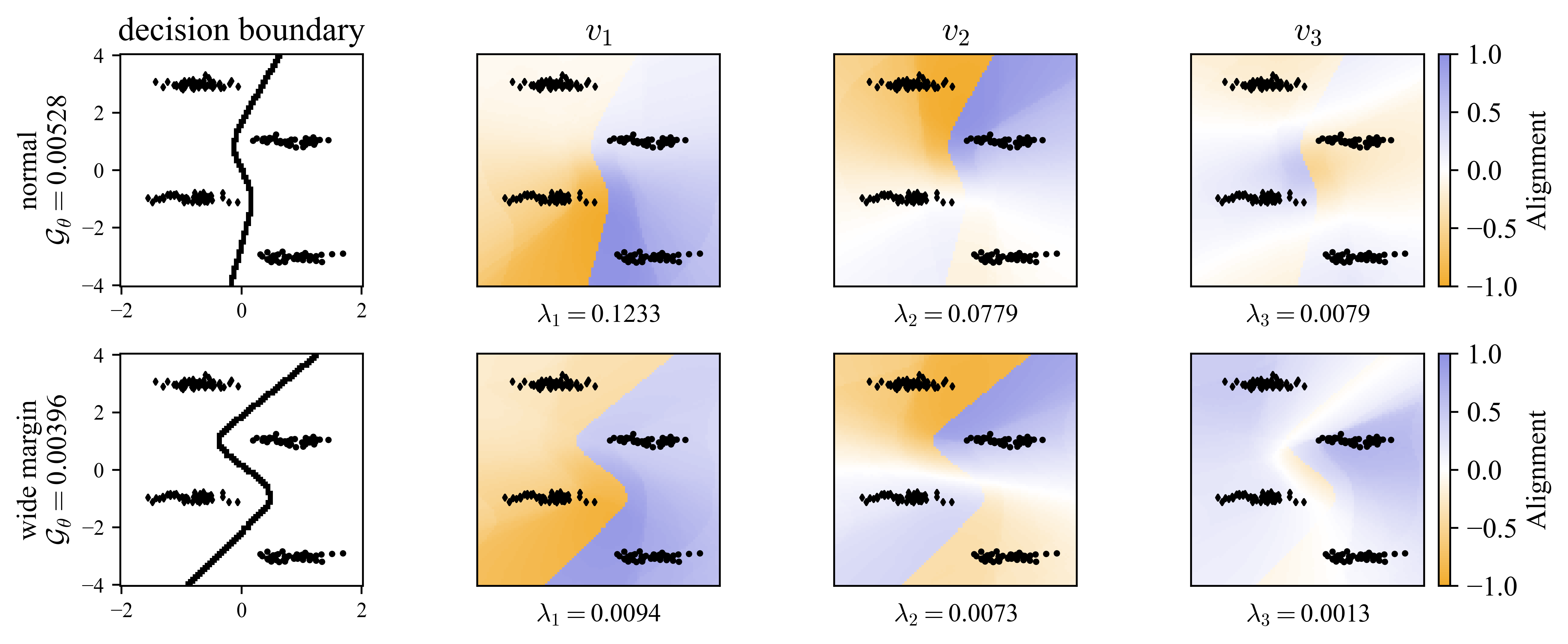}
    \includegraphics[width=0.154\textwidth]{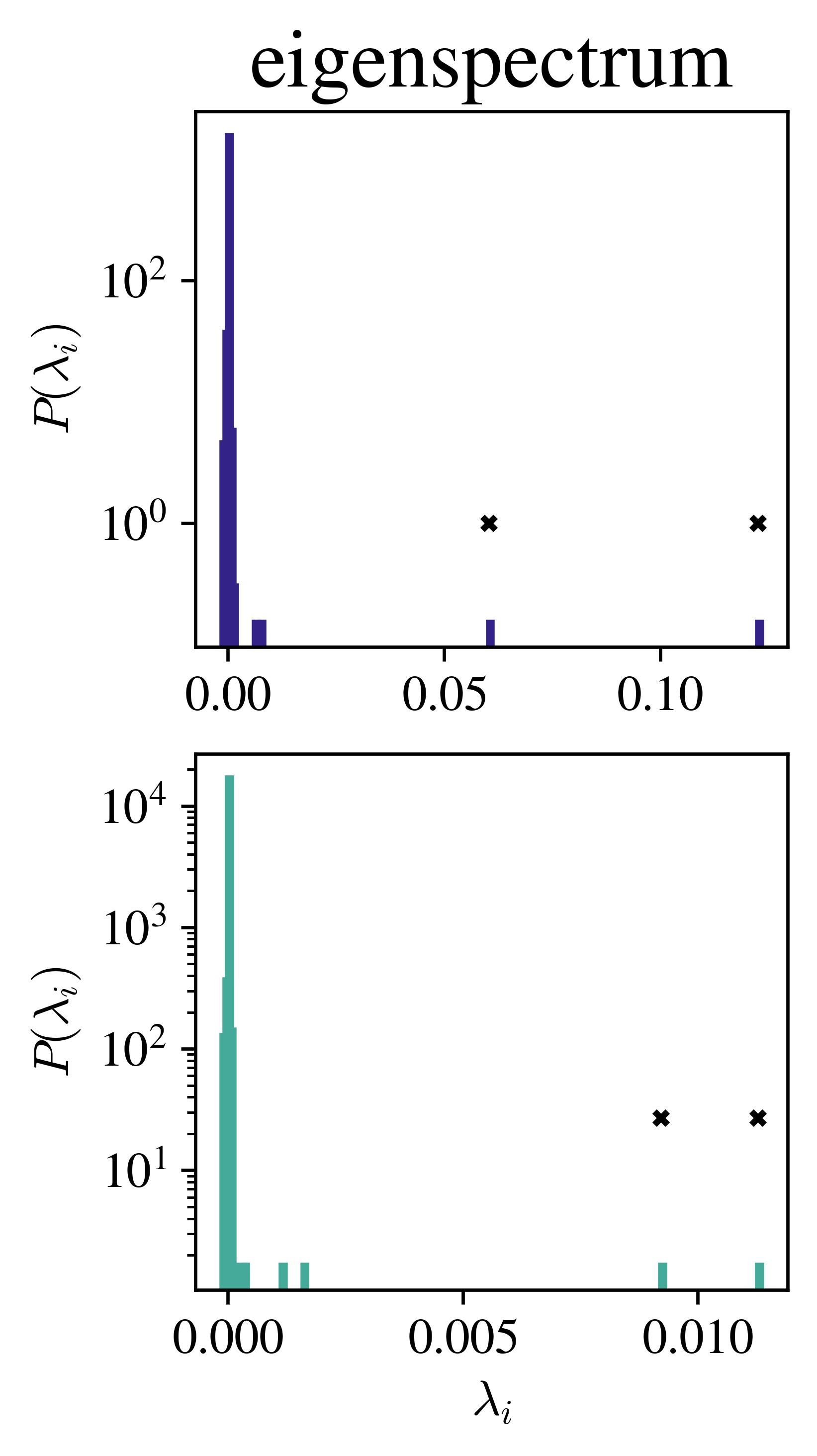}
    \includegraphics[width=0.15\textwidth]{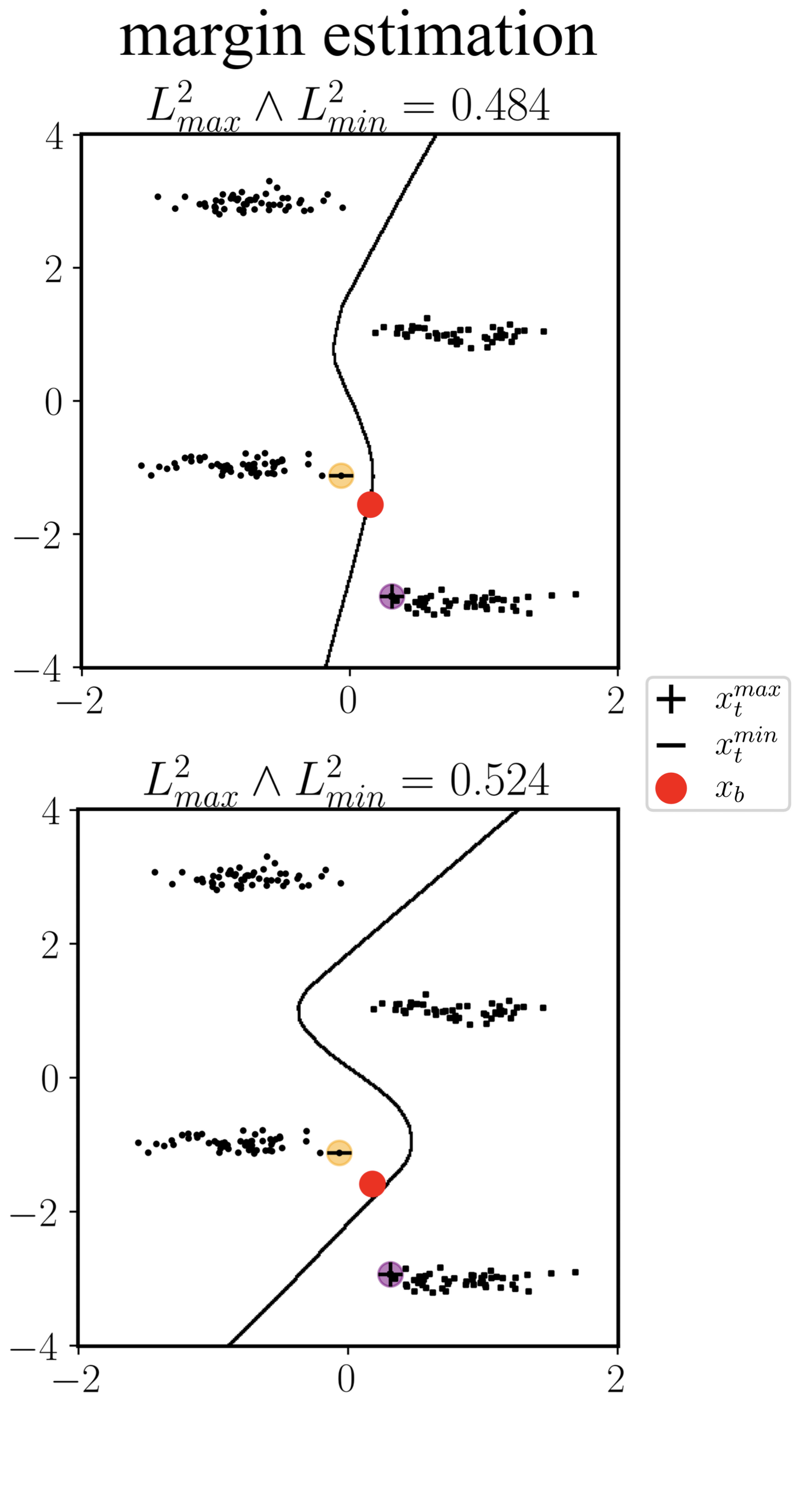}
    \vspace{-0.4em}
    \caption{\textbf{Simplicity bias and margin estimation for \emph{checkerboard}.} We compare the alignments of the top three eigenvectors $v_i$ and the eigenspectrum, trained using two different initializations, normal \emph{(top row)} and  wide-margin initialization \emph{(bottom row)}.
    Both initializations have very similar generalization measures. \emph{(Last column)} Margin estimation from $x_b, x_i^{\rm max}$ and $x_i^{\rm min}$.}
    \label{fig:simplbias_margin}
\end{figure}

This opens up a possibility to \emph{estimate the margin of the decision boundary in higher input dimensions.} To do so, we need two data points from the input space: a training sample $x_t$ that is closest to the boundary (the one that determines the smallest margin of the decision boundary) and a sample on the boundary $x_b$, which should be as close to $x_t$ as possible, as this determines how good estimate of the margin we have. $x_t$ is chosen to have the largest alignment with the top Hessian eigenvector $v_1$. Note that, a priori, we do not know the alignment sign of the training data $x_t$ closest to the artificial sample on the boundary $x_b$. Hence, $x_t^{\text{min}}$ and $x_t^{\text{max}}$ are the smallest and largest alignments with $v_1$ (yellow and purple dots in the last column of Figure~\ref{fig:simplbias_margin}).
To find the sample on the boundary (red dots in Figure~\ref{fig:simplbias_margin}), we can optimize the features of an input sample such that its gradient has a maximum alignment with the top Hessian eigenvector. Then we compute the $L_2$ distance between $x_b$ and $x_t^{\text{min}}$, and between  $x_b$ and $x_t^{\text{max}}$ in the input space and choose the smaller $L_2$ distance as the margin. An example of such a margin width estimation is presented in Figure~\ref{fig:simplbias_margin}, where we correctly see that the less linear decision boundary corresponds to a wider margin compared to the linear decision boundary. 
This simple yet effective margin estimation technique, together with our generalization measure $\mathcal{G}_\theta$, enables a better understanding of the generalization ability of deep neural networks.

\subsection{Validation of our results on real data}
\label{sec:real_data}

We focused our analysis so far on low-dimensional datasets which allow to visualize the decision boundary and avoid relying on a proxy for its complexity. We now extend a part of our analysis to more realistic datasets by studying minima obtained within two settings, that is, normal and adversarial initialization. We obtain the results for \emph{Iris} and four subsets of \emph{MNIST}: \emph{MNIST-017}, \emph{MNIST-179}, \emph{MNIST-0179}, and \emph{MNIST-1379}, where numbers indicate selected classes of digits. Here we focus on results for \emph{MNIST-017}, yet the full analysis is in Appendices \ref{app:Iris}-\ref{app:gen_real}.

\begin{figure}[t]
    \centering
    \includegraphics[width=\textwidth]{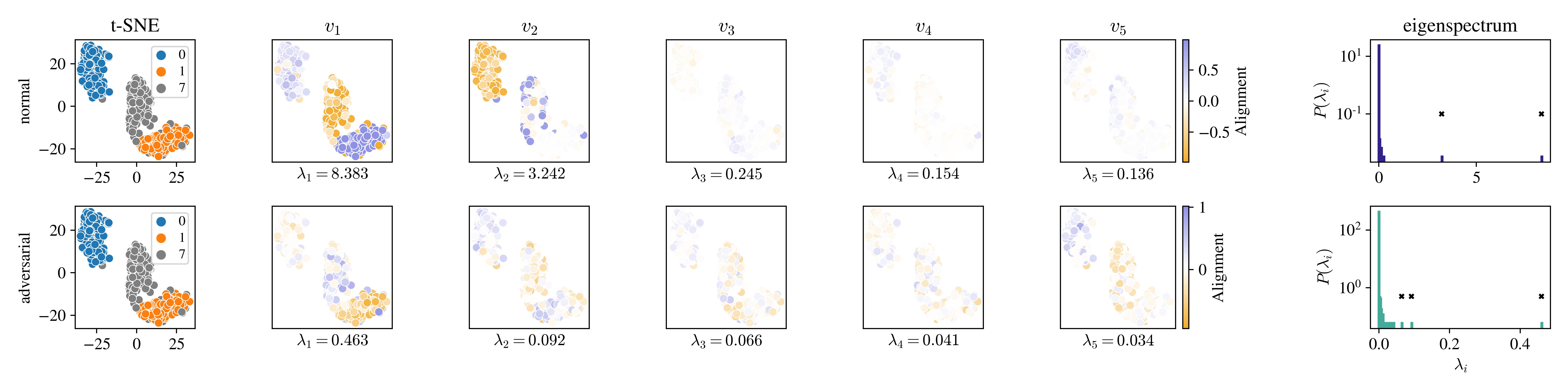}
    \vspace{-2em}\caption{\textbf{Normal and adversarial initialization training for \emph{MNIST-017} with t-SNE visualization.} We visualize the \emph{MNIST-017} dataset with t-SNE and color code the alignments of gradients of all training samples onto the top $5$ eigenvectors. \emph{(Last column)} The eigenspectrum with the outliers.
    \vspace{-1em}}
\label{fig:bad_minima_mnist}
\end{figure}

In particular, we see that \emph{the generalization measure $\mathcal{G}_\theta$ precisely distinguishes models trained with different initializations} having different generalization abilities (Table~\ref{tab:gen_meas_comp_main}). Moreover, we visualize the dataset using t-SNE in Figure~\ref{fig:bad_minima_mnist} and color code the alignment between the training gradients and top Hessian eigenvectors. While t-SNE does not necessarily have the same data representation as a trained neural network, the alignment behavior for models obtained with the normal and adversarial initialization training seems to follow the one for the 2D datasets. It suggests that \emph{the top eigenvectors also encode the decision boundary}. Finally, we also see \emph{a higher number of outliers in the eigenspectrum for complex boundary} (last column of Figure~\ref{fig:bad_minima_mnist}).

\section{Discussion} \label{sec:discussion}

\textbf{Properties of the Hessian.} 
Our work sheds light on various universal aspects of the training loss Hessian. 
%Particularly, we show that the top Hessian eigenvectors encode sections of the decision boundary.
While the localization of the gradient information in the top subspace of the Hessian is known \citep{GurAri2019}, the observation that the top Hessian eigenvectors align with the gradients of samples at the decision boundary, therefore, they encode the decision boundary, provides a new perspective and a simpler way to study the complexity of high-dimensional decision boundary.
In fact, our work aligns with the mathematical analysis of \citet{Papyan2020structure} who connected the emergence of outliers (and therefore the top Hessian eigenvectors) to the ``between-class gradient second moment''.
Moreover, our work illustrates that the number of outliers in the Hessian spectrum is related to the complexity of the learned decision boundary, which was so far rather connected to the number of classes. % \citep{Sagun2016, Sagun2018, Ghorbani2019, Papyan2019structure, Papyan2020structure}. 
%Particularly, we see the outliers emerge with the learned sections of the decision boundary whose number in the normal training setting is naturally connected to the number of classes. When the complexity of the learned decision boundary grows (e.g., via adversarial or large norm initialization), number of outliers also increases. 
Our findings shed light on the observation of \citet{jastrzebski2019relation} that the loss in the subspace of the top Hessian eigenvectors is ``bowl-like'' and that decreasing learning rate within this subspace leads to better generalizing solutions:
this procedure may lead to maximizing the margin of the decision boundary.

\textbf{Generalization.}
We confirm that the Hessian-based metrics like the trace or the largest eigenvalue %\citep{Keskar2017, Wu2017, Izmailov2018, He2019} 
are unreliable proxies for measuring the generalization ability of deep learning models. 
%\citep{Dinh2017, Zhang2021rethinking, Chatterjee2022coherentgrads, andriushchenko2023modern} 
Instead, these metrics are more dependent on the training parameters like the number of epochs, as observed in \citet{Sagun2016}.
%correlated with the training hyperparameters like the number of epochs.
For example, Table~\ref{tab:gen_meas_comp_main} shows that the Hessian trace is the smallest for badly generalizing minima, which required the largest number of epochs to train.
Interestingly, we see that within-class gradients at the minimum are more aligned with each other in the better generalizing case, as noted in \citet{Chatterjee2022coherentgrads}. Our findings suggest that this gradient ``coherence'' emerges from the simplicity of the decision boundary and disappears as the complexity increases.

\textbf{Robustness and interpretability.}
The robustness of neural networks is an active area of research aiming to produce stable outputs towards small, semantically irrelevant input perturbations. \citet{feng2022activityweight} mapped the variations in inputs to variations of specific weight parameters, through which the Hessian connection between the input and parameter space is established.
%the corresponding dual weights, connecting the Hessians in the input and parameter space. 
Combining this with our work may lead to a better understanding of why Hessian-based techniques could lead to more robust models \citep{MoosaviDezfooli2019, Qin2019adversarial, Zhang2019TRADES, Srinivas2022flatten} or more robust gradient-based explanations \citet{Dombrowski2019, Dombrowski2022}. In fact, they may simplify the decision boundary, causing gradients of similar inputs to align in one direction. This also explain the success of Hessian-based interpretation of neural networks \citep{Koh17influence, Madras2020LE, Dawid2021Hessian} and in pruning \citep{LeCun1989, Yu2021}.

\textbf{Practical implications.}
Establishing the connection between Hessian and the decision boundary learned by the network provides a new tool to study the boundary in high input dimensions. However, exact computation of the Hessian and its spectrum is hard for both large data and models. Instead, we can follow the observation from Appendix~\ref{app:grad_hessian} and \citet{Ghorbani2019} that the top eigenvectors of the Hessian at the minimum have a large overlap with the top eigenvectors of the gradient covariance matrix, which enables efficient alignment computation but still requires an expensive eigendecomposition. We can also use efficient approximation techniques based on the Hessian-vector product \citep{Pearlmutter94, Agarwal2017, hessianEigenthingsRepo} and the generalized Gauss-Newton decomposition of the Hessian \citep{Sagun2016, Papyan2019structure, Papyan2020structure}.

\section{Conclusion}\label{s:conclusions}
%\vspace{-1em}
%answer the following questions, \emph{is there a connection between the Hessian and the decision boundary? What is the significance of the outliers in the Hessian spectrum, and is it always approximately equal to the number of classes? Would answers to these questions help us understand generalization better through the complexity of the decision boundary?} 
%inspect the spectrum of the Hessian to establish a connection between the Hessian eigenvectors and the learned decision boundary, which we understand as a hypersurface to separate data from different classes in input space. 
In this work, we establish that the top Hessian eigenvectors characterize the decision boundary learned by the neural networks. With this understanding, we propose a generalization measure that shows promising empirical results in capturing the generalization of neural networks as it aims to quantify the complexity of the decision boundary. 
While this is the strength of our approach, it can also lead to overlooking
%it is also the reason for failing 
the simplicity bias when it hurts generalization, as discussed in Section~\ref{ss:results-simplicity-bias}.
In such a case, we propose a novel technique for estimating the margin of the decision boundary using the alignment with the top Hessian eigenvector.
Naturally, one can also study the per-class margin through the boundary encompassing each class (Appendix~\ref{app:per-class-boundary}). 
While we show the invariance of our main result to architectures and choice of the loss function in Appendix~\ref{app:ablation} and confirm the observations for real datasets in Section~\ref{sec:real_data} and Appendices~\ref{app:Iris}-\ref{app:gen_real}, we acknowledge the need to expand the study to more diverse datasets which is left for future work.

Our work opens various avenues for further research. For instance, the meaning of the exact values of the alignment and their connection to the margin (if any) are elusive. Moreover, a trace of the Hessian sometimes correlates with a good generalization, which begs the question of the meaning of eigenvalues besides their respective order. We also see that eigenvectors encoding the simple boundary tend to be much sparser than the ones encoding the complex boundary. Determining if the alignment behavior of memorized samples is significantly different than the learned examples could shed light on understanding memorization and when it happens. Moreover, we observe differences in the alignment behavior between the optimizers, which may hint at different properties of minima that those optimizers lead to. To answer any of these questions, the key is to rigorously understand the connection between the top Hessian eigenvectors and the decision boundary in simplified models. Our results are also useful for improving pruning techniques and fighting catastrophic forgetting, e.g., by freezing parameters corresponding to the decision boundary learned so far.
The Hessian-based access to the decision boundary can also open new directions in assessing the uncertainty of deep models' predictions. For example, uncertainty could be connected to the distance of the test sample to the closest boundary. Finally, provided efficient computation, our generalization measure and margin estimation technique could be used during training to promote better generalizing minima.

\begin{ack}
We thank Tony Bonnaire, Francesa Mignacco, Stefani Karp, Yatin Dandi, Artem Vysogorets, Boris Hanin, and Julia Kempe for useful discussions. This work started as an open problem posed by A.D. during the 2022 Les Houches Summer School on Statistical Physics and Machine Learning organized by Lenka Zdeborová and Florent Krzakala.

M.S. is supported by the German Research Foundation (Research Training Group GRK 2428).
A.D. acknowledges the financial support from the Foundation for the Polish Science. The Flatiron Institute is a division of the Simons Foundation.
\end{ack}

%\newpage
\bibliographystyle{neurips-23/bibstyle}
\bibliography{neurips-23/citations}

\begin{thebibliography}{49}
\providecommand{\natexlab}[1]{#1}
\providecommand{\url}[1]{\texttt{#1}}
\expandafter\ifx\csname urlstyle\endcsname\relax
  \providecommand{\doi}[1]{doi: #1}\else
  \providecommand{\doi}{doi: \begingroup \urlstyle{rm}\Url}\fi

\bibitem[Agarwal et~al.(2017)Agarwal, Bullins, and Hazan]{Agarwal2017}
Agarwal, N., Bullins, B., and Hazan, E.
\newblock {Second-order stochastic optimization for machine learning in linear
  time}.
\newblock \emph{J. Mach. Learn. Res.}, 18:\penalty0 1--40, 2017.
\newblock ISSN 15337928.

\bibitem[Alain et~al.(2018)Alain, Roux, and Manzagol]{Alain2019}
Alain, G., Roux, N.~L., and Manzagol, P.-A.
\newblock Negative eigenvalues of the {Hessian} in deep neural networks, 2018.
\newblock URL \url{https://openreview.net/forum?id=S1iiddyDG}.

\bibitem[Andriushchenko et~al.(2023)Andriushchenko, Croce, M{\"u}ller, Hein,
  and Flammarion]{andriushchenko2023modern}
Andriushchenko, M., Croce, F., M{\"u}ller, M., Hein, M., and Flammarion, N.
\newblock A modern look at the relationship between sharpness and
  generalization, 2023.
\newblock URL \url{https://arxiv.org/abs/2302.07011}.

\bibitem[Arpit et~al.(2017)Arpit, Jastrzebski, Ballas, Krueger, Bengio, Kanwal,
  Maharaj, Fischer, Courville, Bengio, and
  Lacoste-Julien]{memorisation_DN_2017Arpit}
Arpit, D., Jastrzebski, S., Ballas, N., Krueger, D., Bengio, E., Kanwal, M.~S.,
  Maharaj, T., Fischer, A., Courville, A., Bengio, Y., and Lacoste-Julien, S.
\newblock A closer look at memorization in deep networks.
\newblock In Precup, D. and Teh, Y.~W. (eds.), \emph{Proceedings of the 34th
  International Conference on Machine Learning}, volume~70 of \emph{Proceedings
  of Machine Learning Research}, pp.\  233--242. PMLR, 06--11 Aug 2017.
\newblock URL \url{https://proceedings.mlr.press/v70/arpit17a.html}.

\bibitem[Auer et~al.(1995)Auer, Herbster, and Warmuth]{auer1995exponentially}
Auer, P., Herbster, M., and Warmuth, M. K.~K.
\newblock Exponentially many local minima for single neurons.
\newblock In Touretzky, D., Mozer, M., and Hasselmo, M. (eds.), \emph{Advances
  in Neural Information Processing Systems}, volume~8. MIT Press, 1995.
\newblock URL
  \url{https://proceedings.neurips.cc/paper/1995/file/3806734b256c27e41ec2c6bffa26d9e7-Paper.pdf}.

\bibitem[Chatterjee \& Zielinski(2022)Chatterjee and
  Zielinski]{Chatterjee2022coherentgrads}
Chatterjee, S. and Zielinski, P.
\newblock On the generalization mystery in deep learning, 2022.
\newblock URL \url{https://arxiv.org/abs/2203.10036}.

\bibitem[Choromanska et~al.(2015)Choromanska, Henaff, Mathieu, Ben~Arous, and
  LeCun]{Choromanska2015}
Choromanska, A., Henaff, M., Mathieu, M., Ben~Arous, G., and LeCun, Y.
\newblock {The loss surfaces of multilayer networks}.
\newblock In \emph{AISTATS 2015 - 18th Int. Conf. Artif. Intell. Stat.},
  volume~38, pp.\  192--204. PMLR, 2015.
\newblock URL \url{http://proceedings.mlr.press/v38/choromanska15.html}.

\bibitem[Dauphin et~al.(2014)Dauphin, Pascanu, Gulcehre, Cho, Ganguli, and
  Bengio]{Dauphin2014}
Dauphin, Y.~N., Pascanu, R., Gulcehre, C., Cho, K., Ganguli, S., and Bengio, Y.
\newblock {Identifying and attacking the saddle point problem in
  high-dimensional non-convex optimization}.
\newblock In \emph{NIPS 2014 - Adv. Neural. Inf. Process. Syst.}, volume~27,
  pp.\  2933--2941, 2014.
\newblock URL
  \url{https://papers.nips.cc/paper/2014/hash/17e23e50bedc63b4095e3d8204ce063b-Abstract.html}.

\bibitem[Dawid et~al.(2022)Dawid, Huembeli, Tomza, Lewenstein, and
  Dauphin]{Dawid2021Hessian}
Dawid, A., Huembeli, P., Tomza, M., Lewenstein, M., and Dauphin, A.
\newblock Hessian-based toolbox for reliable and interpretable machine learning
  in physics.
\newblock \emph{Mach. Learn.: Sci. Technol.}, 3:\penalty0 015002, 2022.
\newblock \doi{10.1088/2632-2153/ac338d}.
\newblock URL \url{https://doi.org/10.1088/2632-2153/ac338d}.

\bibitem[Dinh et~al.(2017)Dinh, Pascanu, Bengio, and Bengio]{Dinh2017}
Dinh, L., Pascanu, R., Bengio, S., and Bengio, Y.
\newblock Sharp minima can generalize for deep nets.
\newblock In Precup, D. and Teh, Y.~W. (eds.), \emph{Proceedings of the 34th
  International Conference on Machine Learning}, volume~70 of \emph{Proceedings
  of Machine Learning Research}, pp.\  1019--1028. PMLR, 06--11 Aug 2017.

\bibitem[Dombrowski et~al.(2019)Dombrowski, Alber, Anders, Ackermann,
  M\"{u}ller, and Kessel]{Dombrowski2019}
Dombrowski, A.-K., Alber, M., Anders, C., Ackermann, M., M\"{u}ller, K.-R., and
  Kessel, P.
\newblock Explanations can be manipulated and geometry is to blame.
\newblock In Wallach, H., Larochelle, H., Beygelzimer, A., d\textquotesingle
  Alch\'{e}-Buc, F., Fox, E., and Garnett, R. (eds.), \emph{Advances in Neural
  Information Processing Systems}, volume~32. Curran Associates, Inc., 2019.
\newblock URL
  \url{https://proceedings.neurips.cc/paper/2019/file/bb836c01cdc9120a9c984c525e4b1a4a-Paper.pdf}.

\bibitem[Dombrowski et~al.(2022)Dombrowski, Anders, Müller, and
  Kessel]{Dombrowski2022}
Dombrowski, A.-K., Anders, C.~J., Müller, K.-R., and Kessel, P.
\newblock Towards robust explanations for deep neural networks.
\newblock \emph{Pattern Recognition}, 121:\penalty0 108194, 2022.
\newblock ISSN 0031-3203.
\newblock \doi{https://doi.org/10.1016/j.patcog.2021.108194}.
\newblock URL
  \url{https://www.sciencedirect.com/science/article/pii/S0031320321003769}.

\bibitem[Fawzi et~al.(2018)Fawzi, Moosavi-Dezfooli, Frossard, and
  Soatto]{Fawzi2018topology}
Fawzi, A., Moosavi-Dezfooli, S.-M., Frossard, P., and Soatto, S.
\newblock Empirical study of the topology and geometry of deep networks.
\newblock In \emph{2018 IEEE/CVF Conference on Computer Vision and Pattern
  Recognition}, pp.\  3762--3770, 2018.
\newblock \doi{10.1109/CVPR.2018.00396}.

\bibitem[Feng \& Tu(2022)Feng and Tu]{feng2022activityweight}
Feng, Y. and Tu, Y.
\newblock The activity-weight duality in feed forward neural networks: The
  geometric determinants of generalization, 2022.

\bibitem[Fort \& Ganguli(2019)Fort and Ganguli]{fort2019emerglocalgeom}
Fort, S. and Ganguli, S.
\newblock Emergent properties of the local geometry of neural loss landscapes.
\newblock \emph{CoRR}, abs/1910.05929, 2019.
\newblock URL \url{http://arxiv.org/abs/1910.05929}.

\bibitem[Ghorbani et~al.(2019)Ghorbani, Krishnan, and Xiao]{Ghorbani2019}
Ghorbani, B., Krishnan, S., and Xiao, Y.
\newblock {An investigation into neural net optimization via Hessian eigenvalue
  density}.
\newblock In \emph{ICML 2019 - 36th Int. Conf. Mach. Learn.}, volume~97, pp.\
  2232--2241. PMLR, 2019.
\newblock ISBN 9781510886988.
\newblock URL \url{http://proceedings.mlr.press/v97/ghorbani19b.html}.

\bibitem[Golmant et~al.(2018)Golmant, Yao, Gholami, Mahoney, and
  Gonzalez]{hessianEigenthingsRepo}
Golmant, N., Yao, Z., Gholami, A., Mahoney, M., and Gonzalez, J.
\newblock Efficient {PyTorch Hessian} eigendecomposition, October 2018.
\newblock URL \url{https://github.com/noahgolmant/pytorch-hessian-eigenthings}.

\bibitem[Guan \& Loew(2020)Guan and Loew]{guan2020analysis}
Guan, S. and Loew, M.
\newblock Analysis of generalizability of deep neural networks based on the
  complexity of decision boundary.
\newblock In \emph{2020 19th IEEE International Conference on Machine Learning
  and Applications (ICMLA)}, pp.\  101--106. IEEE, 2020.

\bibitem[Gur-Ari et~al.(2019)Gur-Ari, Roberts, and Dyer]{GurAri2019}
Gur-Ari, G., Roberts, D.~A., and Dyer, E.
\newblock Gradient descent happens in a tiny subspace, 2019.
\newblock URL \url{https://openreview.net/forum?id=ByeTHsAqtX}.

\bibitem[He et~al.(2019)He, Huang, and Yuan]{He2019}
He, H., Huang, G., and Yuan, Y.
\newblock Asymmetric valleys: Beyond sharp and flat local minima.
\newblock In Wallach, H., Larochelle, H., Beygelzimer, A., d\textquotesingle
  Alch\'{e}-Buc, F., Fox, E., and Garnett, R. (eds.), \emph{Advances in Neural
  Information Processing Systems}, volume~32. Curran Associates, Inc., 2019.
\newblock URL
  \url{https://proceedings.neurips.cc/paper/2019/file/01d8bae291b1e4724443375634ccfa0e-Paper.pdf}.

\bibitem[Hochreiter \& Schmidhuber(1997)Hochreiter and
  Schmidhuber]{hochreiter1997flat}
Hochreiter, S. and Schmidhuber, J.
\newblock Flat minima.
\newblock \emph{Neural Computation}, 9\penalty0 (1):\penalty0 1–42, Jan 1997.
\newblock ISSN 0899-7667.
\newblock \doi{10.1162/neco.1997.9.1.1}.
\newblock URL \url{https://doi.org/10.1162/neco.1997.9.1.1}.

\bibitem[Izmailov et~al.(2018)Izmailov, Podoprikhin, Garipov, Vetrov, and
  Wilson]{Izmailov2018}
Izmailov, P., Podoprikhin, D., Garipov, T., Vetrov, D., and Wilson, A.~G.
\newblock {Averaging weights leads to wider optima and better generalization}.
\newblock In \emph{UAI 2018 - 34th Conf. Uncertain. Artif. Intell.}, volume~2,
  pp.\  876--885, 2018.
\newblock ISBN 9781510871601.
\newblock URL \url{https://arxiv.org/abs/1803.05407v3}.

\bibitem[Jastrzebski et~al.(2019)Jastrzebski, Kenton, Ballas, Fischer, Bengio,
  and Storkey]{jastrzebski2019relation}
Jastrzebski, S., Kenton, Z., Ballas, N., Fischer, A., Bengio, Y., and Storkey,
  A.
\newblock On the relation between the sharpest directions of {DNN} loss and the
  {SGD} step length.
\newblock In \emph{International Conference on Learning Representations}, 2019.
\newblock URL \url{https://openreview.net/forum?id=SkgEaj05t7}.

\bibitem[Karimi \& Derr(2022)Karimi and Derr]{Karimi2022DeepDIG}
Karimi, H. and Derr, T.
\newblock Decision boundaries of deep neural networks.
\newblock In \emph{2022 21st IEEE International Conference on Machine Learning
  and Applications (ICMLA)}, pp.\  1085--1092, 2022.
\newblock \doi{10.1109/ICMLA55696.2022.00179}.

\bibitem[Keskar et~al.(2017)Keskar, Mudigere, Nocedal, Smelyanskiy, and
  Tang]{Keskar2017}
Keskar, N.~S., Mudigere, D., Nocedal, J., Smelyanskiy, M., and Tang, P. T.~P.
\newblock On large-batch training for deep learning: Generalization gap and
  sharp minima.
\newblock In \emph{International Conference on Learning Representations}, 2017.
\newblock URL \url{https://openreview.net/forum?id=H1oyRlYgg}.

\bibitem[Kienitz et~al.(2023)Kienitz, Komendantskaya, and
  Lones]{Kienitz2023comparingcomplexities}
Kienitz, D., Komendantskaya, E., and Lones, M.
\newblock Comparing complexities of decision boundaries for robust training:
  A universal approach.
\newblock In Wang, L., Gall, J., Chin, T.-J., Sato, I., and Chellappa, R.
  (eds.), \emph{Computer Vision -- ACCV 2022}, pp.\  627--645, Cham, 2023.
  Springer Nature Switzerland.
\newblock ISBN 978-3-031-26351-4.

\bibitem[Koh \& Liang(2017)Koh and Liang]{Koh17influence}
Koh, P.~W. and Liang, P.
\newblock Understanding black-box predictions via influence functions.
\newblock In \emph{{ICML 2017 - 34th Int. Conf. Mach. Learn.}}, volume~70, pp.\
   1885--1894. PMLR, 2017.
\newblock URL \url{http://proceedings.mlr.press/v70/koh17a.html}.

\bibitem[Kwon et~al.(2021)Kwon, Kim, Park, and Choi]{kwon2021asam}
Kwon, J., Kim, J., Park, H., and Choi, I.~K.
\newblock Asam: Adaptive sharpness-aware minimization for scale-invariant
  learning of deep neural networks.
\newblock In \emph{International Conference on Machine Learning}, pp.\
  5905--5914. PMLR, 2021.

\bibitem[LeCun et~al.(1989)LeCun, Denker, and Solla]{LeCun1989}
LeCun, Y., Denker, J., and Solla, S.
\newblock Optimal brain damage.
\newblock In Touretzky, D. (ed.), \emph{Advances in Neural Information
  Processing Systems}, volume~2. Morgan-Kaufmann, 1989.
\newblock URL
  \url{https://proceedings.neurips.cc/paper/1989/file/6c9882bbac1c7093bd25041881277658-Paper.pdf}.

\bibitem[Liu et~al.(2020)Liu, Papailiopoulos, and Achlioptas]{Liu2020badminima}
Liu, S., Papailiopoulos, D., and Achlioptas, D.
\newblock Bad global minima exist and {SGD} can reach them.
\newblock In Larochelle, H., Ranzato, M., Hadsell, R., Balcan, M., and Lin, H.
  (eds.), \emph{Advances in Neural Information Processing Systems}, volume~33,
  pp.\  8543--8552. Curran Associates, Inc., 2020.
\newblock URL
  \url{https://proceedings.neurips.cc/paper/2020/file/618491e20a9b686b79e158c293ab4f91-Paper.pdf}.

\bibitem[Madras et~al.(2020)Madras, Atwood, and D'Amour]{Madras2020LE}
Madras, D., Atwood, J., and D'Amour, A.
\newblock {Detecting extrapolation with local ensembles}.
\newblock In \emph{ICLR 2020 - Int. Conf. Learn. Represent.}, 2020.
\newblock URL \url{https://openreview.net/forum?id=BJl6bANtwH}.

\bibitem[Moosavi-Dezfooli et~al.(2019)Moosavi-Dezfooli, Fawzi, Uesato, and
  Frossard]{MoosaviDezfooli2019}
Moosavi-Dezfooli, S.~M., Fawzi, A., Uesato, J., and Frossard, P.
\newblock Robustness via curvature regularization, and vice versa.
\newblock In \emph{2019 IEEE/CVF Conference on Computer Vision and Pattern
  Recognition (CVPR)}, pp.\  9070--9078, Los Alamitos, CA, USA, Jun 2019. IEEE
  Computer Society.
\newblock \doi{10.1109/CVPR.2019.00929}.
\newblock URL
  \url{https://doi.ieeecomputersociety.org/10.1109/CVPR.2019.00929}.

\bibitem[Nakkiran et~al.(2019)Nakkiran, Kaplun, Kalimeris, Yang, Edelman,
  Zhang, and Barak]{incr_complexity_learning2019}
Nakkiran, P., Kaplun, G., Kalimeris, D., Yang, T., Edelman, B.~L., Zhang, F.,
  and Barak, B.
\newblock {SGD} on neural networks learns functions of increasing complexity.
\newblock In \emph{NeurIPS 2019 (spotlight)}, volume abs/1905.11604, 2019.
\newblock URL \url{http://arxiv.org/abs/1905.11604}.

\bibitem[Papyan(2019)]{Papyan2019structure}
Papyan, V.
\newblock Measurements of three-level hierarchical structure in the outliers in
  the spectrum of deepnet hessians.
\newblock In Chaudhuri, K. and Salakhutdinov, R. (eds.), \emph{Proceedings of
  the 36th International Conference on Machine Learning}, volume~97 of
  \emph{Proceedings of Machine Learning Research}, pp.\  5012--5021. PMLR,
  09--15 Jun 2019.
\newblock URL \url{https://proceedings.mlr.press/v97/papyan19a.html}.

\bibitem[Papyan(2020)]{Papyan2020structure}
Papyan, V.
\newblock Traces of class/cross-class structure pervade deep learning spectra.
\newblock \emph{J. Mach. Learn. Res.}, 21\penalty0 (1), 2020.
\newblock ISSN 1532-4435.
\newblock \doi{10.5555/3455716.3455968}.

\bibitem[Paszke et~al.(2019)Paszke, Gross, Massa, Lerer, Bradbury, Chanan,
  Killeen, Lin, Gimelshein, Antiga, Desmaison, Kopf, Yang, DeVito, Raison,
  Tejani, Chilamkurthy, Steiner, Fang, Bai, and Chintala]{PyTorch}
Paszke, A., Gross, S., Massa, F., Lerer, A., Bradbury, J., Chanan, G., Killeen,
  T., Lin, Z., Gimelshein, N., Antiga, L., Desmaison, A., Kopf, A., Yang, E.,
  DeVito, Z., Raison, M., Tejani, A., Chilamkurthy, S., Steiner, B., Fang, L.,
  Bai, J., and Chintala, S.
\newblock Pytorch: An imperative style, high-performance deep learning library.
\newblock In Wallach, H., Larochelle, H., Beygelzimer, A., d\textquotesingle
  Alch\'{e}-Buc, F., Fox, E., and Garnett, R. (eds.), \emph{Advances in Neural
  Information Processing Systems}, volume~32. Curran Associates, Inc., 2019.
\newblock URL
  \url{https://proceedings.neurips.cc/paper/2019/file/bdbca288fee7f92f2bfa9f7012727740-Paper.pdf}.

\bibitem[Pearlmutter(1994)]{Pearlmutter94}
Pearlmutter, B.~A.
\newblock Fast exact multiplication by the hessian.
\newblock \emph{Neural Computation}, 6:\penalty0 147--160, 1994.
\newblock \doi{10.1162/neco.1994.6.1.147}.

\bibitem[Petzka et~al.(2021)Petzka, Kamp, Adilova, Sminchisescu, and
  Boley]{petzka2021relative}
Petzka, H., Kamp, M., Adilova, L., Sminchisescu, C., and Boley, M.
\newblock Relative flatness and generalization.
\newblock In Beygelzimer, A., Dauphin, Y., Liang, P., and Vaughan, J.~W.
  (eds.), \emph{Advances in Neural Information Processing Systems}, 2021.
\newblock URL \url{https://openreview.net/forum?id=sygvo7ctb_}.

\bibitem[Qin et~al.(2019)Qin, Martens, Gowal, Krishnan, Dvijotham, Fawzi, De,
  Stanforth, and Kohli]{Qin2019adversarial}
Qin, C., Martens, J., Gowal, S., Krishnan, D., Dvijotham, K., Fawzi, A., De,
  S., Stanforth, R., and Kohli, P.
\newblock Adversarial robustness through local linearization.
\newblock In Wallach, H., Larochelle, H., Beygelzimer, A., d\textquotesingle
  Alch\'{e}-Buc, F., Fox, E., and Garnett, R. (eds.), \emph{Advances in Neural
  Information Processing Systems}, volume~32. Curran Associates, Inc., 2019.
\newblock URL
  \url{https://proceedings.neurips.cc/paper/2019/file/0defd533d51ed0a10c5c9dbf93ee78a5-Paper.pdf}.

\bibitem[Ramamurthy et~al.(2018)Ramamurthy, Varshney, and
  Mody]{ramamurthy2018topological}
Ramamurthy, K.~N., Varshney, K.~R., and Mody, K.
\newblock Topological data analysis of decision boundaries with application to
  model selection, 2018.

\bibitem[Sagun et~al.(2017)Sagun, Bottou, and LeCun]{Sagun2016}
Sagun, L., Bottou, L., and LeCun, Y.
\newblock Eigenvalues of the {Hessian} in deep learning: Singularity and
  beyond, 2017.
\newblock URL \url{https://openreview.net/forum?id=B186cP9gx}.

\bibitem[Sagun et~al.(2018)Sagun, Evci, G{\"{u}}ney, Dauphin, and
  Bottou]{Sagun2018}
Sagun, L., Evci, U., G{\"{u}}ney, V.~U., Dauphin, Y., and Bottou, L.
\newblock {Empirical analysis of the Hessian of over-parametrized neural
  networks}.
\newblock In \emph{ICLR 2018 - 6th Int. Conf. Learn. Represent.}, 2018.
\newblock URL \url{https://openreview.net/forum?id=rJrTwxbCb}.

\bibitem[Shah et~al.(2020)Shah, Tamuly, Raghunathan, Jain, and
  Netrapalli]{Shah2022simplbias}
Shah, H., Tamuly, K., Raghunathan, A., Jain, P., and Netrapalli, P.
\newblock The pitfalls of simplicity bias in neural networks.
\newblock \emph{CoRR}, abs/2006.07710, 2020.
\newblock URL \url{https://arxiv.org/abs/2006.07710}.

\bibitem[Srinivas et~al.(2022)Srinivas, Matoba, Lakkaraju, and
  Fleuret]{Srinivas2022flatten}
Srinivas, S., Matoba, K., Lakkaraju, H., and Fleuret, F.
\newblock Efficiently training low-curvature neural networks, 2022.
\newblock URL \url{https://arxiv.org/abs/2206.07144}.

\bibitem[Wilson et~al.(2017)Wilson, Roelofs, Stern, Srebro, and
  Recht]{wilson2017marginal}
Wilson, A.~C., Roelofs, R., Stern, M., Srebro, N., and Recht, B.
\newblock The marginal value of adaptive gradient methods in machine learning.
\newblock \emph{Advances in neural information processing systems}, 30, 2017.

\bibitem[Wu et~al.(2017)Wu, Zu, and E]{Wu2017}
Wu, L., Zu, Z., and E, W.
\newblock Towards understanding generalization of deep learning: Perspective of
  loss landscapes.
\newblock \emph{arXiv:1706.10239}, 2017.
\newblock URL \url{https://arxiv.org/abs/1706.10239}.

\bibitem[Yu et~al.(2021)Yu, Yao, Gholami, Dong, Mahoney, and Keutzer]{Yu2021}
Yu, S., Yao, Z., Gholami, A., Dong, Z., Mahoney, M.~W., and Keutzer, K.
\newblock Hessian-aware pruning and optimal neural implant.
\newblock \emph{CoRR}, abs/2101.08940, 2021.
\newblock URL \url{https://arxiv.org/abs/2101.08940}.

\bibitem[Zhang et~al.(2021)Zhang, Bengio, Hardt, Recht, and
  Vinyals]{Zhang2021rethinking}
Zhang, C., Bengio, S., Hardt, M., Recht, B., and Vinyals, O.
\newblock Understanding deep learning (still) requires rethinking
  generalization.
\newblock \emph{Commun. ACM}, 64\penalty0 (3):\penalty0 107–115, Feb 2021.
\newblock ISSN 0001-0782.
\newblock \doi{10.1145/3446776}.
\newblock URL \url{https://doi.org/10.1145/3446776}.

\bibitem[Zhang et~al.(2019)Zhang, Yu, Jiao, Xing, Ghaoui, and
  Jordan]{Zhang2019TRADES}
Zhang, H., Yu, Y., Jiao, J., Xing, E., Ghaoui, L.~E., and Jordan, M.
\newblock Theoretically principled trade-off between robustness and accuracy.
\newblock In Chaudhuri, K. and Salakhutdinov, R. (eds.), \emph{Proceedings of
  the 36th International Conference on Machine Learning}, volume~97 of
  \emph{Proceedings of Machine Learning Research}, pp.\  7472--7482. PMLR,
  09--15 Jun 2019.
\newblock URL \url{https://proceedings.mlr.press/v97/zhang19p.html}.

\end{thebibliography}
\newpage

\appendix
%\tableofcontents

\begin{center}
    \huge{\textbf{Supplementary material}}\\
    \vspace{0.3cm}
    \large{accompanying ``\textit{Unveiling the Hessian's Connection to~the~Decision~Boundary}''}
    \vspace{0.6cm}
\end{center}
\normalsize

We provide the following results in the supplementary material.
\begin{itemize}
    \item \textbf{Section~\ref{app:dotvscosine}}: Alignment of vectors: cosine similarity vs scalar product
    \item \textbf{Section~\ref{app:sim_datasets}}: The top Hessian eigenvectors and decision boundary for the additional datasets
    \item \textbf{Section~\ref{sec:theory}}: Theoretical analysis
    \item \textbf{Section~\ref{app:other_directions}}: Directions other than the top Hessian eigenvectors do not align with the decision boundary
    \item \textbf{Section~\ref{app:per-class-boundary}}: Decision boundary per class
    \item \textbf{Section~\ref{app:ablation}}: Results are invariant to an architecture, loss, and optimizer: \emph{gaussian}
    \item \textbf{Section~\ref{app:grad_hessian}}: The gradient covariance matrix vs the Hessian at the minimum
    \item \textbf{Section~\ref{app:dynamics}}: Decision boundaries during the training
    \item \textbf{Section~\ref{app:generalization_measure_sim}}: Generalization measure for all simulated datasets and its limitations
    \item \textbf{Section~\ref{app:reparameterization}}: Generalization measure is invariant to model reparameterization
    \item \textbf{Section~\ref{app:Iris}}: Hessian analysis for \emph{Iris}
    \item \textbf{Section~\ref{app:MNIST}}: Hessian analysis for \emph{MNIST}
    \item \textbf{Section~\ref{app:gen_real}}: Generalization measure for \emph{Iris} and \emph{MNIST}
\end{itemize}
    \vspace{0.6cm}
\section{Alignment of vectors: cosine similarity vs scalar product}\label{app:dotvscosine}

In Equation~\eqref{eq:alignment}, we have defined the \emph{alignment} between the reinforcing gradient $g_\theta(x)$ from Equation~\eqref{eq:reinf_grads} of an input $x$ with eigenvector $v_i$ in terms of the cosine similarity as
\begin{align}
     \mathcal{A}_i(x) = \frac{\left \langle g_\theta(x) , v_i \right\rangle}{\|g_\theta(x)\| \|v_i\|},
\end{align}
where $\langle\cdot,\cdot\rangle$ is the scalar product, and $\|\cdot\|$ is the Euclidean norm of the vector. 

\begin{figure}[h]
    \centering
\includegraphics[width=\textwidth]{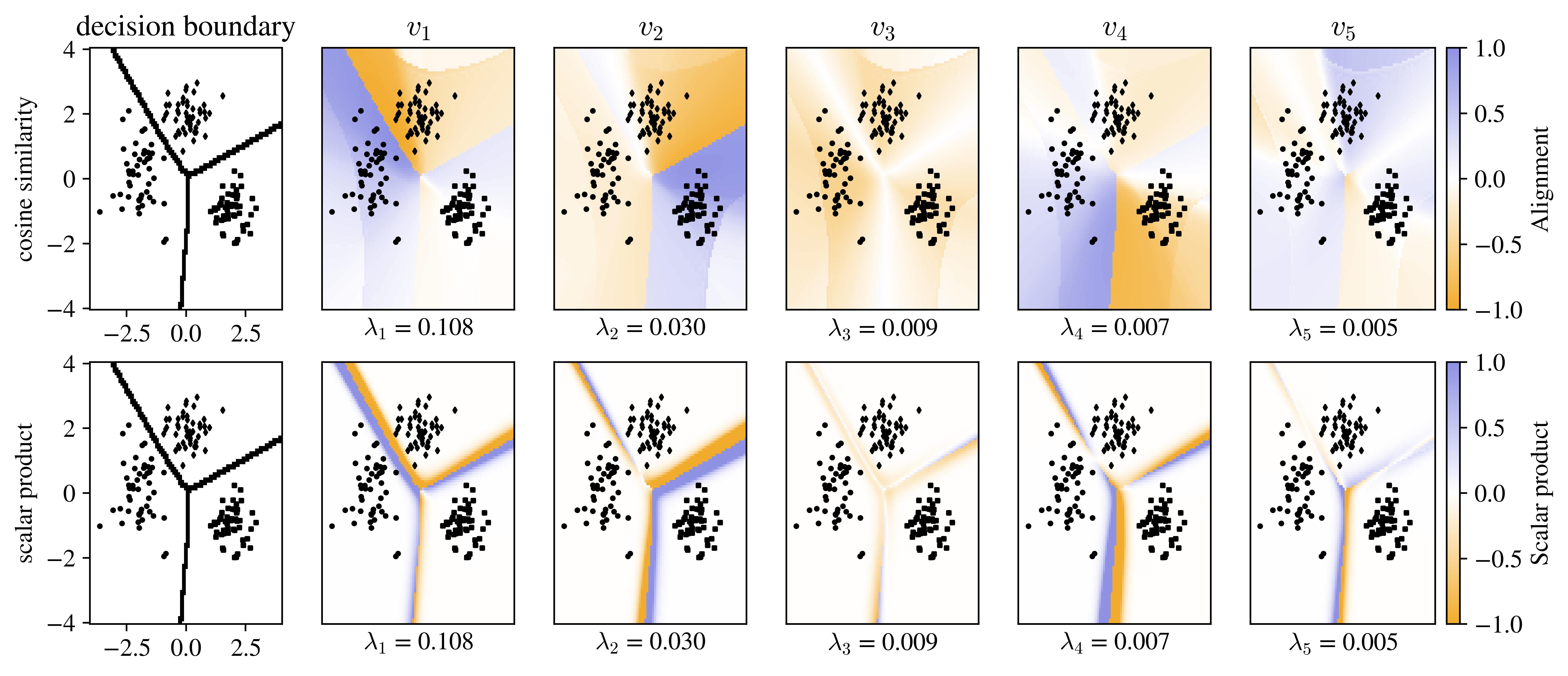}
    \caption{\textbf{Comparison of similarity metrics for \emph{gaussian}.} \emph{(Top)} Cosine similarity. \emph{(Bottom)} Scalar product with a color bar limited to $[-1,+1]$ (without normalizing the scalar product values).}
    \label{fig:scalar_vs_cosine}
\end{figure}

As the main findings of our work result from comparing gradients of loss of individual training samples with the Hessian top eigenvectors, one may ask why we chose cosine similarity as the measure of similarity between the vectors instead, e.g., of the scalar product itself, $\left \langle g_\theta(x) , v_i \right\rangle$. We compare the two similarity metrics in Figure~\ref{fig:scalar_vs_cosine}, where we immediately see the main weakness of the scalar product when it comes to studying the input space. First of all, due to the lack of gradients' normalization, the overlap highlights only points on the decision boundary. Large norm of gradients on the boundary dominates any existing alignment between the Hessian eigenvectors and gradients of samples far from the boundary. Secondly, contrary to the cosine similarity, the scalar product has no maximal value, which could guide the analysis of the decision boundary decomposition in terms of the Hessian eigenvectors.

\section{The top Hessian eigenvectors and decision boundary for the additional simulated datasets}\label{app:sim_datasets}
Within this work, we use five simulated 2D datasets: \emph{gaussian} with three classes, concentric \emph{circle} and \emph{half-moon} datasets with two classes, \emph{hierarchical gaussian} with four classes, and \emph{checkerboard} dataset with two classes. Details on their generation can be found in the code available \href{https://github.com/Shmoo137/Hessian-and-Decision-Boundary}{here} (\texttt{src/datasets.py}).

\begin{figure}[hb!]
    \centering
    \includegraphics[width=\textwidth]{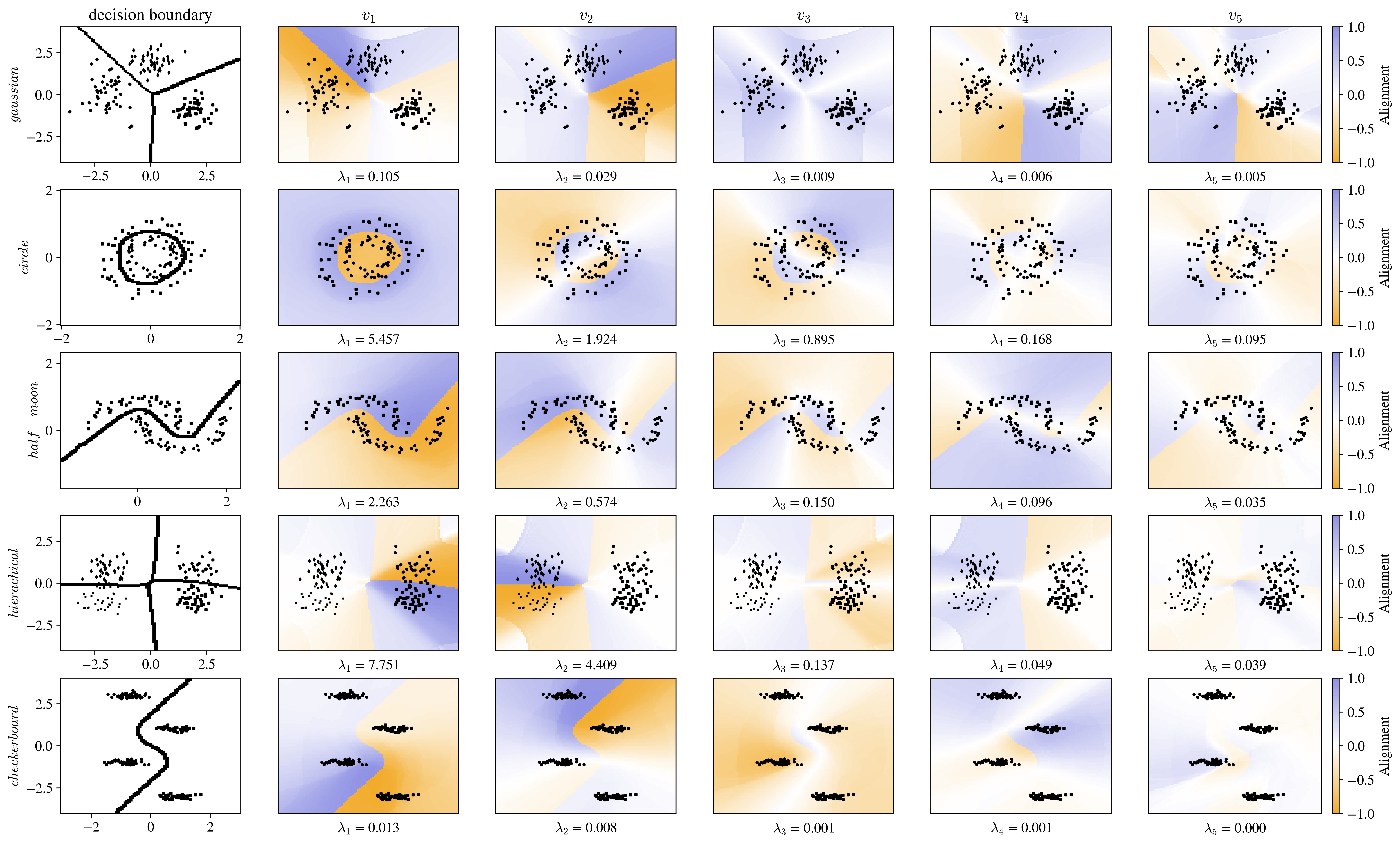}
    \caption{\textbf{Top Hessian eigenvectors and decision boundaries for all simulated datasets.} \emph{(First column)} The decision boundary in the data space obtained by training a two-layered fully connected network. \emph{(Other columns)} The alignment with the top five eigenvectors illustrates that the top eigenvectors encode the decision boundary.}
    \label{fig:all_sim_datasets}
\end{figure}

In Figure~\ref{fig:all_sim_datasets}, we present the alignment of the reinforcing gradients of input samples with the top five Hessian eigenvectors for all the simulated datasets. We see consistently that \emph{the top Hessian eigenvectors align with gradients of loss of samples at the boundary.}

\begin{figure}[ht!]
    \centering
    \subfigure{\includegraphics[width=0.3\linewidth, scale=0.7]{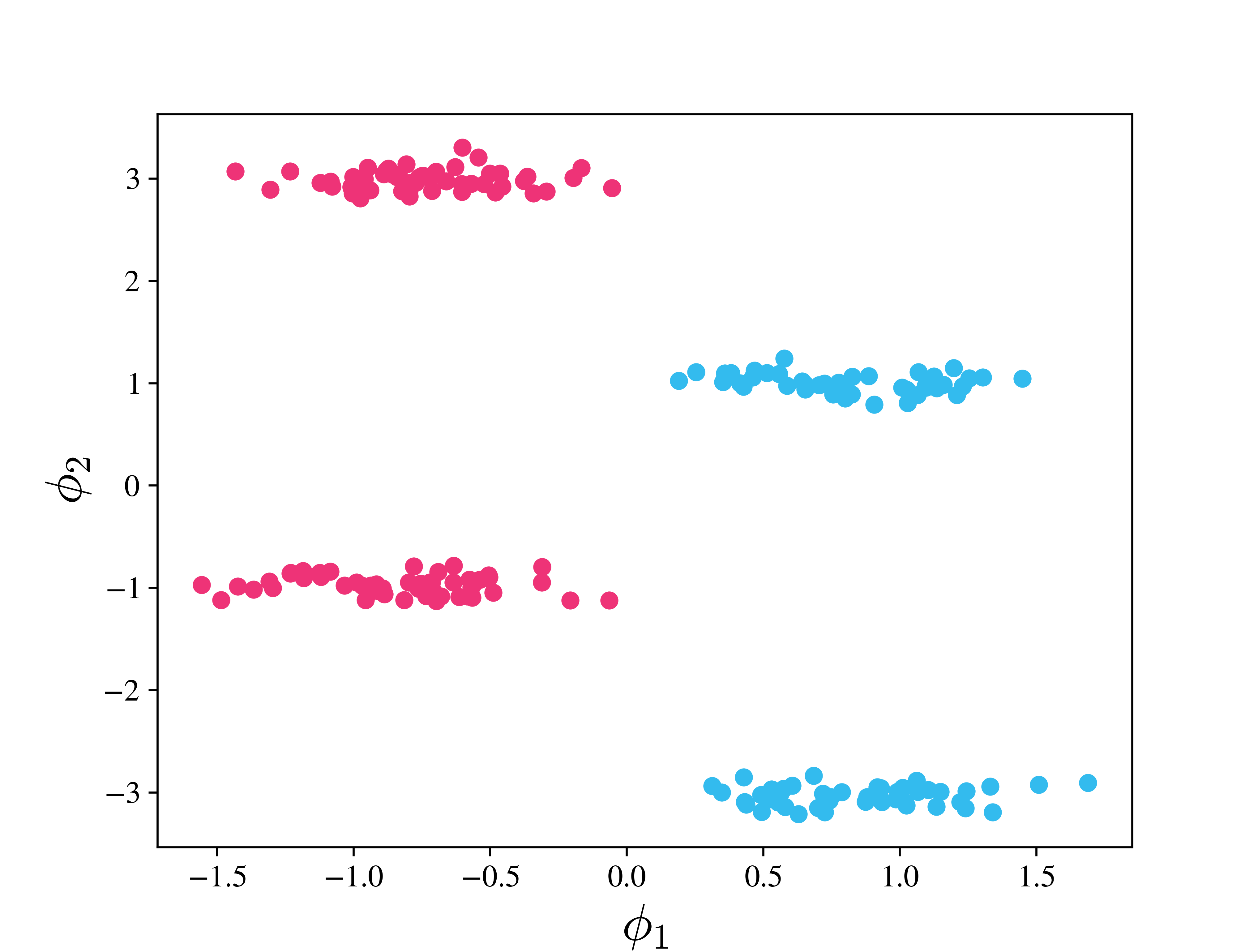}}
    \subfigure{\includegraphics[width=0.3\linewidth, scale=0.7]{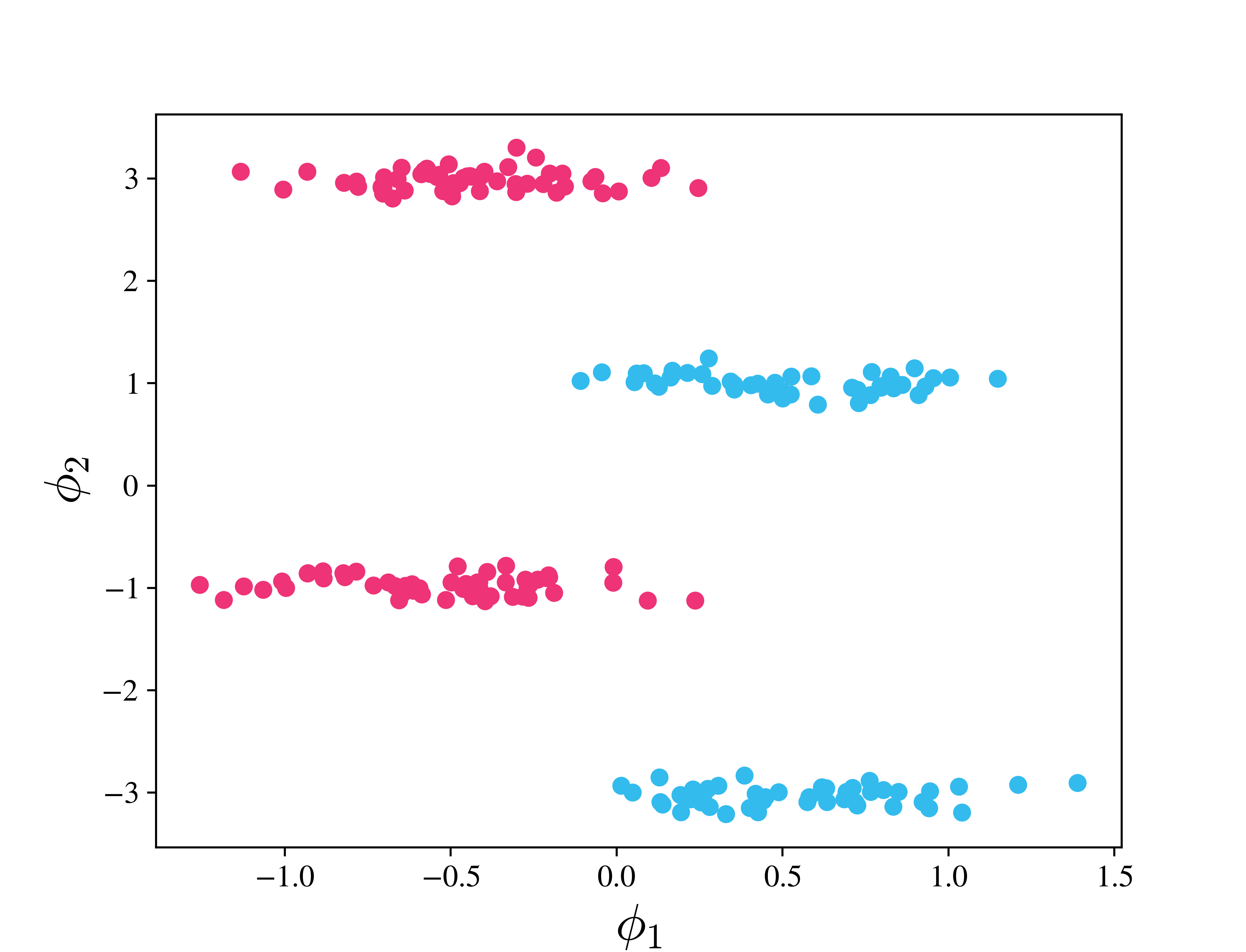}}
    \caption{\textbf{The \emph{checkerboard} datasets} used for \emph{(Left)} training and \emph{(Right)} pretraining the wide-margin setup in section~\ref{ss:results-simplicity-bias} on simplicity bias analysis.}
    \label{fig:checkerboard_datasets}
\end{figure}
 
Finally, let us here specify \textit{how we have obtained the narrow- and wide-margin solutions} for the classification of the \emph{checkerboard} dataset in section~\ref{ss:results-simplicity-bias}. The mentioned \emph{checkerboard} dataset consists of the two-class Gaussian mixture and is presented in the left panel of Figure~\ref{fig:checkerboard_datasets}. We train on it to achieve the linear boundary with a narrow margin due to simplicity bias inherent to neural networks. The narrow-margin solution is presented in the top row of Figure~\ref{fig:simplbias_margin}. We call this ``normal initialization'', and we follow the same training procedure here as for all other normally initialized datasets. 
To obtain the wide-margin solutions presented both in the bottom row in Figure~\ref{fig:simplbias_margin} and in the last row of Figure~\ref{fig:all_sim_datasets}, we initialize the network with what we call \textit{the wide-margin initialization}. We achieve this initialization as follows. First, we pretrain until $100\%$ accuracy using the usual procedure on the dataset presented in the right panel of Figure~\ref{fig:checkerboard_datasets}. These are also Gaussian mixture clusters with the same variance parameters as the \emph{checkerboard} dataset but with means in feature $\phi_1$ that bring the clusters closer, keeping feature $\phi_2$ the same. Then, we train on the first \emph{checkerboard} dataset until $100\%$ accuracy. Such a pretraining results in a decision boundary with a wider margin than without pretraining.

Let us also note the \textit{difference between the wide-margin solutions} presented in the bottom row in Figure~\ref{fig:simplbias_margin} and in the last row of Figure~\ref{fig:all_sim_datasets}. Interestingly, the top Hessian eigenvectors aligns with samples at the decision boundary in different ways resulting from the stochasticity of the training. In the bottom row in Figure~\ref{fig:simplbias_margin}, we see that the top Hessian eigenvector aligns with almost all samples across the decision boundary. Here, in the last row of Figure~\ref{fig:all_sim_datasets}, the alignment separates across two top Hessian eigenvectors. Both solutions have approximately equally wide margins and similar generalization abilities. It further shows that the separation across the top Hessian eigenvectors results not only from the complexity of the decision boundary, but can change across runs and initializations.

\section{Theoretical analysis}
\label{sec:theory}
We consider that the converged minimum $\theta := \theta^*$ is an exact minimum, meaning that the loss and its gradient at $\theta^*$ is zero, i.e., $\mathcal{L}( \theta^*; \mathcal{D} ) = 0$ and $\nabla_\theta \mathcal{L}( \theta^*; \mathcal{D} ) = \mathbf{0}$. Using this information, we expand the loss using the second-order Taylor's approximation at $\theta := \theta^*$.
\begin{align}
    \mathcal{L}( \theta^* + \Delta \theta; \mathcal{D} ) &= \mathcal{L}( \theta^*; \mathcal{D} ) + \nabla_\theta \mathcal{L}( \theta^*; \mathcal{D} )^T \Delta \theta + \dfrac{1}{2} \Delta \theta^T \nabla_\theta^2 \mathcal{L}( \theta^*; \mathcal{D} ) \Delta \theta  \nonumber \\
    \mathcal{L}( \theta^* + \Delta \theta; \mathcal{D} ) &= \dfrac{1}{2} \Delta \theta^T H_{\theta^*} \Delta \theta  \nonumber
\end{align}
$H_{\theta^*}$ is the Hessian of the training loss function evaluated at the minimum. We denote its eigenvectors and corresponding eigenvalues as $v_i$ and $\lambda_i$. Now, let's consider $\Delta \theta := \frac{g_\theta(x)}{|| g_\theta(x) ||}$ to be a reinforcing gradient of some input $x$ in the dataset $\mathcal{D} := \{ \mathcal{X}, \mathcal{Y} \}$, and an overparametrized classifier (e.g., neural network) with $p$ parameters denoted by $f(\theta, \mathcal{X})$. 
\begin{align}
    \mathcal{L}\left( f \left( \theta^* + \frac{g_\theta(x)}{|| g_\theta(x) ||} , \mathcal{X} \right), \mathcal{Y} \right) &= \dfrac{1}{2} \dfrac{g_\theta(x)^T}{|| g_\theta(x) ||} \sum_{i=1}^{p} \lambda_i v_i v_i^T \dfrac{g_\theta(x)}{|| g_\theta(x) ||}  \nonumber \\
    &= \dfrac{1}{2} \sum_{i=1}^{p} \lambda_i \dfrac{\left \langle g_\theta(x), v_i\right \rangle}{|| g_\theta(x) ||} \dfrac{\left \langle g_\theta(x),v_i \right \rangle}{|| g_\theta(x) ||} \nonumber \\
    \mathcal{L}\left( f \left( \theta^*, \mathcal{X}  \right) + \nabla_\theta f \left( \theta^* , \mathcal{X} \right)^T \frac{g_\theta(x)}{|| g_\theta(x) ||} , \mathcal{Y} \right) &= \dfrac{1}{2} \sum_{i=1}^p \lambda_i \mathcal{A}_{i}(x)^2 \nonumber \\
    \mathcal{L}\left( \mathcal{Y} + \nabla_\theta f \left( \theta^*, \mathcal{X} \right)^T \frac{g_\theta(x)}{|| g_\theta(x) ||} , \mathcal{Y} \right) &= \dfrac{1}{2} \sum_{i=1}^p \lambda_i \mathcal{A}_{i}(x)^2 \label{eq:taylor_l_a}
\end{align}
From Equation~\eqref{eq:taylor_l_a}, for the loss to have a maximal change, the reinforcing gradient of $x$ should be aligned with the direction of the steepest ascent of $f(\theta^*, \mathcal{X})$. 
This implies that moving data $x$ in the direction of the gradient of $f(\theta^*, \mathcal{X})$ potentially changes the predicted class for $x$, thus increasing the loss.
In other words, the alignment of the reinforcing gradient of $x$ with the function's gradient is high for $x$ near the decision boundary.
From this understanding, we infer that the alignment of $x$ with the Hessian eigenvectors is larger for $x$ near the boundary than data points farther away from the right-hand side of Equation~\eqref{eq:taylor_l_a}. 
%implying that the data $x$ is at the decision boundary. Therefore, the right side of the equation has to be maximal for $x$ at the decision boundary and small for $x$ that is far from the boundary. 
As the alignment $\mathcal{A}(x)$ considers a normalized $g_\theta(x)$, this variability comes only from a different alignment of $g_{\theta}(x)$ for $x$'s close and far from the boundary with different Hessian eigenvectors. As stated multiple times, the behavior of the Hessian spectra in deep learning setups is universal. Its spectrum has a small number of positive non-zero eigenvalues, $\lambda_0, \lambda_1, \ldots, \lambda_t$, and the rest of the eigenvalues is close to zero. This, in turn, implies that the right side of the equation is large when $g_\theta(x)$ is aligned with the top Hessian eigenvectors with the largest eigenvalues $\lambda_0, \lambda_1, \ldots, \lambda_t$. This implication strengthens our numerical observations about the top Hessian eigenvectors being aligned with the gradients of loss of data at the boundary.

\section{Directions other than the top Hessian eigenvectors do not align with the decision boundary}\label{app:other_directions}

\begin{figure}[hb!]
    \centering
    \includegraphics[width=0.75\textwidth]{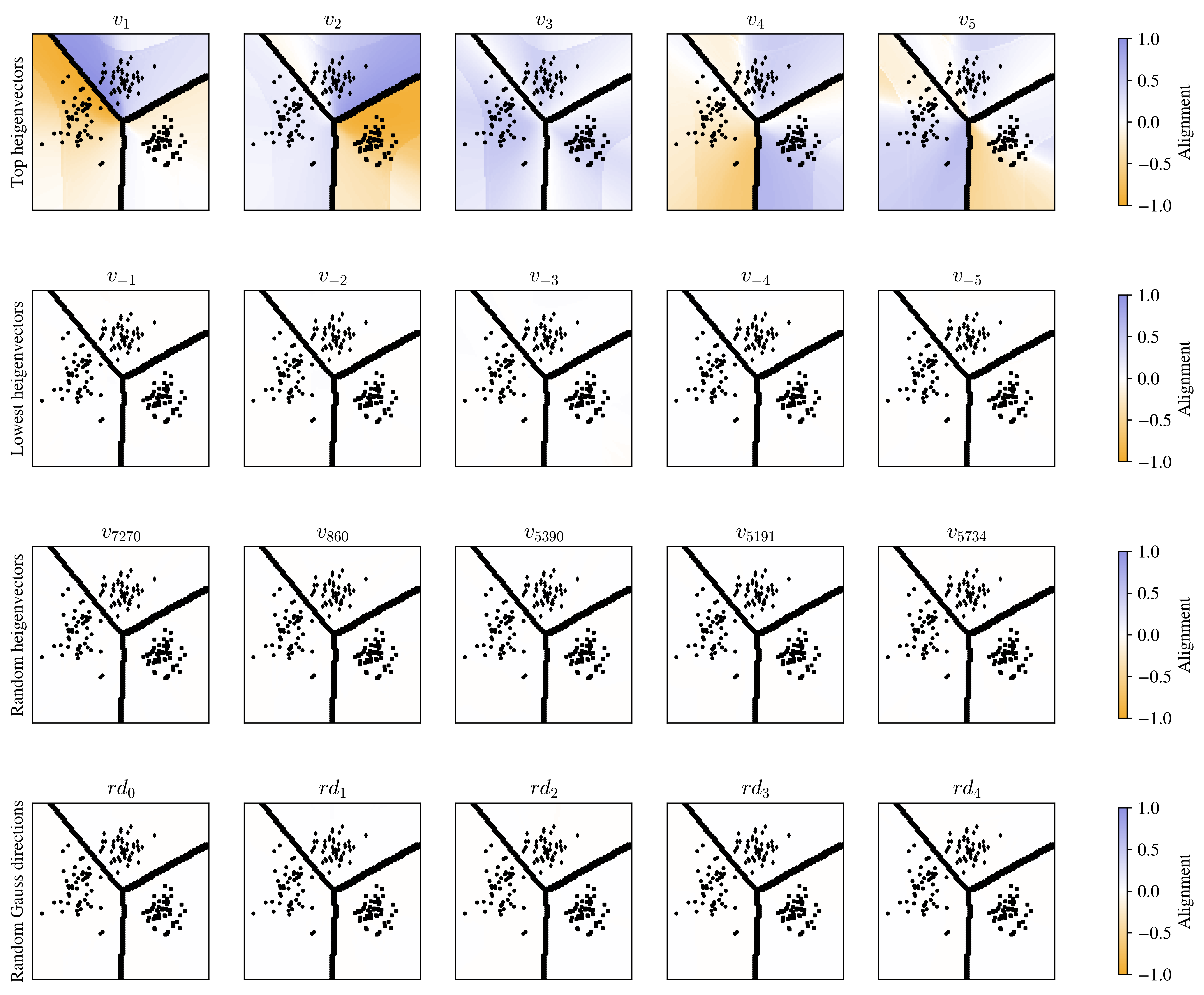}\caption{\textbf{Comparison of different directions in parameter space and their alignment with the reinforcing gradients of input points in the 2D plane.}
    Alignment color normalized to the interval between $[-1,+1]$. 
    We show the top eigenvectors $v_1,\ldots,v_5$ and compare them to the eigenvectors with the smallest eigenvalues where $v_{-1}$ has the smallest eigenvalue and $v_{-5}$ the fifth-smallest eigenvalue. We also sample some random directions from the Hessian eigenspace and finally compare to random directions in parameter space where each entry of the vector is sampled from a standard Gaussian.}
    \label{fig:other_directions}
\end{figure}

To test our finding connecting the top Hessian eigenvectors and decision boundary, we check the alignment of reinforcing gradients of individual input samples with vectors pointing in other directions in parameter space. In Figure~\ref{fig:other_directions} we compare the gradients' alignment with \emph{(1)} the top five Hessian eigenvectors, \emph{(2)} the bottom five Hessian eigenvectors (that correspond to the five smallest eigenvalues, i.e., five largest negative eigenvalues), \emph{(3)} five randomly selected Hessian eigenvectors, and \emph{(4)} five random
directions in parameter space where each vector element is sampled from a standard Gaussian. Within this comparison, the color bar is normalized across the plots to $[-1,+1]$. We immediately see that \emph{directions other than the top Hessian eigenvectors are not aligned with reinforcing gradients in the slightest.}

\begin{figure}[ht!]
    \centering
    \includegraphics[width=0.75\textwidth]{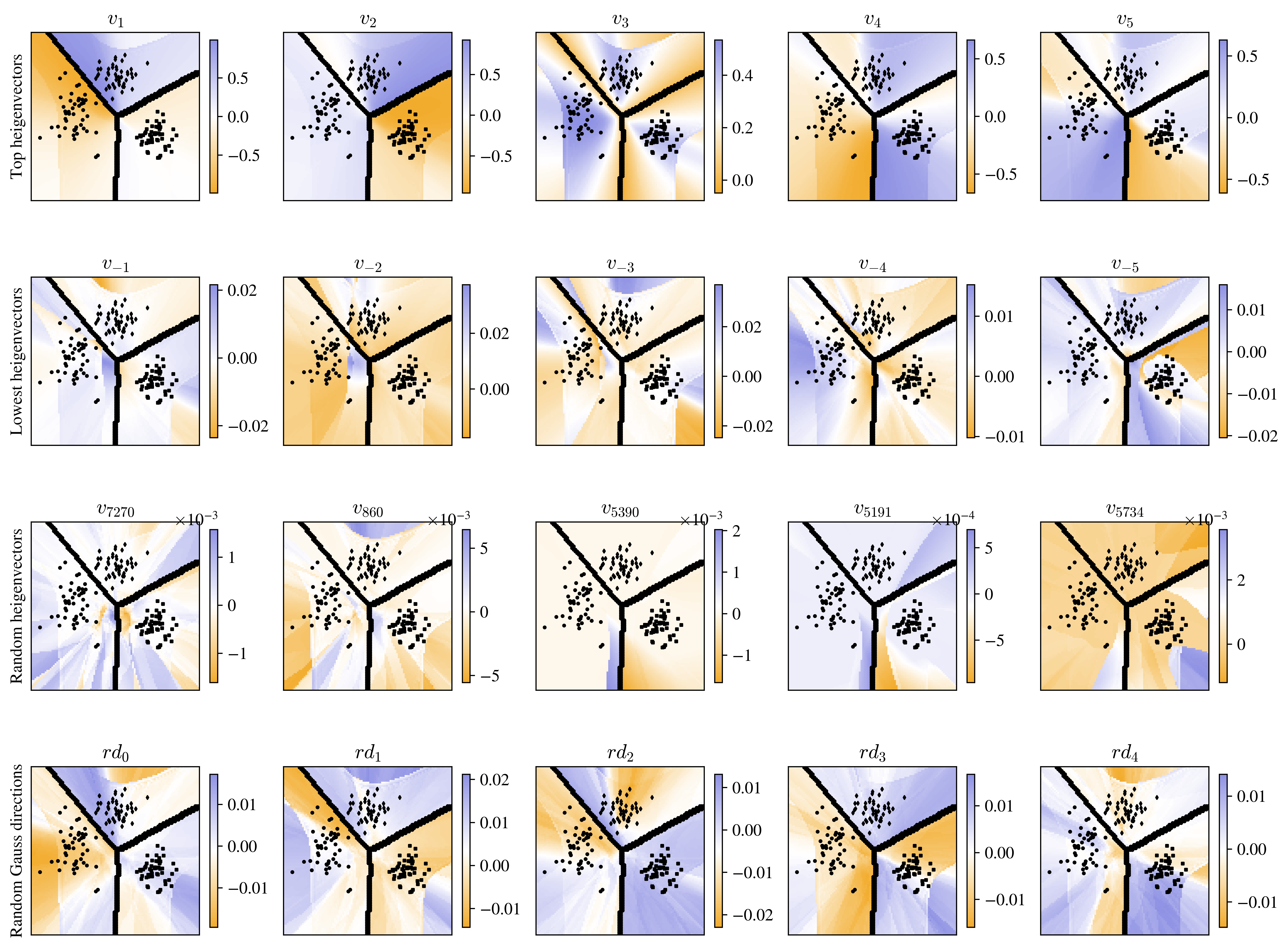}
    \caption{\textbf{Comparison of different directions in parameter space and their alignment with the reinforcing gradients of input points in the 2D plane - The color bars are normalized per plot.}
    We show the top eigenvectors $v_1,\ldots,v_5$ and compare them to the eigenvectors with the smallest eigenvalues where $v_{-1}$ has the smallest eigenvalue and $v_{-5}$ the fifth-smallest eigenvalue. We also sample some random directions from the Hessian eigenspace and finally compare to random directions in parameter space where each entry of the vector is sampled from a standard Gaussian.}
    \label{fig:other_directions_individ}
\end{figure}

Interestingly, when we use separate color bars for each plot and repeat the comparison in Figure~\ref{fig:other_directions_individ}, we see that the random directions may sometimes reflect the alignment of the gradients resulting from the decision boundary. Tiny alignment values around $10^{-2}$ show, however, that it reflects the coherence of gradients rather than encodes relevant directions in parameter space. \emph{The average of the maximal alignment of training reinforcing gradients with five randomly sampled directions in the parameter space serves therefore as a threshold $\epsilon$ in Equation~\ref{eq:gen_measure} for counting Hessian eigenvectors with ``non-zero'' alignment with reinforcing gradients.}

\section{Decision boundary per class}\label{app:per-class-boundary}

\begin{figure}[h!]
    \centering
    \includegraphics[width=0.6\textwidth]{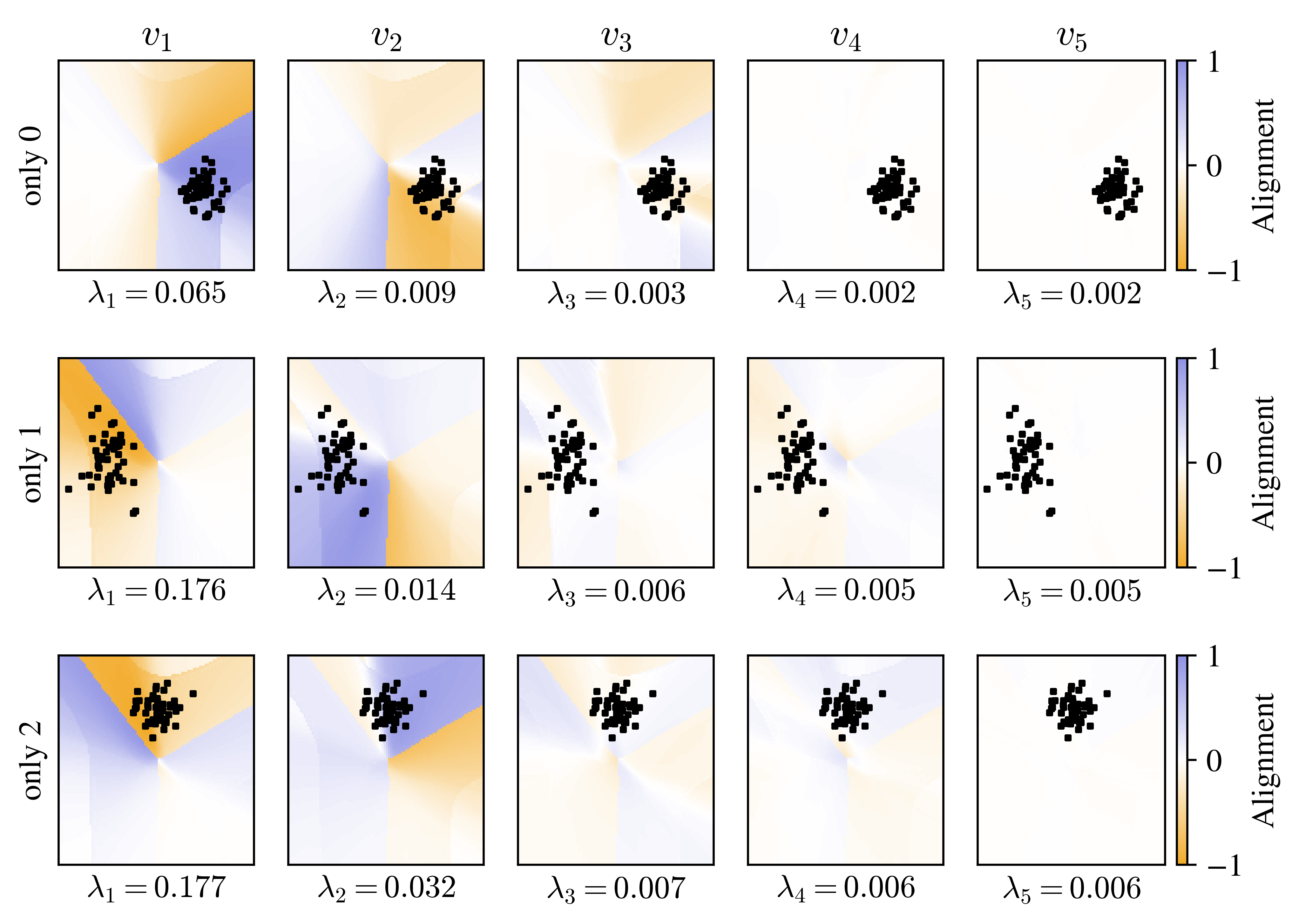}
    \caption{\textbf{Loss decomposed into separate classes.} The decision boundaries are equivalent to those from Figure~\ref{fig:gauss_normal_training}. Here, the training loss is decomposed into the losses of individual training points associated with a given class. The Hessian eigenvectors look different for each decomposition, and the top eigenvectors exactly show the decision boundary enclosing this class. Only the relevant class is shown.}
    \label{fig:encboundary2}
\end{figure}

The Hessian eigenvectors crucially depend on the loss landscape, which in turn depends on the data. 
When we analyze the Hessian with respect to the loss that only considers a specific class $c$, we observe that \emph{the top eigenvectors are now restricted to the boundaries that are relevant to deciding the ``all-against-one'' for the selected class $c$}.
The results are presented in Figure~\ref{fig:encboundary2}.

\section{Results are invariant to an architecture, loss, and optimizer: \emph{gaussian}}\label{app:ablation}

\begin{figure}
    \centering
\includegraphics[width=\textwidth]{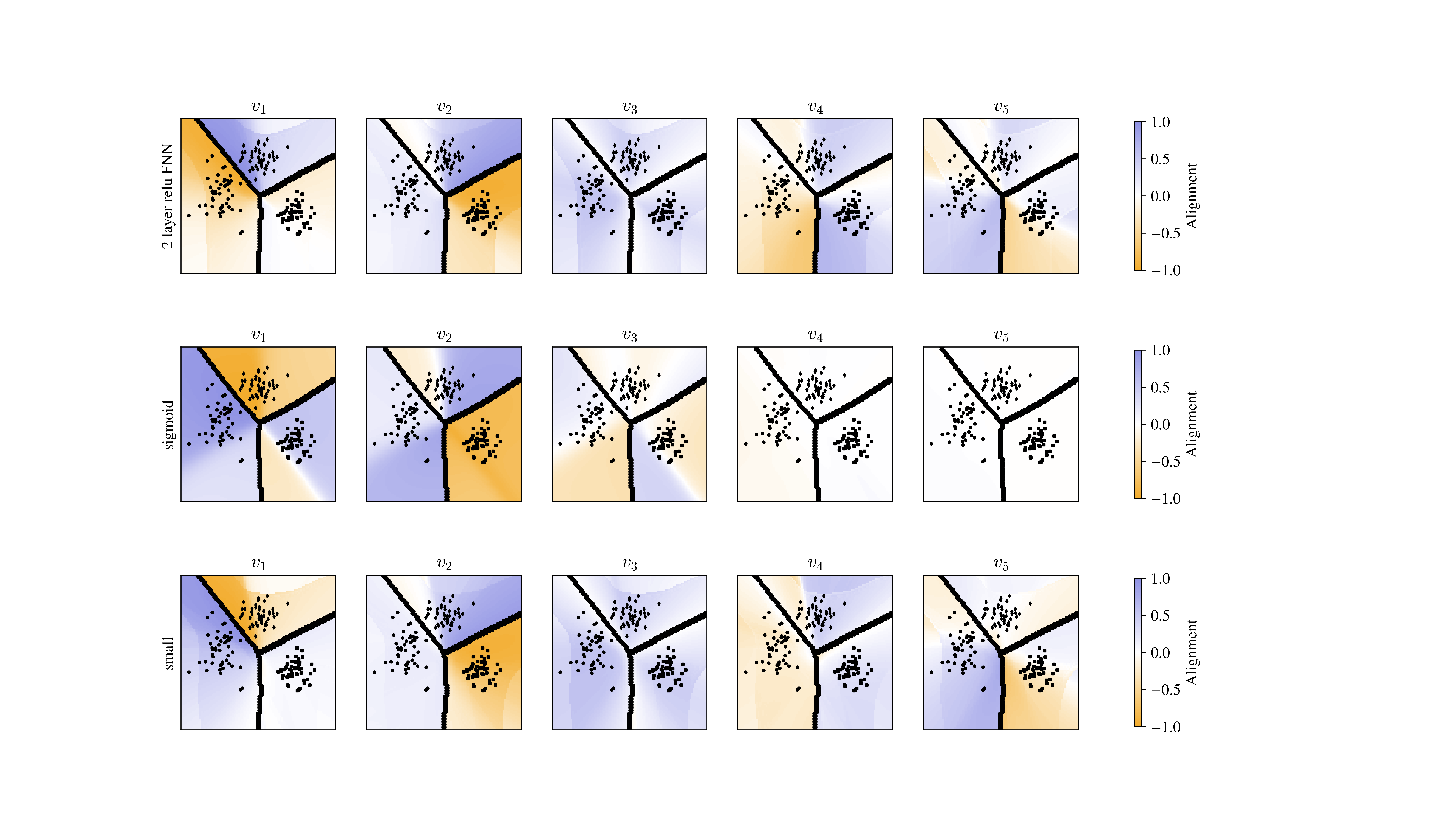}
    \vspace{-1em}
    \caption{\textbf{Top eigenvectors of the Hessian for alternative model architectures.} \emph{(First row)} A two-layer neural network with 100 neurons per layer and the ReLU activation function. \emph{(Second row)} The same with sigmoid activations. \emph{(Third row)} The same as first but with 50 neurons per layer instead.}
    \label{fig:vary_architecture}
\end{figure}

\begin{figure}
    \centering
\includegraphics[width=\textwidth]{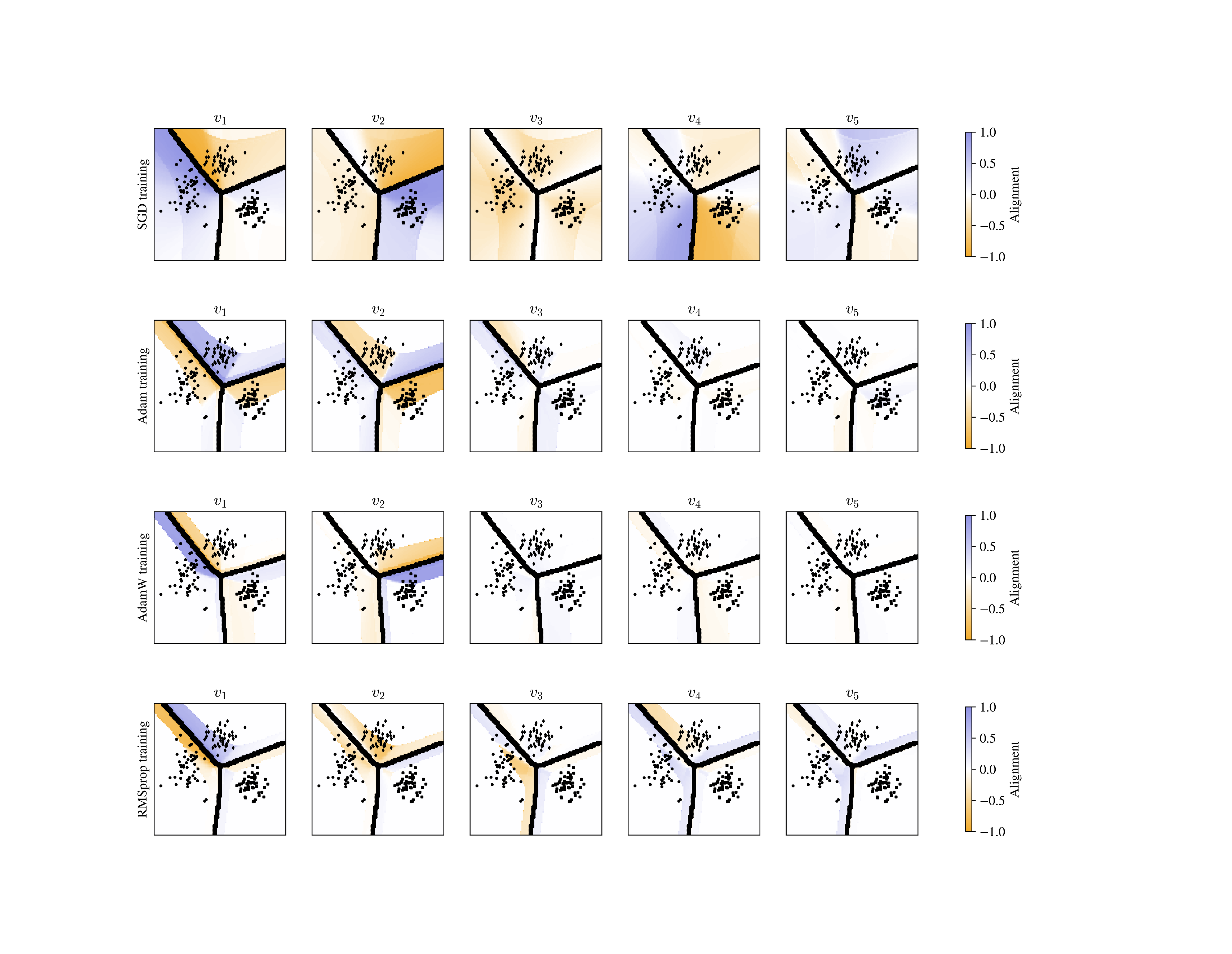}%{figures/invoptimizer_normal_training_adam_5_normcolor=True.overlap.png}
    \vspace{-1em}
    \caption{\textbf{Top eigenvectors of the Hessian for different optimizers:} \emph{(First row)} SGD, \emph{(Second row)} Adam, \emph{(Third row)} AdamW, and \emph{(Fourth row)} RMSprop with the same learning rate of 0.2 and a batch size of 64. For optimizers besides SGD, the boundaries are more ``clear cut''; The gradient at many places in the input space has zero alignment with their counterparts on the boundary.}
    \label{fig:vary_optimizer}
\end{figure}

\begin{figure}
    \centering
\includegraphics[width=\textwidth]{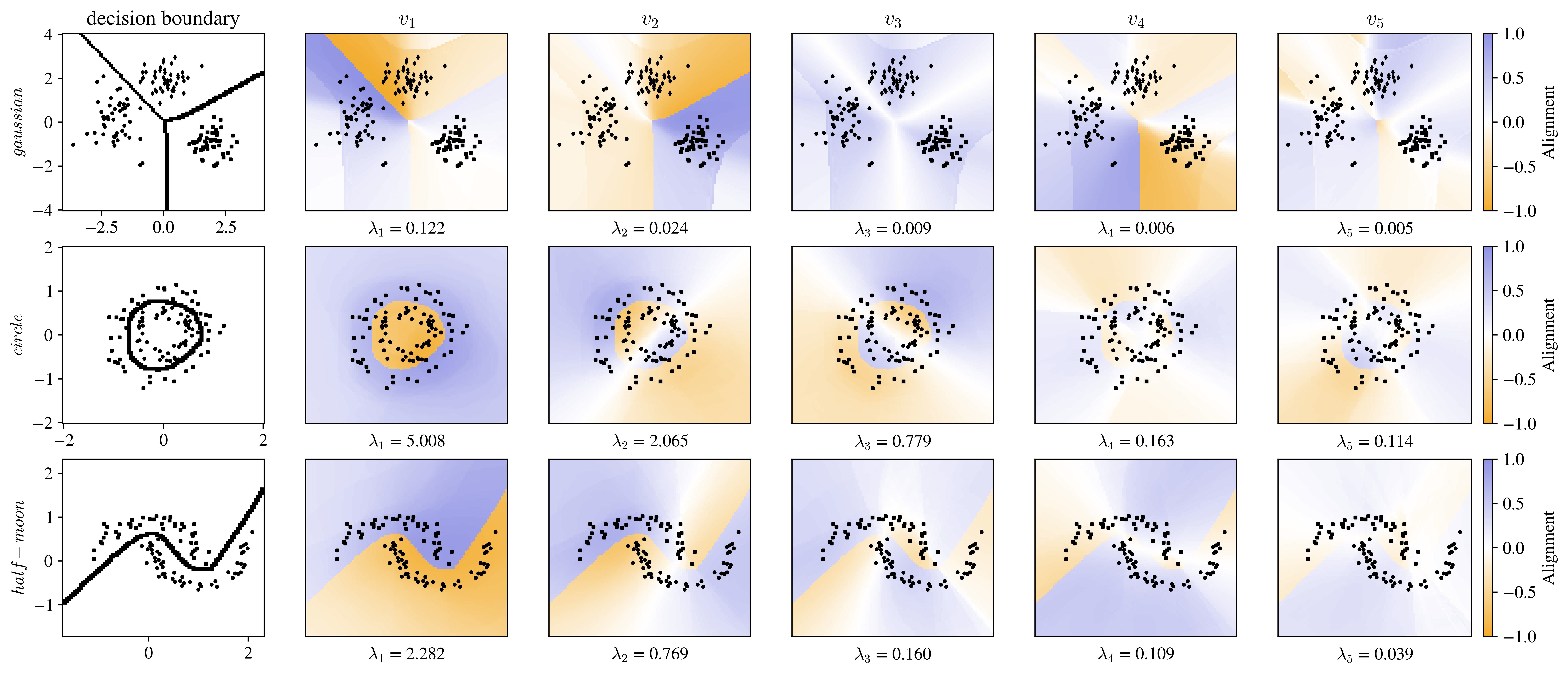}
    \caption{\textbf{Top eigenvectors of the Hessian for the negative log-likelihood loss (NLLLoss):} \emph{(Top)} \emph{gaussian} dataset. \emph{(Middle)} \emph{circle} dataset. \emph{(Bottom)} \emph{half-moon} dataset.}
    \label{fig:vary_loss}
\end{figure}

Here, we show that while the decision boundary may shift and the alignment values change, the connection between the topmost eigenvectors and \emph{the decision boundary is invariant to the architecture} (Figure~\ref{fig:vary_architecture}), \emph{the choice of the optimizer }(Figure \ref{fig:vary_optimizer}), and \emph{loss function} (Figure~\ref{fig:vary_loss}). 

Interestingly, while SGD, Adam, AdamW, and RMSprop in Figure \ref{fig:vary_optimizer} reach similar decision boundaries, values of alignment of gradients across the input space with the top eigenvectors vary significantly. In particular, in SGD, the non-zero alignment with the top eigenvectors is preserved for gradients far from the boundary. In Adam, AdamW, and RMSprop, the alignment goes quickly to zero with the distance from the boundary. It may suggest that in such cases, the gradient-based interpretation methods such as influence functions \citep{Koh17influence} may fail to find examples that are similar to a sample far from the boundary. At the same time, the connection between the decision boundary itself and the top Hessian eigenvectors is preserved regardless of the optimizer.

\newpage
\clearpage
\section{The gradient covariance matrix vs the Hessian at the minimum}\label{app:grad_hessian}
\begin{figure}[h]
    \centering
\includegraphics[width=\textwidth]{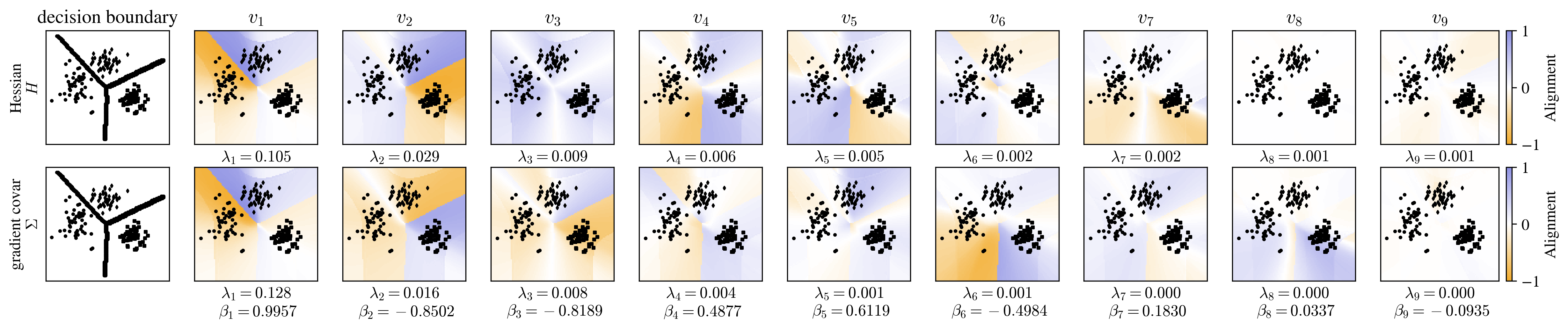}
    \caption{\textbf{Gradient covariance matrix vs. Hessian.} We compare the alignment of gradients of loss of input samples with the top eigenvectors of \emph{(Top)} the Hessian $H$ and \emph{(Bottom)} the gradient covariance matrix $\Sigma(\theta; \mathcal{D})$ defined in Equation~\ref{eq:cov_mat}. For the first two eigenvectors of both matrices, their alignment with gradients of input samples is very similar across the input space.
    The cosine similarity $\beta_i = \langle v_i^H, v_i^\Sigma\rangle$ of between the Hessian eigenvectors $v_i^H$ and the gradient covariance matrix' eigenvectors $v_i^C$  decreases with the values of their eigenvalues.}
    \label{fig:grad_covariance}
\end{figure}

Here, we study the covariance matrix of gradients of loss of individual training samples at the minimum defined as 
\begin{align}\label{eq:cov_mat}
\Sigma(\theta; \mathcal{D}) &= \frac{1}{n} \sum_{i=1}^n g_{\theta} (x_i) g_{\theta}^T (x_i) \nonumber \\
&= \frac{1}{n} \sum_{i=1}^n  \frac{\partial}{\partial \theta} \mathcal{L}\left(\theta;\{x_i,y_i\}\right) \frac{\partial}{\partial \theta} \mathcal{L}^T\left(\theta;\{x_i,y_i\}\right)\,,
\end{align}
where training data $\mathcal{D} = \{x_i,y_i\}_{i=1}^{n}$, $x_i \in \mathbb{R}^d$, and $y_i \in \{1,\dots, C\}$ is the class label.
In Figure~\ref{fig:grad_covariance}, we show that the top few eigenvectors of the covariance matrix $\Sigma(\theta; \mathcal{D})$ actually encode the same information as the top few Hessian eigenvectors at the minimum as observed by \citet{Ghorbani2019} and \citet{fort2019emerglocalgeom}. This observation can also be justified by the Hessian approximation with the gradient outer product holds well at the minimum. Therefore, \emph{the top subspace of $\Sigma(\theta; \mathcal{D})$ can be used instead of the more computationally expensive Hessian to study the decision boundary.}
%\newpage
\section{Decision boundaries during the training}\label{app:dynamics}

\begin{figure}
    \centering
\includegraphics[width=1.1\textwidth]{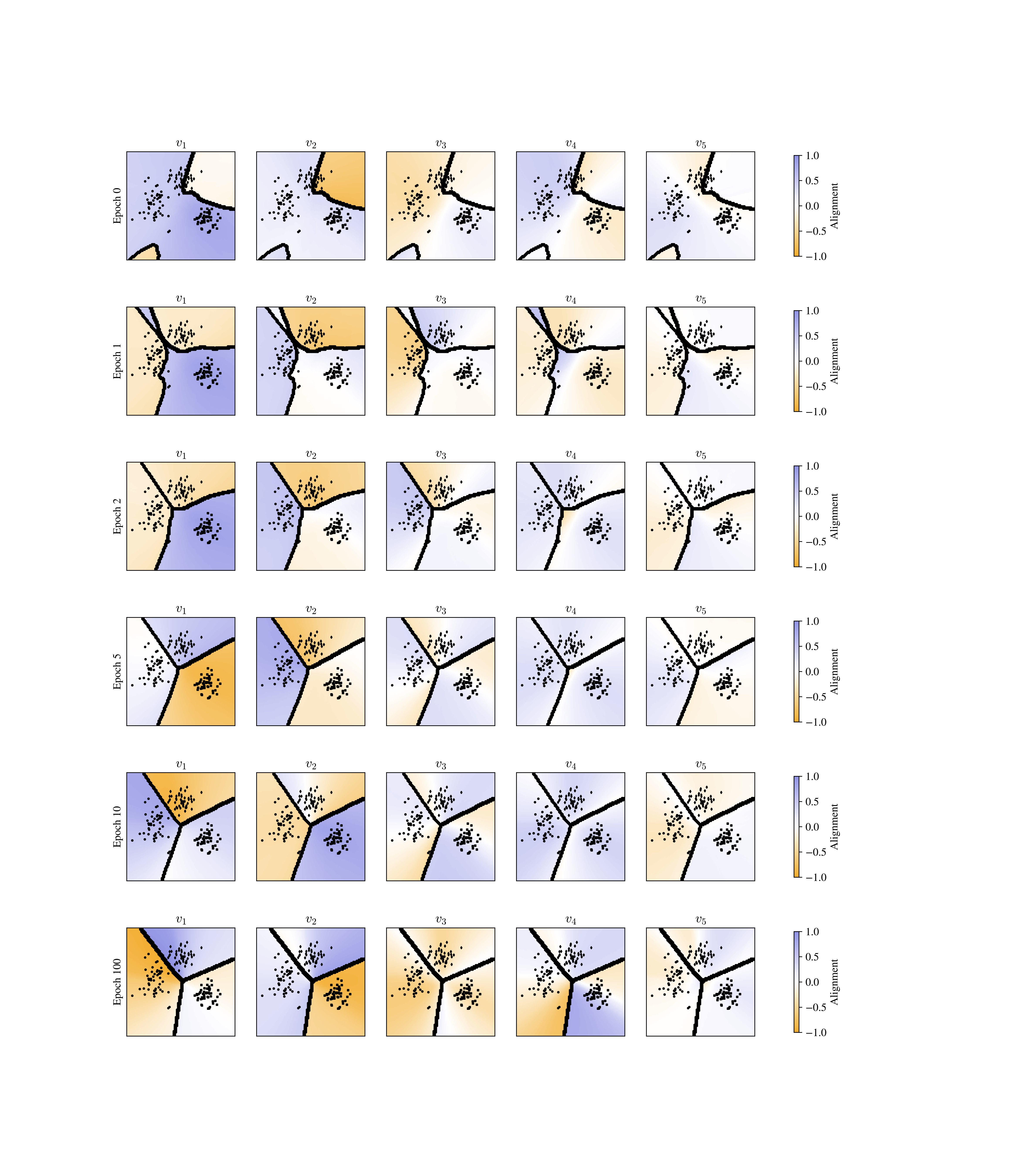}
    \caption{\textbf{Top Hessian eigenvectors encode boundaries also away from the minimum.} The overlap plots during different epochs in for normal training on \emph{gaussian}. Epoch 0 is the boundary at initialization before training.} 
    \label{fig:dynamics}
\end{figure}

\begin{figure}
    \centering
\includegraphics[width=1.1\textwidth]{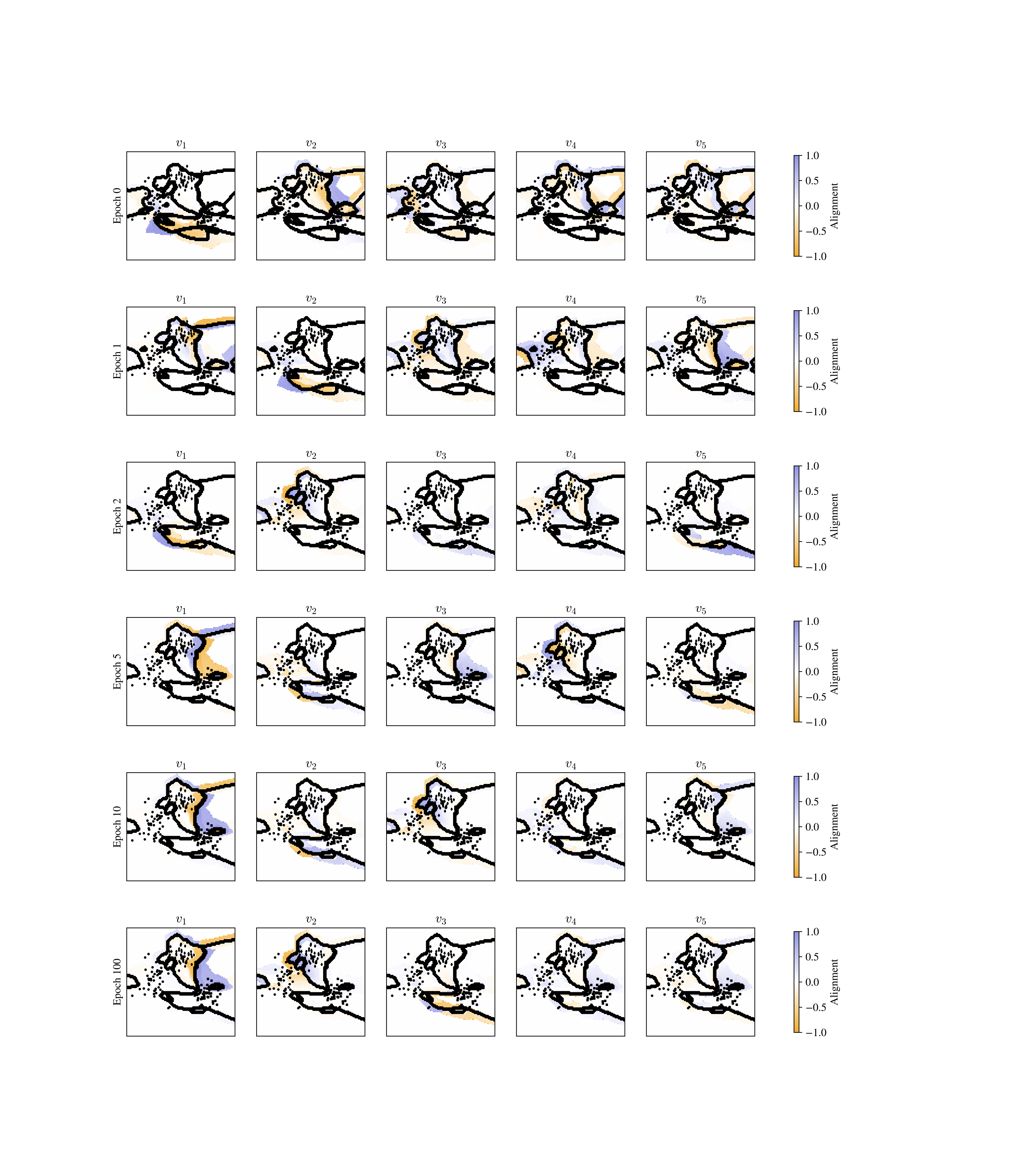}
    \caption{\textbf{Top Hessian eigenvectors encode boundaries also away from the minimum for an adversarial initialization on \emph{gaussian}.} Epoch 0 is the boundary at initialization before training.}
    \label{fig:dynamics2}
\end{figure}

Interestingly, we see that \emph{the top Hessian eigenvectors encode the decision boundary also away from the minimum during the training dynamics.} We believe that the gradient covariance matrix will not exhibit this behavior since it is not at the minimum. We present the usual alignment analysis between gradients of loss of input samples and the top five Hessian eigenvectors for selected epochs of the regular training in Figure~\ref{fig:dynamics} and training starting from the adversarial initialization of \citet{Liu2020badminima} in Figure~\ref{fig:dynamics2}.

\newpage
\clearpage
\section{Generalization measure for all simulated datasets and its limitations}
\label{app:generalization_measure_sim}

In Table~\ref{tab:gen_meas_comp_sim_appendix}, we present values of the generalization measure $\mathcal{G}_{\theta}$ introduced in Equation~\ref{eq:gen_measure} for models trained on various simulated datasets and from different initializations. ``Normal training'' indicates the regular initialization of the neural networks, ``adversarial initialization'' follows the initialization procedure by \citet{Liu2020badminima} that consists in pretraining the model on the same data but with random labels, and the ``large norm training'' means starting from a default random initialization with the imposed large norm, as discussed in Section~\ref{ss:results-complex-many-heigenvectors}. We compare $\mathcal{G}_{\theta}$ with other metrics calculated at the minimum like the Hessian trace, its spectral norm, and $L_2$ norm of the solution. We consistently see that models initialized adversarially or with a large norm learn more complex decision boundaries that generalize worse than the simple boundary learned by regularly trained models. 
\emph{The generalization measure $\mathcal{G}_{\theta}$ successfully distinguishes between those models in a large majority of cases}, and other metrics are unreliable. The analogous results for real datasets like \emph{Iris} and \emph{MNIST} are in Appendix~\ref{app:gen_real}. 

\begin{table}[h]
\centering
\caption{\textbf{Generalization measures comparison for the five simulated 2D datasets.} We provide the mean of those measures and their standard deviation over $5$ runs. A bold font marks the best generalizing minimum according to the studied metric, green (red) color indicates whether the indication is correct (wrong). With yellow, we mark correct indications with standard deviations being larger than the difference of compared means.}
%\small
\resizebox{\linewidth}{!}{
\begin{tabular}{@{}llllll@{}}
\toprule
\multicolumn{1}{c}{\textbf{Dataset}} & \multicolumn{1}{c}{\textbf{Training}} & \multicolumn{1}{c}{\textbf{$\mathcal{G}_\theta \downarrow$}} & \multicolumn{1}{c}{\textbf{$\mathrm{trace}(H) \downarrow$}} & \multicolumn{1}{c}{\textbf{$\lambda_{\max}(H) \downarrow$}} & \multicolumn{1}{c}{\textbf{$|| \theta^* ||_2 \downarrow$}} \\ \midrule  

\multirow{3}{*}{\emph{gaussian}}   
& normal      & $\color{applegreen}{\textbf{0.055}} \pm 0.004$& $0.176 \pm 0.010$ & $0.114 \pm 0.010$ & $\color{applegreen}{\textbf{19.60}} \pm 0.15$ \\
 & adversarial &$0.156 \pm 0.035$&$\color{brightmaroon}{\textbf{0.003}} \pm 0.001 $&$ \color{brightmaroon}{\textbf{0.002}}\pm 0.001$&$ 105.00 \pm 0.005$             \\
& large norm &$ 0.114 \pm 0.040$&$0.021\pm 0.018$&$ 0.017\pm 0.012$&$ 98.169 \pm 0.413$\\ \midrule 

\multirow{3}{*}{\emph{circle}}   
& normal      & $ \color{applegreen}{\textbf{0.044}} \pm 0.007$& $8.028 \pm 0.777$ & $4.965 \pm 0.514$ & $\color{applegreen}{\textbf{22.884}} \pm 0.139$ \\
 & adversarial &$0.059 \pm 0.003$&$\color{brightmaroon}{\textbf{0.795}} \pm 0.051 $&$ \color{brightmaroon}{\textbf{0.439}}\pm 0.037$&$ 41.630 \pm 0.006$             \\
& large norm &$ 0.057 \pm 0.003$ &$6.320\pm 0.833$&$ 4.350\pm 0.735$&$ 41.840 \pm 0.226$\\ \midrule 

\multirow{3}{*}{\emph{half-moon}}   
& normal      & $ \color{applegreen}{\textbf{0.036 }} \pm 0.003$& $4.202 \pm 0.637$ & $2.958 \pm 0.479$ & $\color{applegreen}{\textbf{21.529}} \pm 0.285$ \\
 & adversarial &$0.072 \pm 0.006$&$\color{brightmaroon}{\textbf{0.037}} \pm 0.001 $&$ \color{brightmaroon}{\textbf{0.017}}\pm 0.001$&$ 68.988 \pm 0.009$             \\
& large norm &$ 0.042 \pm 0.004$ &$1.119\pm 0.563$&$ 0.868\pm 0.431$&$ 64.807 \pm 0.201$\\ \midrule 

\multirow{3}{*}{\emph{hierarchical}}   
& normal      & $ \color{warningyellow}{\textbf{0.053}} \pm 0.001$& $12.450 \pm 0.595$ & $7.102 \pm 0.189$ & $\color{applegreen}{\textbf{20.034}} \pm 0.202$ \\

 & adversarial &$0.118 \pm 0.024$&$\color{brightmaroon}{\textbf{3.394}} \pm 1.035 $&$ \color{brightmaroon}{\textbf{2.645}}\pm 0.877$&$ 121.675 \pm 0.062$             \\
 
& large norm &$ 0.059 \pm 0.009$ &$93.104\pm 12.985$&$ 36.787\pm 3.936$&$ 112.579 \pm 0.404$\\ \midrule

\multirow{2}{*}{\emph{checkerboard}}   
& narrow-margin      & $ 0.046 \pm 0.005 $& $0.240 \pm 0.050$ & $0.127 \pm 0.028$ & $\color{brightmaroon}{\textbf{19.267}} \pm 0.097$ \\
 & wide-margin$^{a}$ & $ \color{warningyellow}{\textbf{0.043}} \pm 0.001$ &\color{applegreen}{$\textbf{0.029} \pm 0.000 $}&$ \color{applegreen}{\textbf{0.014}}\pm 0.000$&$ 19.910 \pm 0.000$             \\\bottomrule

\end{tabular}}
\label{tab:gen_meas_comp_sim_appendix}
\vspace{-1.4em}
\end{table}
{\scriptsize ${}^{a}$The standard deviation is almost $0$ since we initialize models across runs with the same pretrained solution to promote a wide margin.}
\vspace{0.5em}

The first case where $\mathcal{G}_{\theta}$ gives ambiguous results is distinguishing between the narrow- and wide-margin minima as already discussed in Section~\ref{ss:results-simplicity-bias}. There, the $\mathcal{G}_{\theta}$ correctly indicated that both minima have similarly simple decision boundaries. The difference in the margin width can be detected with the margin width estimation technique proposed in Section~\ref{ss:results-simplicity-bias}.

\begin{figure}[b]
    \centering
    \includegraphics[width=0.82\textwidth]{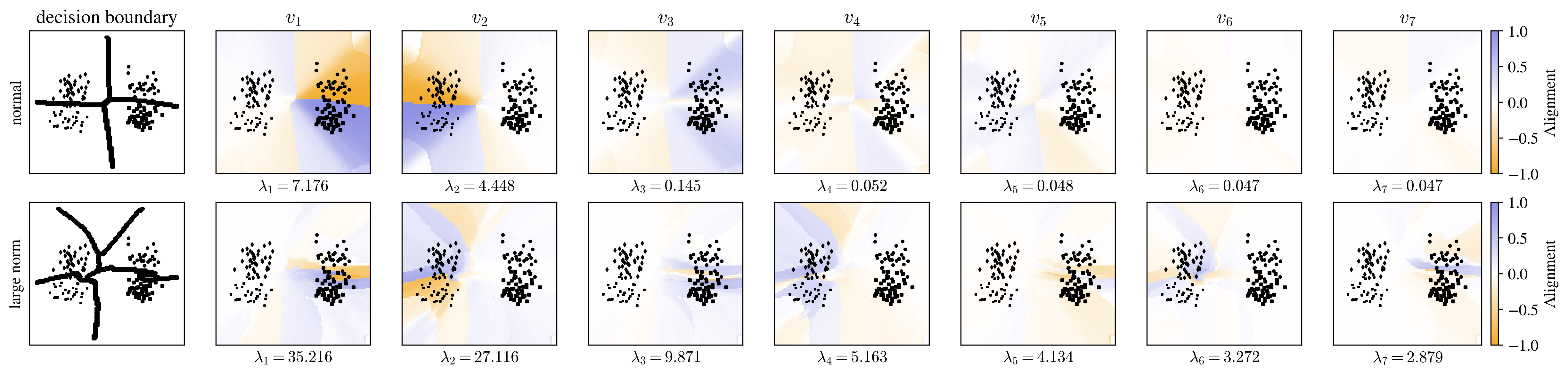}
    \includegraphics[width=0.139\textwidth]
    {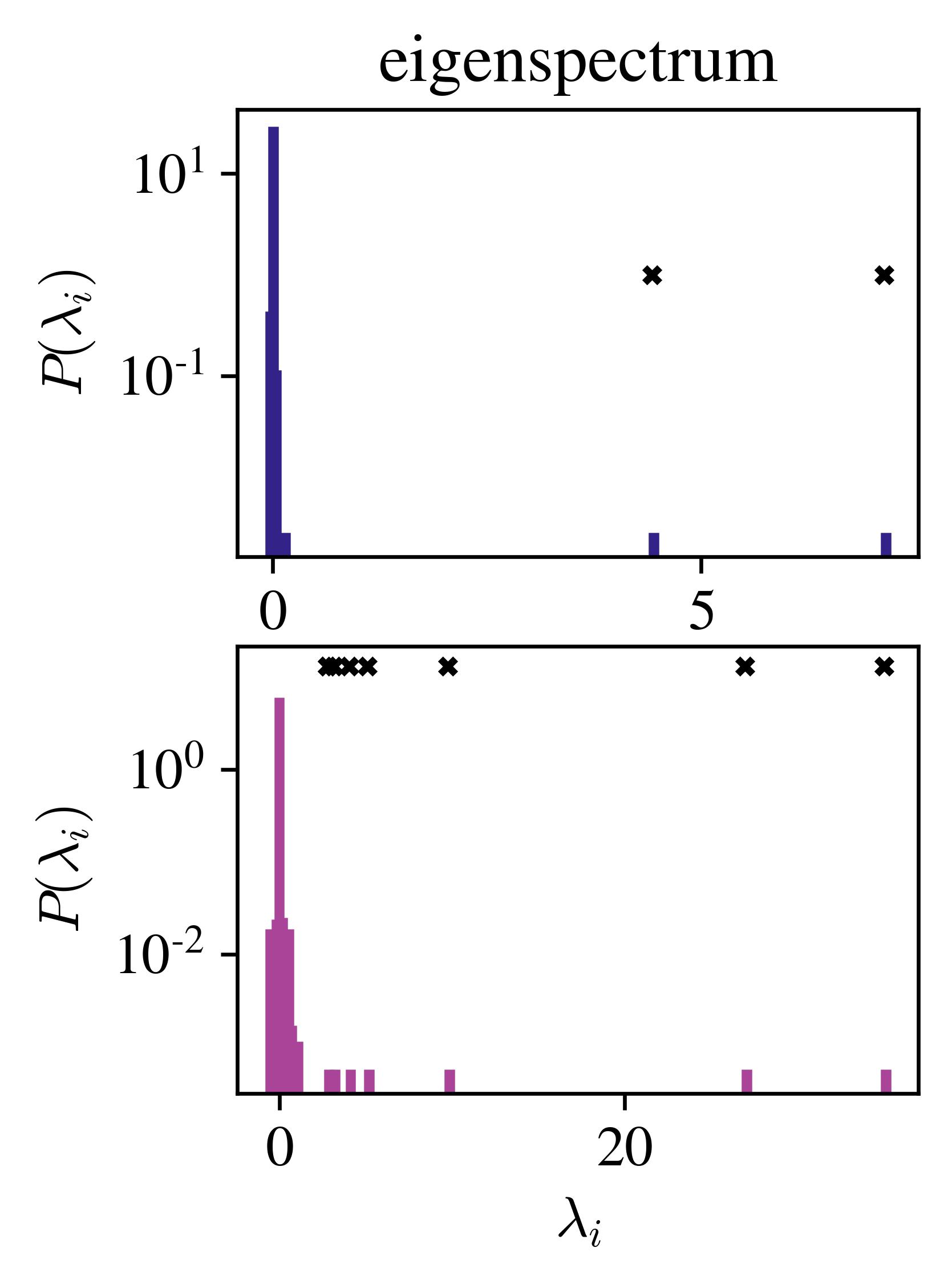}
    \caption{\textbf{Decision boundaries of different complexities for \emph{hierarchical gaussian}.} Alignment plots and histograms of the Hessian spectra for models obtained from normal training and a large norm initialization.}
\label{fig:bad_minima_hierarchical}
\end{figure}

\begin{figure}[t]
    \centering
    \includegraphics[width=\textwidth]{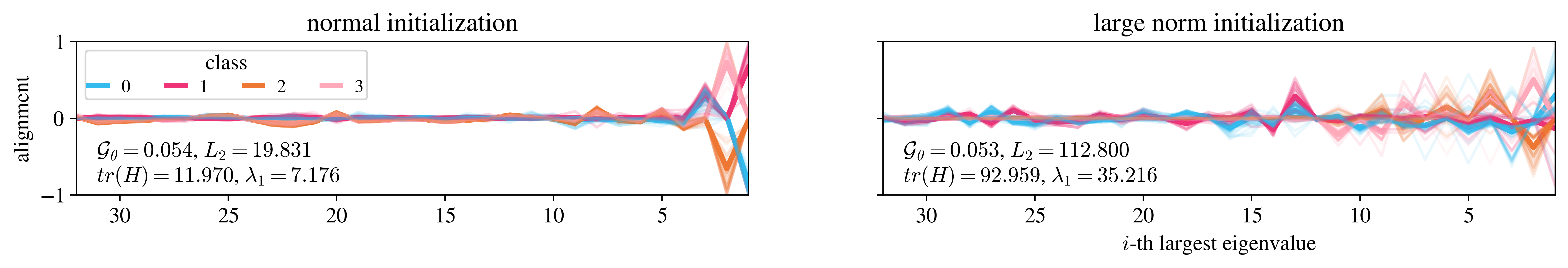}
    \caption{\textbf{Alignment of all training data with the top $25$ eigenvectors for \emph{hierarchical gaussian}} for models obtained from the \emph{(Left)} normal training and the\emph{(Right)} large norm initialization. There are four classes $\{0,1,2,3\}$. The dark lines show the mean of each class alignment.}
    \label{fig:hierarchical_alignment}
\end{figure}

The second case of ambiguous results takes place when distinguishing between the minima obtained with the normal and large norm training for \emph{hierarchical gaussian}, marked in yellow in Table~\ref{tab:gen_meas_comp_sim_appendix}. While on average $\mathcal{G}_{\theta}$ successfully indicates that minima obtained with normal initialization have simpler decision boundaries than those obtained with the large norm initialization, the standard deviation exceeds the difference between the means. We identify a single case where our $\mathcal{G}_{\theta}$ fails in distinguishing minima with different complexities of decision boundaries and make a full analysis of the alignment in Figures~\ref{fig:bad_minima_hierarchical} and~\ref{fig:hierarchical_alignment} for the \emph{hierarchical gaussian} with four classes. Firstly, we see from the first column of Figure~\ref{fig:bad_minima_hierarchical} that the complexity of the decision boundary increases in the region with a small number of training samples. It indicates a limitation of our measure that is based on the ``interaction'' between the training samples and neighboring decision boundary. If the decision boundary is simple close to the training samples but complex away from them, $\mathcal{G}_{\theta}$ may struggle in detecting this. Secondly, the number of outliers in the Hessian spectrum in the large norm case remains larger than in the normal case as visible in the last column of Figure~\ref{fig:bad_minima_hierarchical}. Finally, we take a closer look at the alignment of the top Hessian eigenvectors and gradients of loss of training samples in Figure~\ref{fig:hierarchical_alignment} at the minima studied in Figure~\ref{fig:bad_minima_hierarchical}. We still see that for simpler decision boundaries the alignment of the training gradients localizes much more in the top Hessian subspace than for the complex boundaries. At the same time, $\mathcal{G}_{\theta}$ is almost the same for both cases, meaning it is an imperfect measure for the gradient alignment localization that seems to be a prevailing characteristic of minima with simple decision boundaries. We leave improvement of this scalar measure for further study. At the same time, we stress that $\mathcal{G}_{\theta}$ has correctly distinguished between minima with simple and complex decision boundaries from adversarial initializations in all the studied cases, and the ambiguity arises only in the large norm initializations.

\begin{figure}[b]
    \centering
\includegraphics[width=0.7\linewidth]{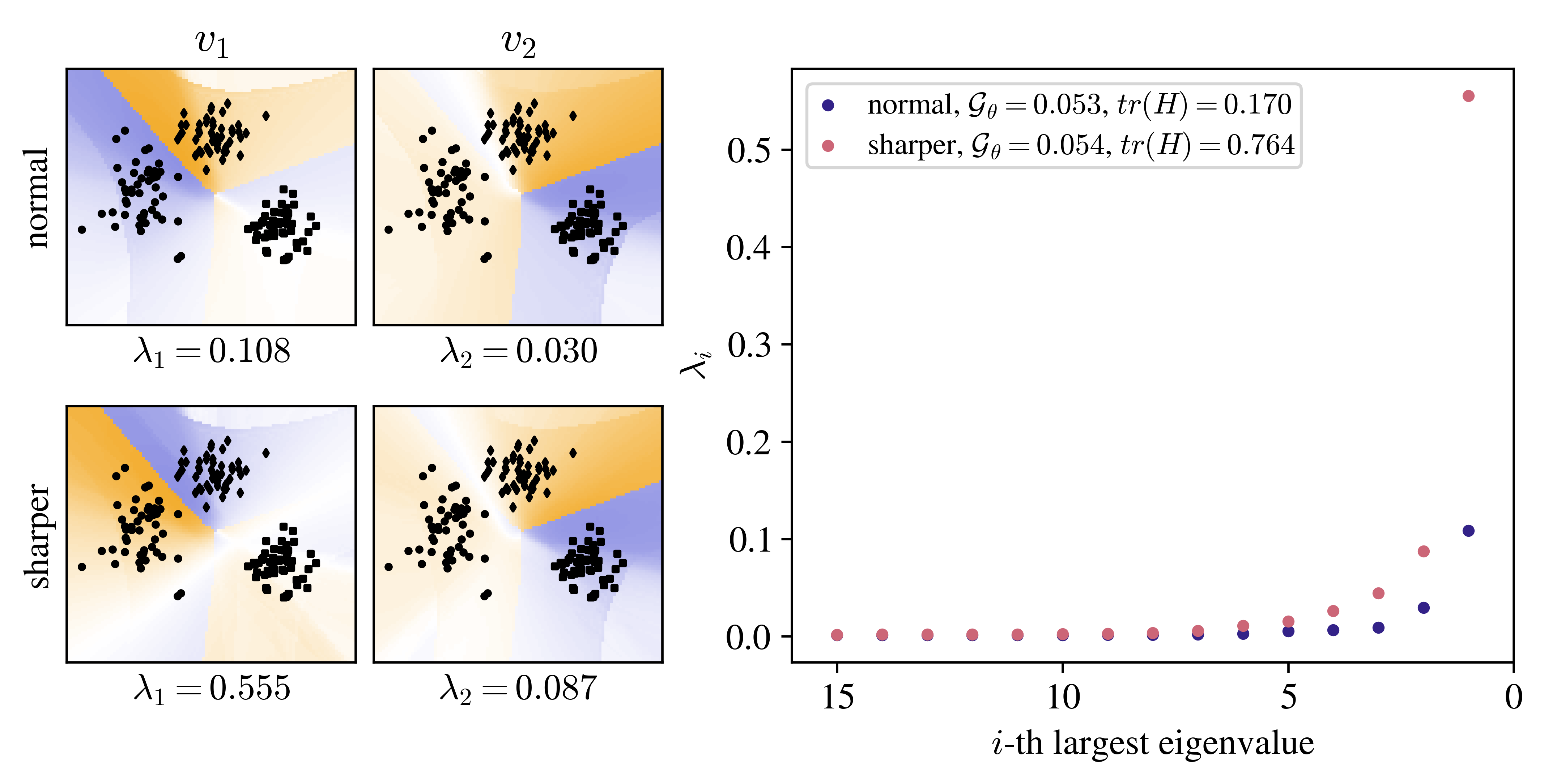}
    %\vspace{-2em}
    \caption{\textbf{Reparameterization.} A sharp $\alpha$-scale transformation for a 2-layer ReLU-network that rescales the weights and biases of the original model according to \citep[App. B]{Dinh2017} while keeping the predictions identical. \emph{(Left)} The alignments for the top eigenvectors. \emph{(Right)} The spectra for both parameterizations.}
\label{fig:reparameterization}
\end{figure}

\section{Generalization measure is invariant to model reparameterization}\label{app:reparameterization}
A natural assumption is that if a model after a reparameterization yields the same output as the original one, their generalization abilities (and measures) should also be equal.
This is not the case for metrics based on Hessian trace and ReLU networks; One can ``artificially'' sharpen a minimum while retaining the predictions of the original model using the $\alpha$-scale transformation proposed by \citet{Dinh2017}.
In Figure~\ref{fig:reparameterization}, we see that while such a reparameterization indeed affects the Hessian spectrum, it does not impact the connection between the top Hessian eigenvectors and decision boundary. As a result, \emph{our generalization metric is also invariant to the reparameterization} as it is based on the simplicity of the decision boundary that stays the same. At the same time, we see that the reparameterization may change the sign and values of the alignment. 
It is yet unclear why this happens, but this may further suggest that the exact values of the alignment are not informative.

%\newpage
\section{Hessian analysis for \emph{Iris}}\label{app:Iris}
We have conducted our Hessian analysis for 2D datasets enabling straightforward visualization of the learned decision boundary. This approach has enabled a clear visual distinction between a simple or complex decision boundaries. Such a visual distinction is much needed in view of a limited (to our knowledge) theoretical description of the complexity of the decision boundary. At the same time, visualization of the decision boundary is hardly possible for high-dimensional datasets. 

\begin{figure}[t]
    \centering
    \includegraphics[width=\textwidth]{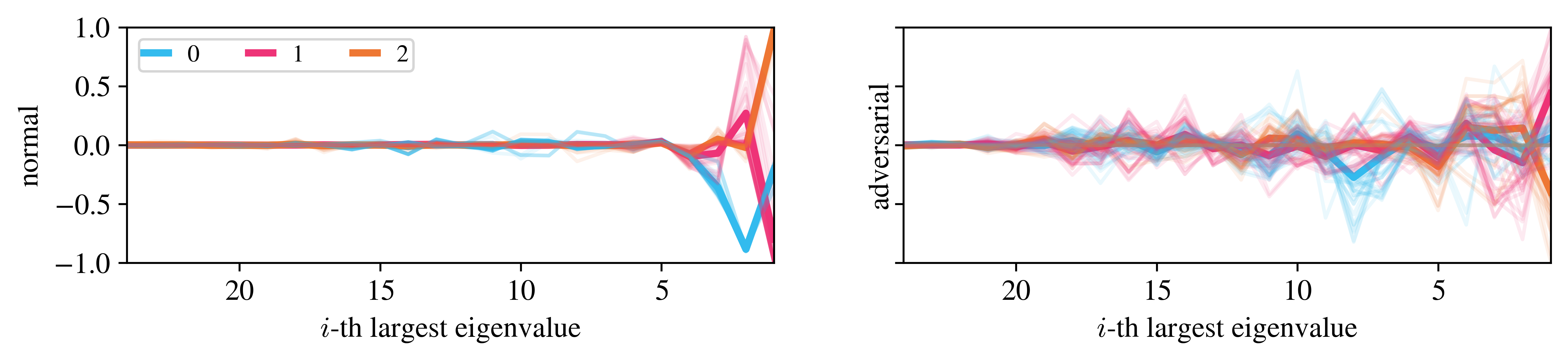}
    \caption{\textbf{Alignment of all training data with the top $25$ Hessian eigenvectors for \emph{Iris}.} \emph{(Left)} Normal training. \emph{(Right)} Adversarial initialization. There are three classes $\{0,1,2\}$. The dark lines show the mean of each class alignment.}
    \label{fig:iris_overlap}
\end{figure}

\begin{figure}[b]
    \centering
    \includegraphics[width=\textwidth]{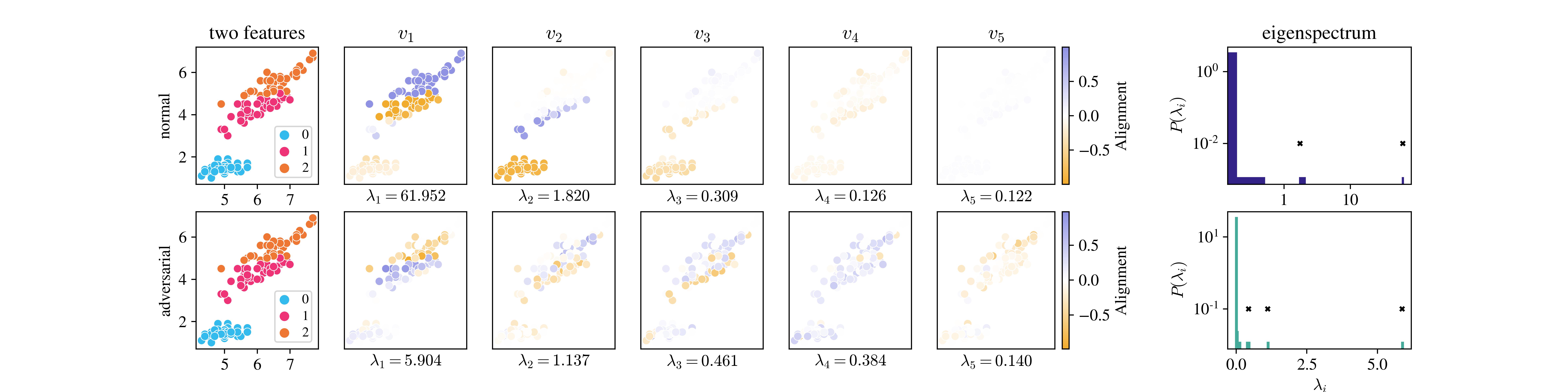}
    \caption{\textbf{Experimental results on \emph{Iris}.} \emph{(First column)} Two features (petal and sepal length) out of four of the Iris dataset with color-coded classes. \emph{(Other columns)} The alignment of gradients of the loss of individual training samples with the top five Hessian eigenvectors. \emph{(Last column)} Histograms of the Hessian spectra. \emph{(Top)} Well-generalizing minimum obtained with normal training. \emph{(Bottom)} Badly generalizing minimum obtained with an adversarial initialization \citep{Liu2020badminima}.}
    \label{fig:iris_overlap_decision_boundary}
\end{figure}

Here, we extend our analysis to real datasets, that is to \emph{Iris} dataset in this section and \emph{MNIST}-based datasets in Appendix~\ref{app:MNIST}. Firstly, we show that \emph{with our Hessian analysis, we distinguish between well- and badly generalizing minima in realistic deep learning setups}. To do so, we compare the models trained with a regular initialization and the adversarial initialization \citep{Liu2020badminima}, which are believed to reach a well- and badly generalizing minimum, respectively. For \emph{Iris}, we show the corresponding Hessian spectra in the last column of Figure~\ref{fig:iris_overlap_decision_boundary} and alignments of gradients of loss of individual training samples with the Hessian eigenvectors in Figure~\ref{fig:iris_overlap}, respectively. We again see the larger number of outliers in the spectra in the case of more complex decision boundary. Most importantly, we see that the gradients have non-zero alignment with a much smaller number of Hessian eigenvectors in the case of normal training than in the adversarial case (Figure~\ref{fig:iris_overlap}). We again see that the gradients are more aligned with each other in the well generalizing than badly generalizing minimum, as observed during the training dynamics in \citet{Chatterjee2022coherentgrads}. The generalization metric $\mathcal{G}_{\theta}$ captures this difference as expected (is lower for well generalizing minimum) and is listed along with the \emph{MNIST} results in Table~\ref{tab:generalization_measure_real} in Appendix~\ref{app:gen_real}.

Moreover, our low-dimensional analysis in the main body shows that the drastically different behavior of training gradients alignment with the Hessian eigenvectors results from a different complexity of the decision boundary. In other words, training gradients align with a larger number of the top Hessian eigenvectors because around training samples in input space, there are numerous sections of the decision boundary encoded in multiple directions in parameter space. While we could make this connection clear in the case of 2D datasets, it is more challenging for four dimensions and impractical for significantly larger dimensions. For Irises, we instead visualize samples by selecting only two features out of four and without the decision boundaries. Then we color code the alignment of the gradient of their individual losses at the minimum with the top Hessian eigenvectors. We present the normal and adversarial training results in Figure~\ref{fig:iris_overlap_decision_boundary}. We can see a clearly different behavior of the alignment between the well- and badly-generalizing minimum. This suggests a different complexity of the decision boundary following results from the low-dimensional data.

\section{Hessian analysis for \emph{MNIST}}\label{app:MNIST}

Finally, we make an analogous Hessian analysis for the \emph{MNIST}-based datasets. To decrease the complexity of the dataset and better understand the dependence of the results on the number of classes, we create four subsets of \emph{MNIST}: \emph{MNIST-017}, \emph{MNIST-179}, \emph{MNIST-0179}, and \emph{MNIST-1379}, where numbers indicate selected classes of digits. Each class has a few hundred samples sampled randomly from the \emph{MNIST} dataset.

\begin{figure}[b]
    \centering
    \subfigure{\includegraphics[width=0.48\linewidth]{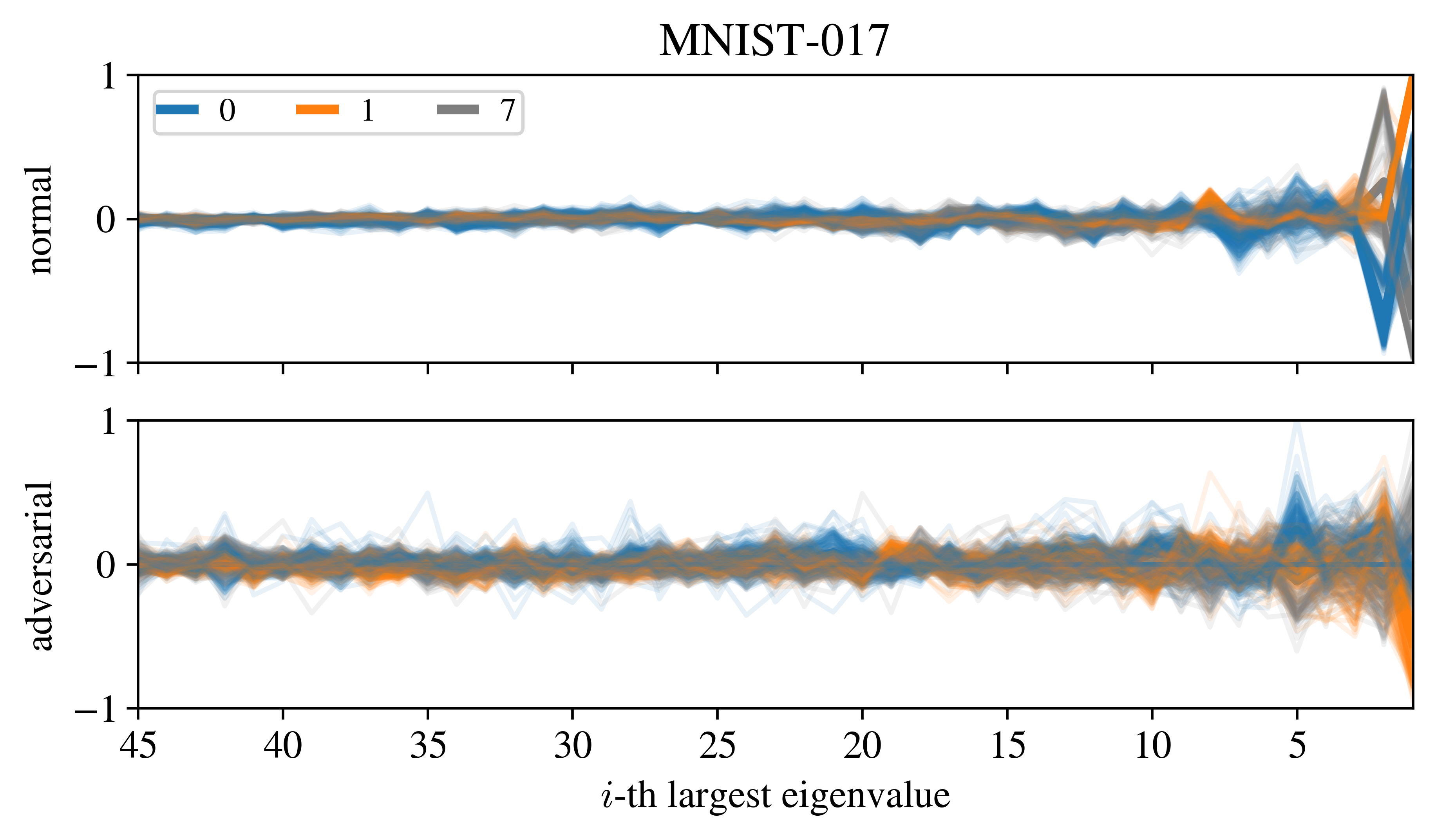}}
    \subfigure{\includegraphics[width=0.48\linewidth]{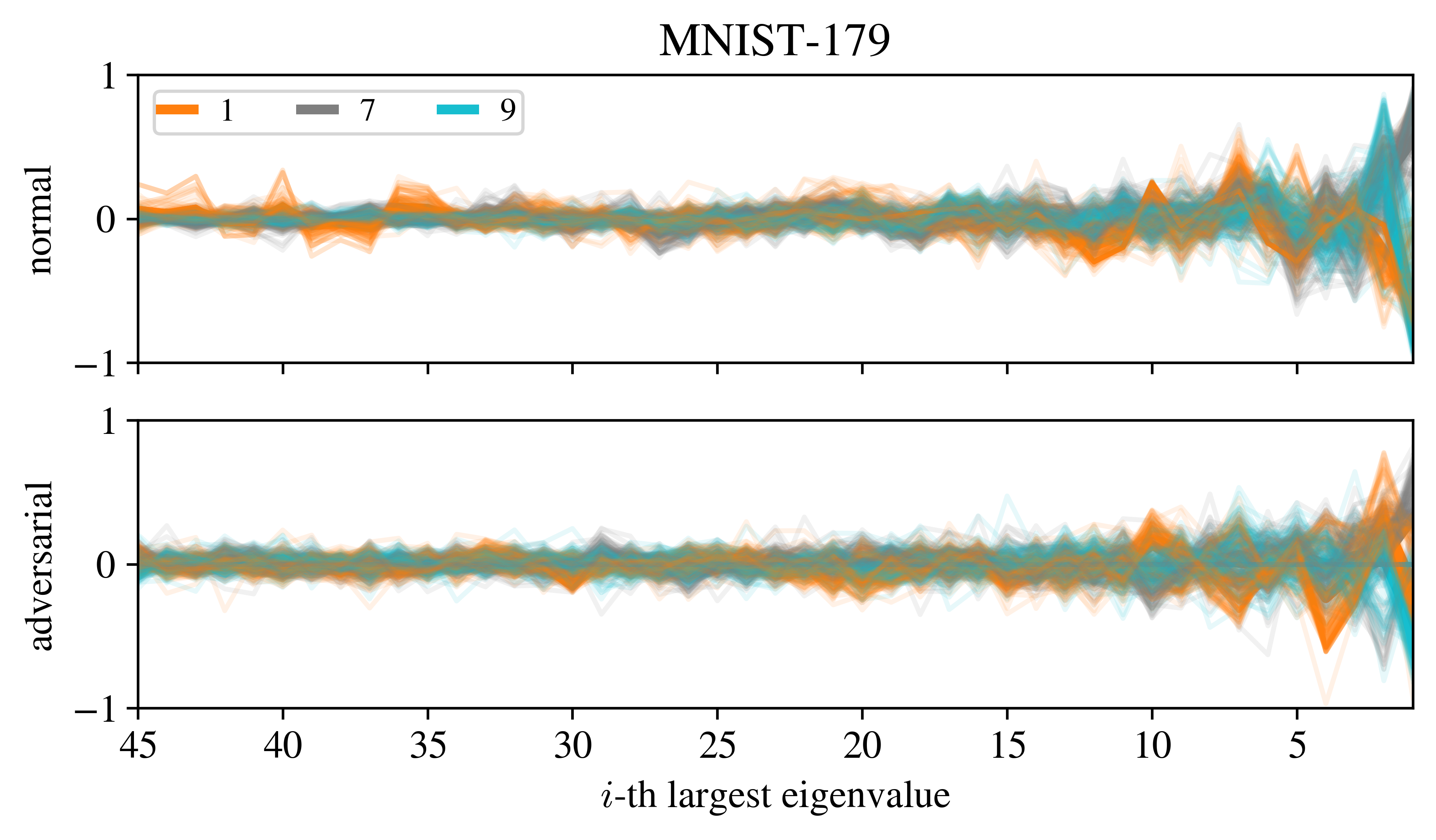}}
    \subfigure{\includegraphics[width=0.48\linewidth]{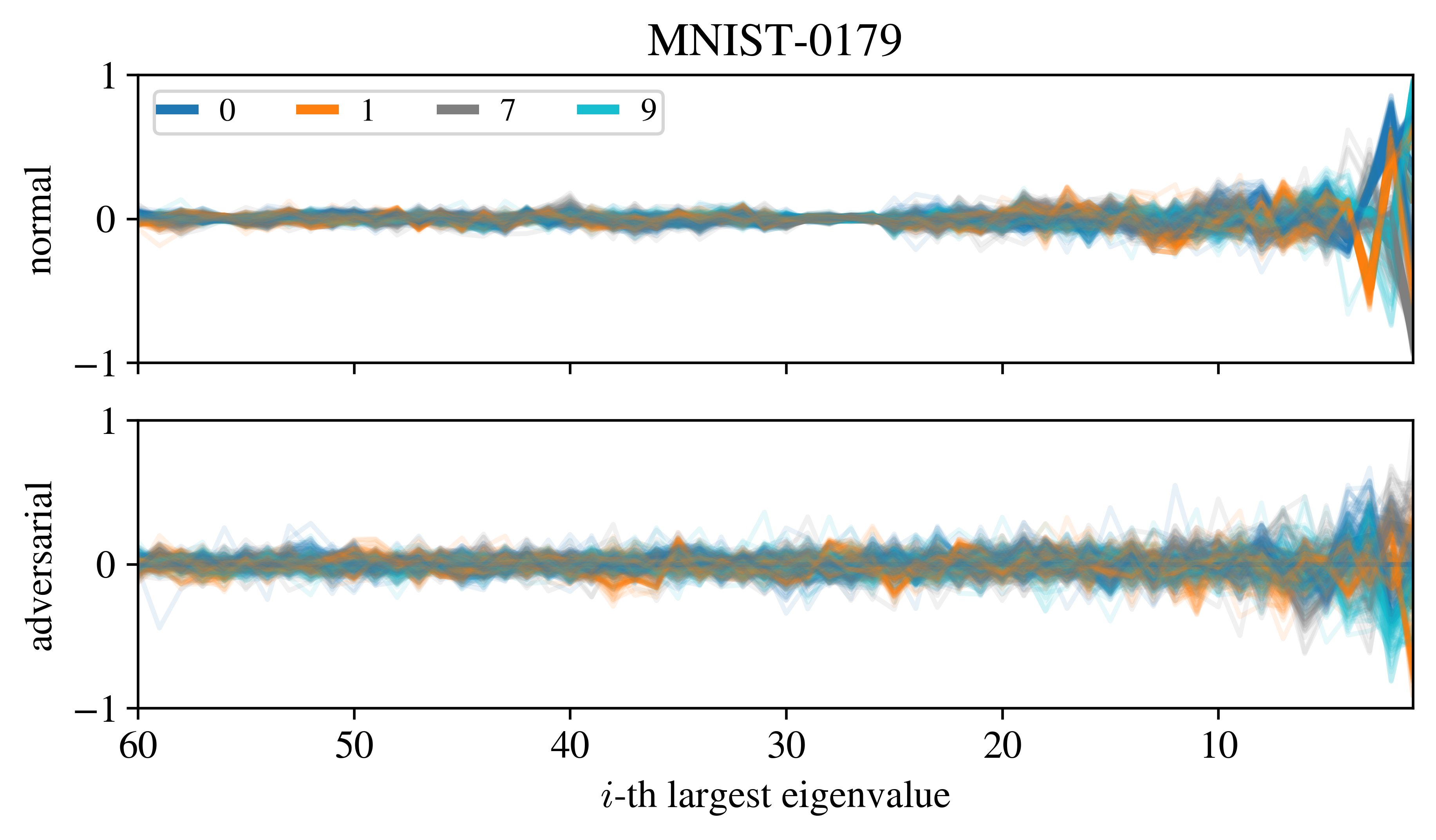}}
    \subfigure{\includegraphics[width=0.48\linewidth]{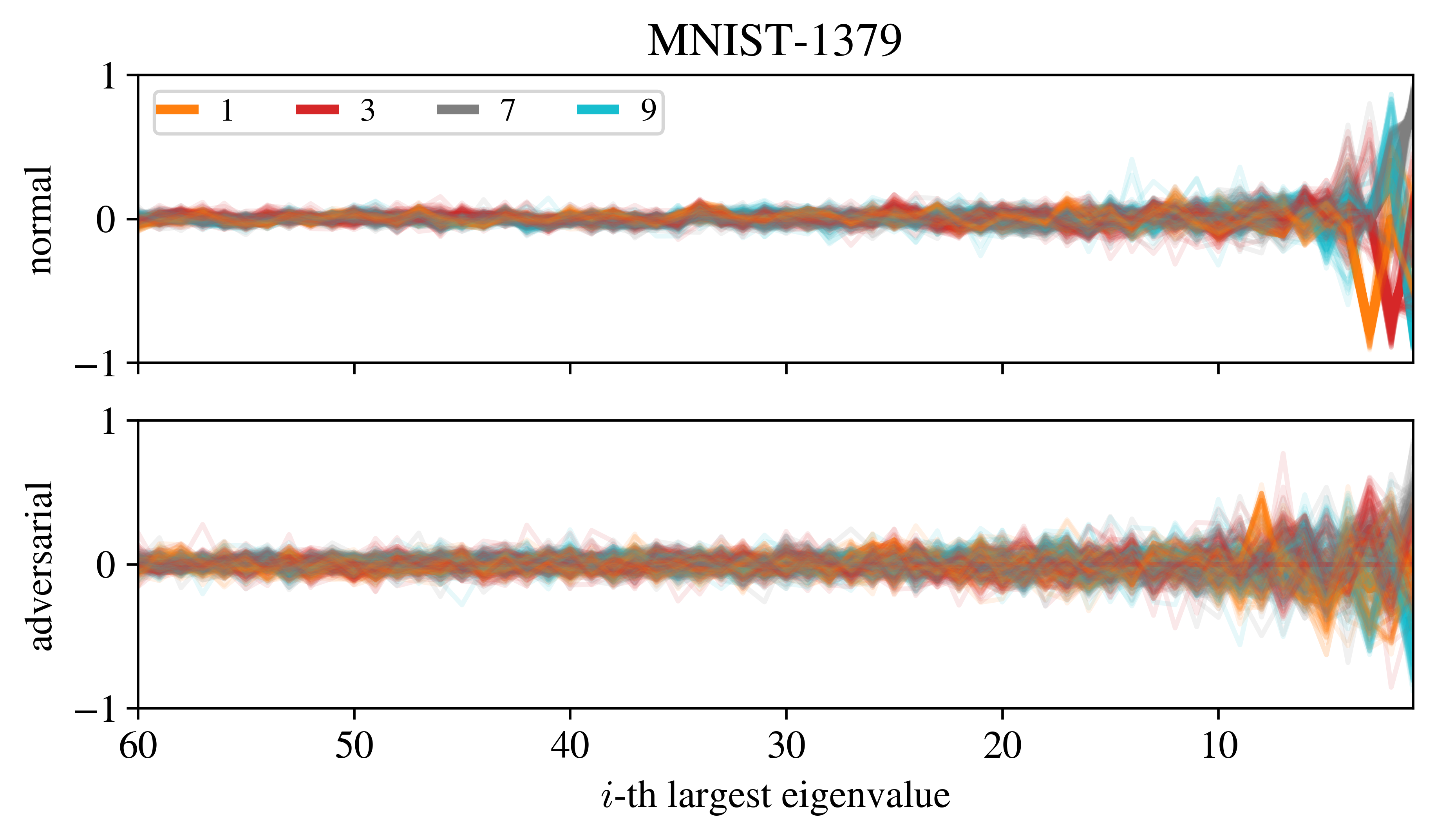}}

    \caption{\textbf{Normal and adversarial initialization training for \emph{MNIST-017}, \emph{MNIST-179}, \emph{MNIST-0179}, and \emph{MNIST-1379}}. We plot the alignments of gradients of all training samples onto all eigenvectors ordered by their eigenvalues. Only the largest eigenvectors have non-zero alignment with the gradients of training samples, and their number increases for the training from the adversarial initialization.}
    \label{fig:MNIST_overlaps}
\end{figure}

The analysis of the alignment of the gradients of loss of individual training samples and the top Hessian eigenvectors is presented in Figure~\ref{fig:MNIST_overlaps}. We continue to see that the alignment for the regular training is more localized in the space spanned by the top Hessian eigenvectors. We also see self-alignment of the gradients \citep{Chatterjee2022coherentgrads} that maybe stops being so apparent in the top few eigenvectors.

\begin{figure}
    \centering
    \includegraphics[width=\textwidth]{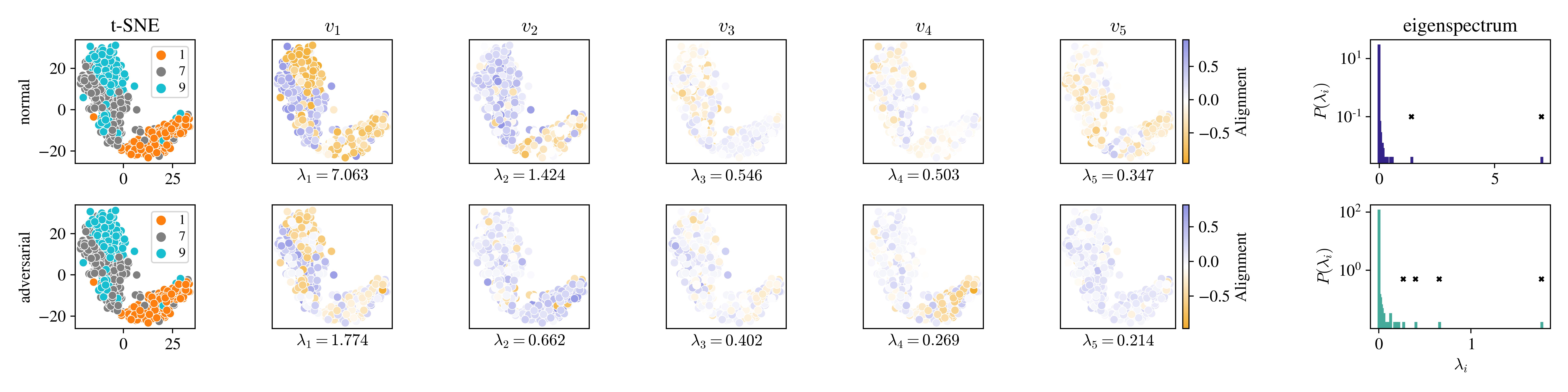} \\
    \includegraphics[width=\textwidth]{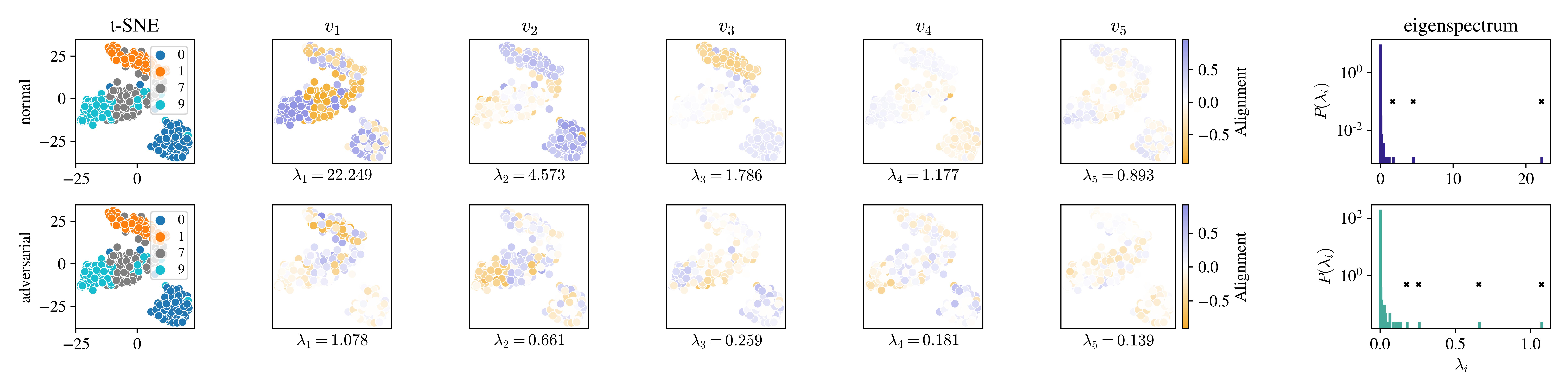} \\
    \includegraphics[width=\textwidth]{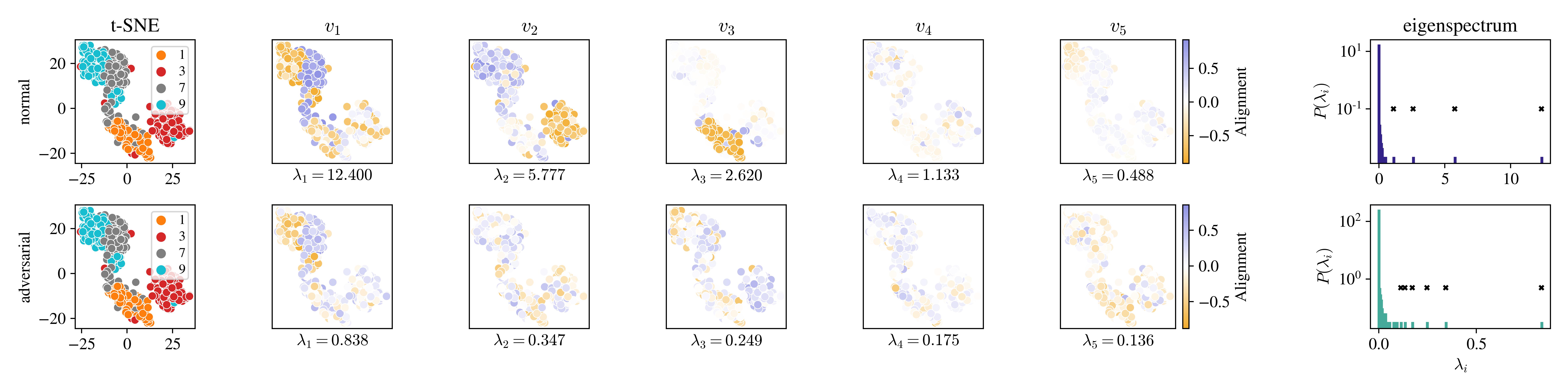}
    \caption{\textbf{Normal and adversarial initializations training for \emph{MNIST-179}, \emph{MNIST-0179}, and \emph{MNIST-1379} with t-SNE visualization and Hessian eigenspectra}. We visualize the \emph{MNIST}-based datasets with t-SNE and color code the alignments of gradients of all training samples onto all eigenvectors ordered by their eigenvalues. Multiple eigenvectors have non-zero alignment with the gradients of training samples, and there is little ordering of the samples' colors suggesting complex decision boundaries. \textit{(Last column)} Hessian eigenspectra.}
    \label{fig:mnist_tsne_spectrum}
\end{figure}

Moreover, as we mentioned in Appendix~\ref{app:Iris}, while we could make a clear connection between Hessian-gradient alignment and complexity of decision boundary in the case of 2D datasets, such a visualization is impractical for high input dimensions. Instead, we make the following non-rigorous analysis. We visualize the high-dimensional \emph{MNIST} samples in a 2D plot using t-distributed stochastic neighbor embedding (t-SNE) and then color code the alignment of the gradient of their individual losses at the minimum. We present the normal and adversarial training results in Figure~\ref{fig:mnist_tsne_spectrum}. Even if there is no guarantee that the neural network representation of the data is related to the one obtained by t-SNE nor that the learned decision boundary in the input space corresponds simply to the boundaries between t-SNE generated clusters, we still can see a clearly different behavior of the alignment between the well- and badly-generalizing minimum.

Finally, the Hessian spectra for all \emph{MNIST}-based datasets are in the last column of Figure~\ref{fig:mnist_tsne_spectrum}. We consistently see that the number of outliers increases for the badly generalizing minima. We also see that metrics like the Hessian trace or its largest eigenvalue fail to capture the difference between the minima's generalizing abilities. On the other hand, our generalization measure, $\mathcal{G}_{\theta}$, consistently provides correct indications. We make this comparison apparent in Table~\ref{tab:generalization_measure_real} in Appendix~\ref{app:gen_real}.

\newpage
\clearpage
\section{Generalization measure for \emph{Iris} and \emph{MNIST}}\label{app:gen_real}
For the convenience of the reader, in Table~\ref{tab:generalization_measure_real}, we present values of the generalization metric $\mathcal{G}_{\theta}$ introduced in Equation~\ref{eq:gen_measure} for models trained on various real datasets and from different initializations. ``Normal training'' indicates the regular initialization of the neural networks. Adversarial initialization follows the initialization by \citet{Liu2020badminima} that consists in pretraining on the same data but with random labels, as discussed in Section~\ref{ss:results-complex-many-heigenvectors}. In the low-dimensional datasets, we consistently see that models initialized adversarially learn more complex decision boundaries that generalize worse than the simple boundary learned by regularly trained models. As we cannot visualize the decision boundary for the high-dimensional data, we skip the analysis of the large-norm initialization here. Instead, we use only the established adversarial initialization by \citet{Liu2020badminima}, which has been shown to produce complex boundaries and badly generalizing minima.
The generalization metric successfully distinguishes between those models. The analogous results for simulated 2D datasets are in Appendix~\ref{app:generalization_measure_sim}.

\begin{table}
\caption{\textbf{Generalization measure comparison for real datasets \emph{Iris} and different subsets of \emph{MNIST} under different initializations.} We provide the mean of those measures and their standard deviation over $5$ runs. A bold font marks the best generalizing minimum according to the studied metric, green (red) color indicates whether the indication is correct (wrong).}
%\small
\resizebox{\linewidth}{!}{
\begin{tabular}{@{}llllll@{}}
\toprule
\multicolumn{1}{c}{\textbf{Dataset}} & \multicolumn{1}{c}{\textbf{Training}} & \multicolumn{1}{c}{\textbf{$\mathcal{G}_\theta \downarrow$}} & \multicolumn{1}{c}{\textbf{$\mathrm{trace}(H) \downarrow$}} & \multicolumn{1}{c}{\textbf{$\lambda_{\max}(H) \downarrow$}} & \multicolumn{1}{c}{\textbf{$|| \theta^* ||_2 \downarrow$}} \\ \midrule 
\multirow{2}{*}{\emph{Iris}}       & normal      & $\color{applegreen}{\textbf{0.031} {  \pm 0.006}}$                & $67.857 { \pm 5.943}$  & $65.005 { \pm 6.072}$    & $\color{applegreen}\textbf{13.998} { \pm 0.221}$      \\
           & adversarial & $0.094 { \pm 0.002}$                & $\color{brightmaroon}\textbf{8.324} { \pm 0.235}$  & $\color{brightmaroon}\textbf{5.934} { \pm 0.067}$    & $68.361 { \pm 2.040}$      \\ \midrule
\multirow{2}{*}{\emph{MNIST-017}}  & normal      & $\color{applegreen}\textbf{0.037} { \pm 0.028}$                & $6.288 { \pm 4.697}$   & $3.758 { \pm 3.215}$     & $2237.6 { \pm 3526.8}$ \\
           & adversarial & $0.109 { \pm 0.002}$                & $\color{brightmaroon}\textbf{0.945} { \pm 0.117}$   & $\color{brightmaroon}\textbf{0.479} { \pm 0.097}$     & $\color{brightmaroon}\textbf{938.38} { \pm 0.03} $    \\ \midrule
\multirow{2}{*}{\emph{MNIST-179}}  & normal      & $\color{applegreen}\textbf{0.045} { \pm 0.063}$                & $14.714 { \pm 14.783}$ & $8.270 { \pm 8.262}$     & $731.65 { \pm 716.37}$   \\
           & adversarial & $0.209 { \pm 0.006}$                & $\color{brightmaroon}\textbf{5.222} { \pm 0.740}$   & $\color{brightmaroon}\textbf{1.611} { \pm 0.271}$     & $\color{brightmaroon}\textbf{268.65} { \pm 0.06}$     \\ \midrule
\multirow{2}{*}{\emph{MNIST-0179}} & normal      & $\color{applegreen}\textbf{0.043} { \pm 0.010}$                & $28.472 { \pm 9.840}$  & $13.254 { \pm 6.312}$    & $387.54 { \pm 129.66}$   \\
           & adversarial & $0.110 { \pm 0.003}$                & $\color{brightmaroon}\textbf{3.180} { \pm 0.399}$   & $\color{brightmaroon}\textbf{0.946} { \pm 0.203}$     & $\color{brightmaroon}\textbf{209.41} { \pm 0.01}$     \\ \midrule
\multirow{2}{*}{\emph{MNIST-1379}} & normal      & $\color{applegreen}\textbf{0.077} { \pm 0.031}$                & $18.380 { \pm 16.588}$ & $6.868 { \pm 6.368}$     & $\color{applegreen}\textbf{249.63} { \pm 135.32}$   \\
           & adversarial & $0.135 { \pm 0.015}$                & $\color{brightmaroon}\textbf{2.938} { \pm 0.164}$   & $\color{brightmaroon}\textbf{0.844} { \pm 0.029}$     & $398.19 { \pm 0.17}$   \\\bottomrule 
\end{tabular}
}
\label{tab:generalization_measure_real}
\end{table}

%\section{Sparsity of the top eigenvector depends on the non-linearity of the encoded boundary}\label{app:sparsity}
%Often eigenvectors encoding linear boundaries are more sparse in terms of parameter contributions than those encoding highly non-linear boundaries. Figure~\ref{fig:top_eigenvector_sparsity} presents a comparison between the normal training, adversarial training, and random label training, leading to increasingly non-linear decision boundaries. Top eigenvectors are sparser in the case of the normal training. This observation holds very often across setups.
%\begin{figure}
%\vskip 0.2in
%    \centering
%    \includegraphics[width=\textwidth]{figures/sparsity_of_largest_eigenvectors.png}
%    \caption{Normal / adversarial / random training, 2D Gausses. The largest eigenvector in the normal training is the sparsest compared to the adversarial and random training.}
%    \label{fig:top_eigenvector_sparsity}
%\vskip -0.2in
%\end{figure}

%%%%%%%%%%%%%%%%%%%%%%%%%%%%%%%%%%%%%%%%%%%%%%%%%%%%%%%%%%%%%%%%%%%%%%%%%%%%%%%
%%%%%%%%%%%%%%%%%%%%%%%%%%%%%%%%%%%%%%%%%%%%%%%%%%%%%%%%%%%%%%%%%%%%%%%%%%%%%%%
%\end{comment}

%%%%%%%%%%%%%%%%%%%%%%%%%%%%%%%%%%%%%%%%%%%%%%%%%%%%%%%%%%%%

\end{document}